\documentclass[sigconf]{acmart}
\usepackage[capitalize]{cleveref}
\usepackage{graphicx}
\usepackage{amsmath}
\usepackage{booktabs}
\usepackage{bbm}
\usepackage[linesnumbered,ruled,vlined]{algorithm2e}
\usepackage{enumitem}
\usepackage{amsthm}
\usepackage{bbold}
\usepackage{mathtools}
\usepackage{algorithmic}
\usepackage{stackengine}
\usepackage{subcaption}

\newtheorem{theorem}{Theorem}%[section]
\newtheorem{lemma}[theorem]{Lemma}

\floatstyle{ruled}
\newfloat{listing}{tb}{lst}{}
\floatname{listing}{Listing}
\newtheorem{proposition}{Proposition}

\SetKwInput{KwInput}{Input}                % Set the Input
\SetKwInput{KwOutput}{Output}              % set the Output
\SetKwInput{Initialize}{Initialize}
\DeclareMathOperator*{\argsortDesc}{arg\,sortDesc}
\DeclareMathOperator*{\argsortAsc}{arg\,sortAsc}
\DeclareMathOperator*{\argmax}{arg\,max}

\usepackage{xspace} % Needed for et al.
\newcommand{\ie}{\emph{i.e.,}\xspace}
\newcommand{\eg}{\emph{e.g.,}\xspace}

\AtBeginDocument{%
  \providecommand\BibTeX{{%
    \normalfont B\kern-0.5em{\scshape i\kern-0.25em b}\kern-0.8em\TeX}}}

\copyrightyear{2022}
\acmYear{2022}
\setcopyright{acmcopyright}\acmConference[KDD '22]{Proceedings of the 28th ACM SIGKDD Conference on Knowledge Discovery and Data Mining}{August 14--18, 2022}{Washington, DC, USA}
\acmBooktitle{Proceedings of the 28th ACM SIGKDD Conference on Knowledge Discovery and Data Mining (KDD '22), August 14--18, 2022, Washington, DC, USA}
\acmPrice{15.00}
\acmDOI{10.1145/3534678.3539264}
\acmISBN{978-1-4503-9385-0/22/08}

\begin{document}

%%
%% The "title" command has an optional parameter,
%% allowing the author to define a "short title" to be used in page headers.
%\title{Active Deep Multiple Instance Learning}
\title{Balancing Bias and Variance for Active Weakly Supervised Learning}
\author{Hitesh Sapkota 
}
\email{hxs1943@rit.edu}
\affiliation{%
  \institution{Rochester Institute of Technology}
  \city{Rochester, NY}
  \country{USA}}

\author{Qi Yu 
}
\email{qi.yu@rit.edu}
\affiliation{%
  \institution{Rochester Institute of Technology}
  \city{Rochester, NY}
  \country{USA}}

%Rochester Institute of Technology\\
%{\tt\small \{hxs1943, qi.yu\}@rit.edu}
%\author {
%    % Author
%    Anonymous Authors 
%}

%%
%% By default, the full list of authors will be used in the page
%% headers. Often, this list is too long, and will overlap
%% other information printed in the page headers. This command allows
%% the author to define a more concise list
%% of authors' names for this purpose.
%\renewcommand{\shortauthors}{Anonymous}

%%
%% The abstract is a short summary of the work to be presented in the
%% article.
\begin{abstract}
As a widely used weakly supervised learning scheme, modern multiple instance learning (MIL) models achieve competitive performance at the bag level. However, instance-level prediction, which is essential for many important applications, remains largely unsatisfactory. We propose to conduct novel active deep multiple instance learning that samples a small subset of informative instances for annotation, aiming to significantly boost the instance-level prediction. A variance regularized loss function is designed to properly balance the bias and variance of instance-level predictions, aiming to effectively accommodate the highly imbalanced instance distribution in MIL and other fundamental challenges.  Instead of directly minimizing the variance regularized loss that is non-convex, we optimize a distributionally robust bag level likelihood as its convex surrogate. The robust bag likelihood provides a good approximation of the variance based MIL loss with a strong theoretical guarantee. It also automatically balances bias and variance, making it effective to identify the potentially positive instances to support active sampling. The robust bag likelihood can be naturally integrated with a deep architecture to support deep model training using mini-batches of positive-negative bag pairs. Finally, a novel P-F sampling function is developed that combines a probability vector and predicted instance scores, obtained by optimizing the robust bag likelihood. By leveraging the key MIL assumption, the sampling function can explore the most challenging bags and effectively detect their positive instances for annotation, which significantly improves the instance-level prediction. Experiments conducted over multiple real-world datasets clearly demonstrate the state-of-the-art instance-level prediction achieved by the proposed model. 
\end{abstract}

\begin{CCSXML}
<ccs2012>
   <concept>
       <concept_id>10010147.10010257.10010282.10011304</concept_id>
       <concept_desc>Computing methodologies~Active learning settings</concept_desc>
       <concept_significance>500</concept_significance>
       </concept>
   <concept>
       <concept_id>10010147.10010257.10010282</concept_id>
       <concept_desc>Computing methodologies~Learning settings</concept_desc>
       <concept_significance>500</concept_significance>
       </concept>
   <concept>
       <concept_id>10010147.10010257</concept_id>
       <concept_desc>Computing methodologies~Machine learning</concept_desc>
       <concept_significance>500</concept_significance>
       </concept>
 </ccs2012>
\end{CCSXML}

\ccsdesc[500]{Computing methodologies~Active learning settings}
\ccsdesc[500]{Computing methodologies~Learning settings}
\ccsdesc[500]{Computing methodologies~Machine learning}

% \begin{CCSXML}
% <ccs2012>
%   <concept>
%       <concept_id>10010147</concept_id>
%       <concept_desc>Computing methodologies</concept_desc>
%       <concept_significance>500</concept_significance>
%       </concept>
%   <concept>
%       <concept_id>10010147.10010257.10010293.10010315</concept_id>
%       <concept_desc>Computing methodologies~Instance-based learning</concept_desc>
%       <concept_significance>500</concept_significance>
%       </concept>
%   <concept>
%       <concept_id>10010147.10010257</concept_id>
%       <concept_desc>Computing methodologies~Machine learning</concept_desc>
%       <concept_significance>500</concept_significance>
%       </concept>
%   <concept>
%       <concept_id>10010147.10010257.10010293</concept_id>
%       <concept_desc>Computing methodologies~Machine learning approaches</concept_desc>
%       <concept_significance>500</concept_significance>
%       </concept>
%  </ccs2012>
% \end{CCSXML}

% \ccsdesc[500]{Computing methodologies}
% \ccsdesc[500]{Computing methodologies~Instance-based learning}
% \ccsdesc[500]{Computing methodologies~Machine learning}
% \ccsdesc[500]{Computing methodologies~Machine learning approaches}
%\ccsdesc[500]{Label Computation~Multiple Instance Active Learning}

%%
%% Keywords. The author(s) should pick words that accurately describe
%% the work being presented. Separate the keywords with commas.
\keywords{multiple instance learning; active learning; weak supervision}

%%
%% This command processes the author and affiliation and title
%% information and builds the first part of the formatted document.
\maketitle

\vspace{-2mm}\section{Introduction}%\vspace{-2mm}
Multiple Instance Learning (MIL) offers an attractive weakly supervised learning paradigm, where instances are naturally organized into bags and training labels are assigned at the bag level to reduce annotation cost \cite{Dietterich1997, Settles2008, Li2015, Sultani2018}. State-of-the-art MIL models achieve competitive performance at the bag level. However, instance-level prediction, which is essential for many important applications (\eg anomaly detection from surveillance videos~\cite{Sultani2018} and medical image segmentation~\cite{Ilse2018}) remains largely unsatisfactory. 

In MIL, a bag is considered to be positive if at least one of the instances is positive otherwise negative \cite{Dietterich1997, Haubmann2017}. To achieve a high bag level prediction, most existing MIL models primarily focus on the most positive instance from a positive bag that is mainly responsible for determining the bag label \cite{Andrews2002, Kim2010, Sultani2018, Haubmann2017}. However, they suffer from two major limitations, which lead to poor instance-level predictions. First, solely focusing on the most positive instance is sensitive to outliers, which are negative instances that look very different from other negative ones \cite{Carbonneau2018MultipleIL}. As a result, these instances may be wrongly assigned a high score indicating they are positive. 
%If these outliers are from a positive bag, the model may still predict the bag label correctly but via the wrongly identified positive instances. 
Second, there may be multiple types (\ie multimodal) of positive instances  in a single bag (\eg different types of anomalies in a surveillance video or different types of skin lesions in a dermatology image). Thus, focusing on a single most positive instance will miss other positive ones. Both cases will result in a low instance-level prediction performance. A possible solution to improve the detection of positive instances is to consider the top-$k$ most positive instances. However, the number of positive instances may vary significantly across different bags and applying the same $k$ to all bags may be inappropriate. Furthermore, finding an optimal $k$ for each bag is highly challenging as it takes a discrete value. 

\begin{figure*}[t!]
\centering
\begin{subfigure}{0.19\textwidth}
  \centering
  \vspace{0mm}
  \includegraphics[width=0.96\linewidth]{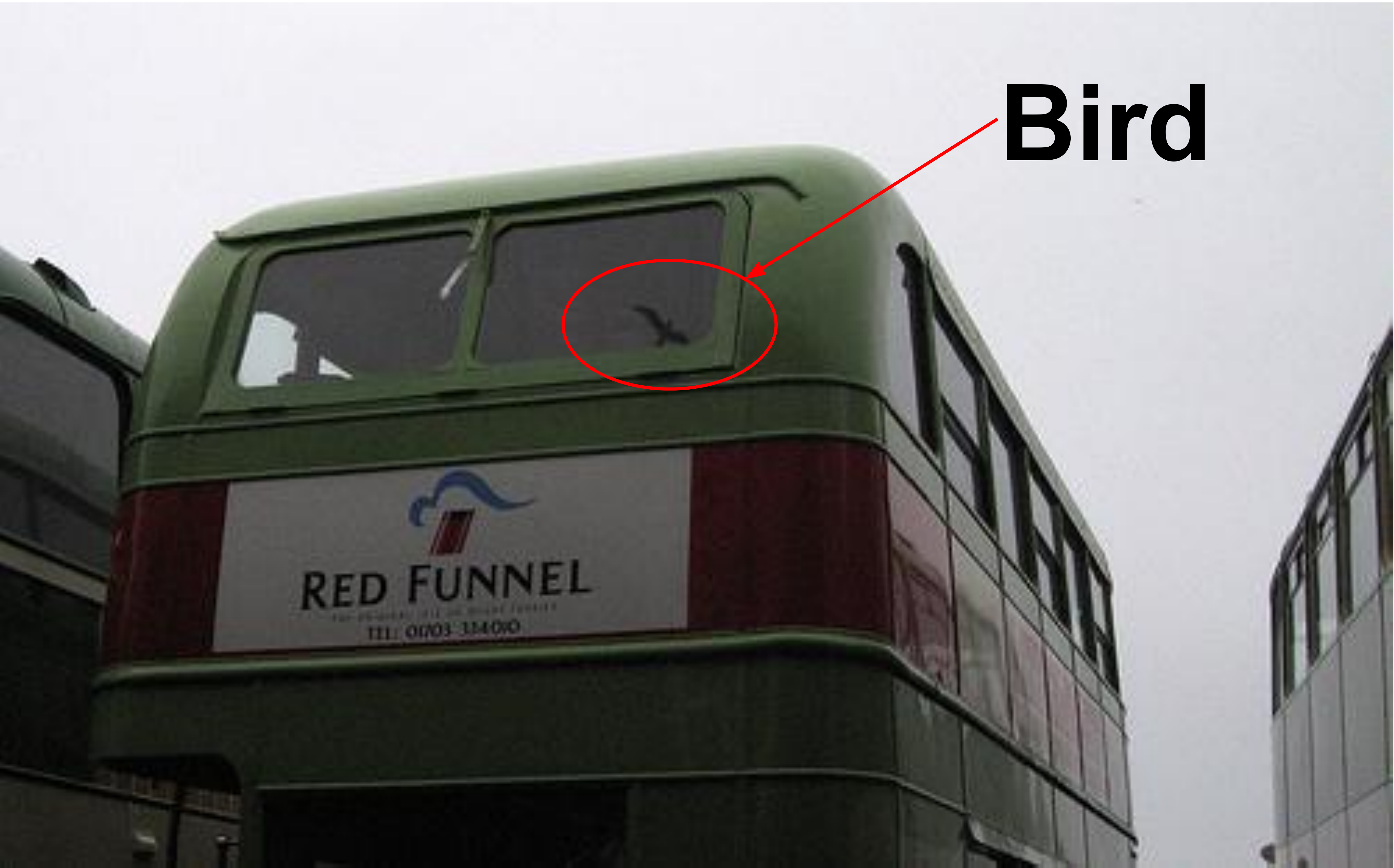}
\vspace{4mm}
  \caption{{\sc Sample bag B$_1$}}
\end{subfigure}%
\begin{subfigure}{0.19\textwidth}
  \centering
  \includegraphics[width=\linewidth]{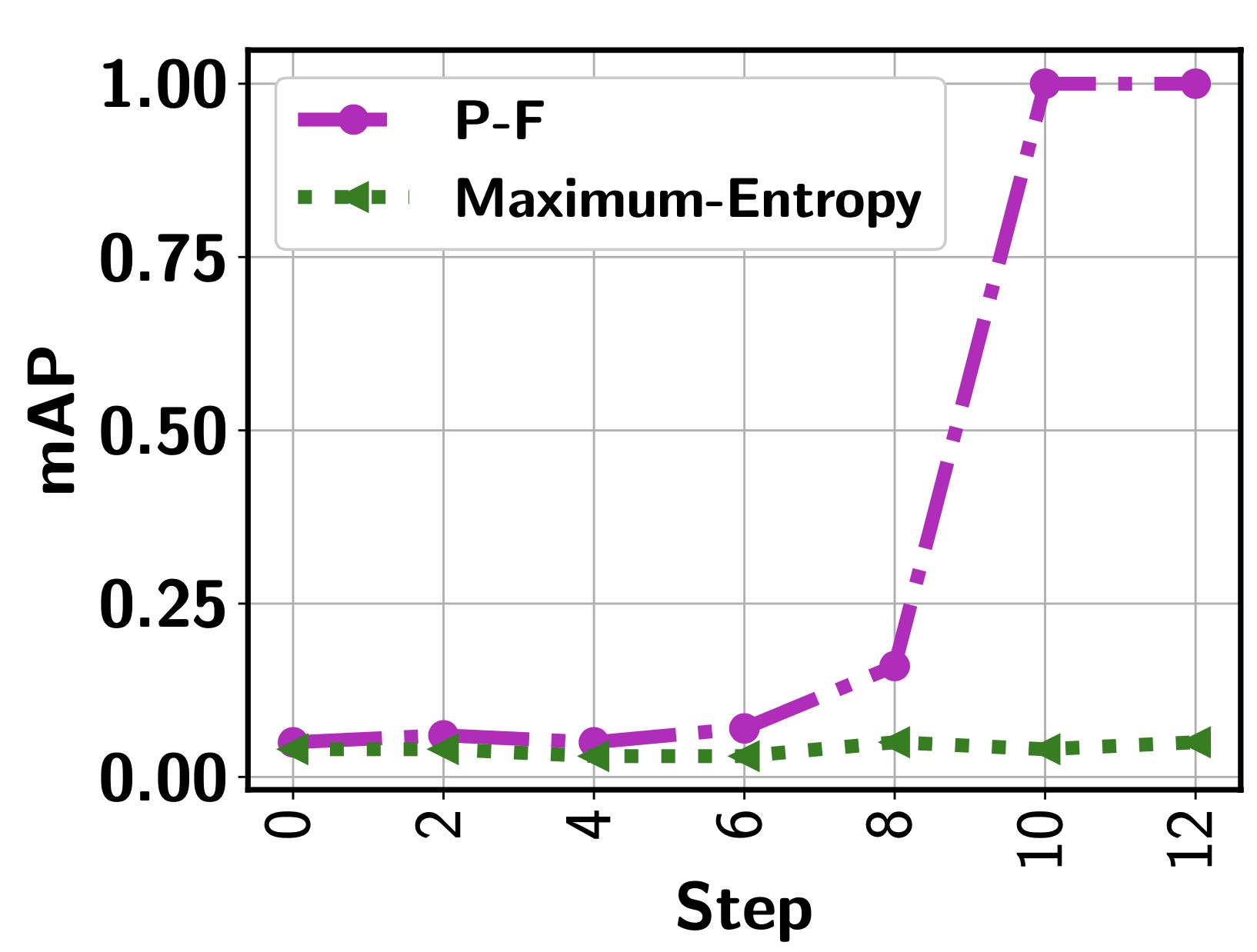}
  \vspace{-4.5mm}
  \caption{MAP Score}
\end{subfigure}%
\begin{subfigure}{0.19\textwidth}
  \centering
  \includegraphics[width=\linewidth]{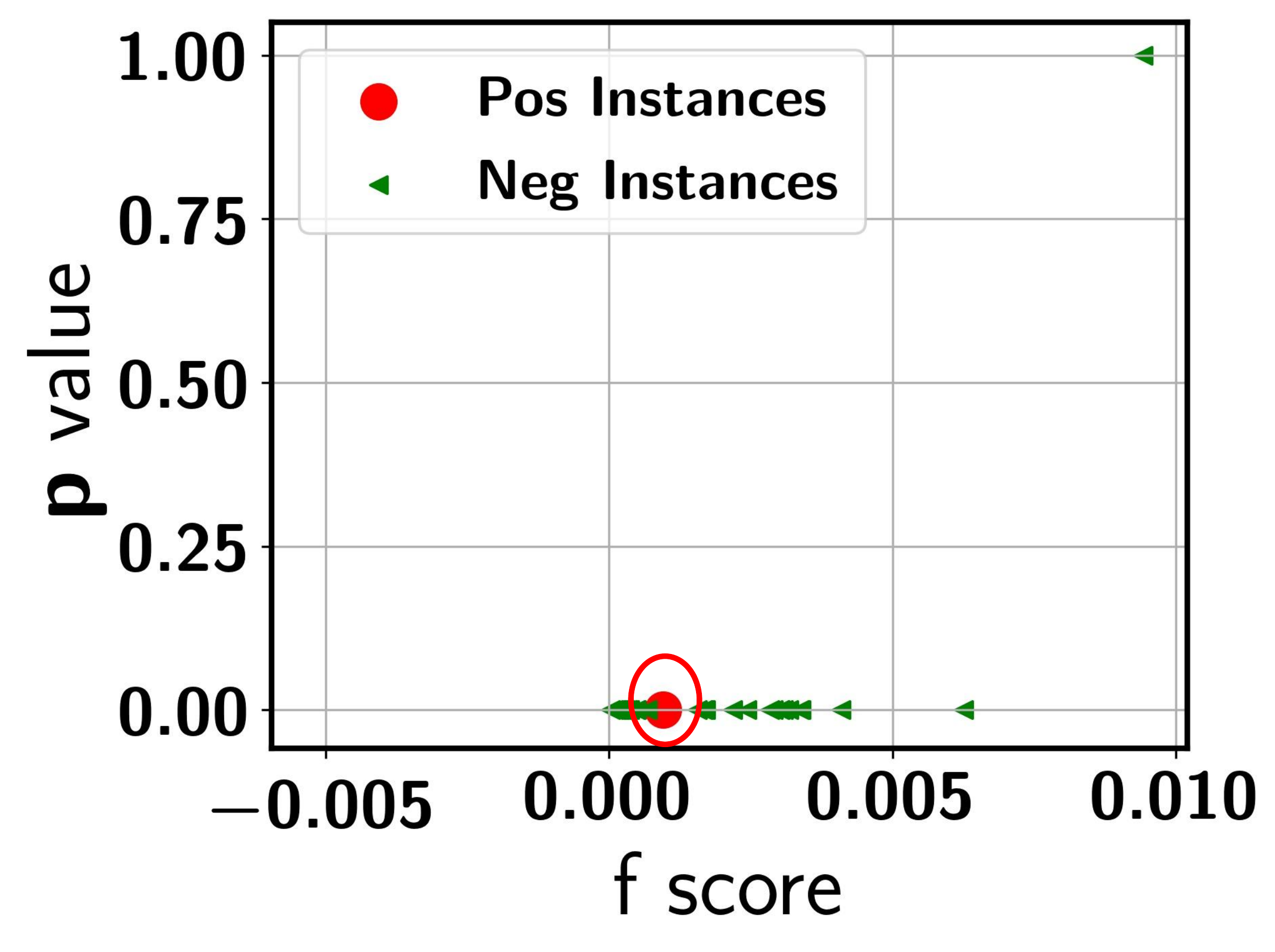}
  \vspace{-5mm}
  \caption{Step 0}
\end{subfigure}%
\begin{subfigure}{0.19\textwidth}
  \centering
  \includegraphics[width=\linewidth]{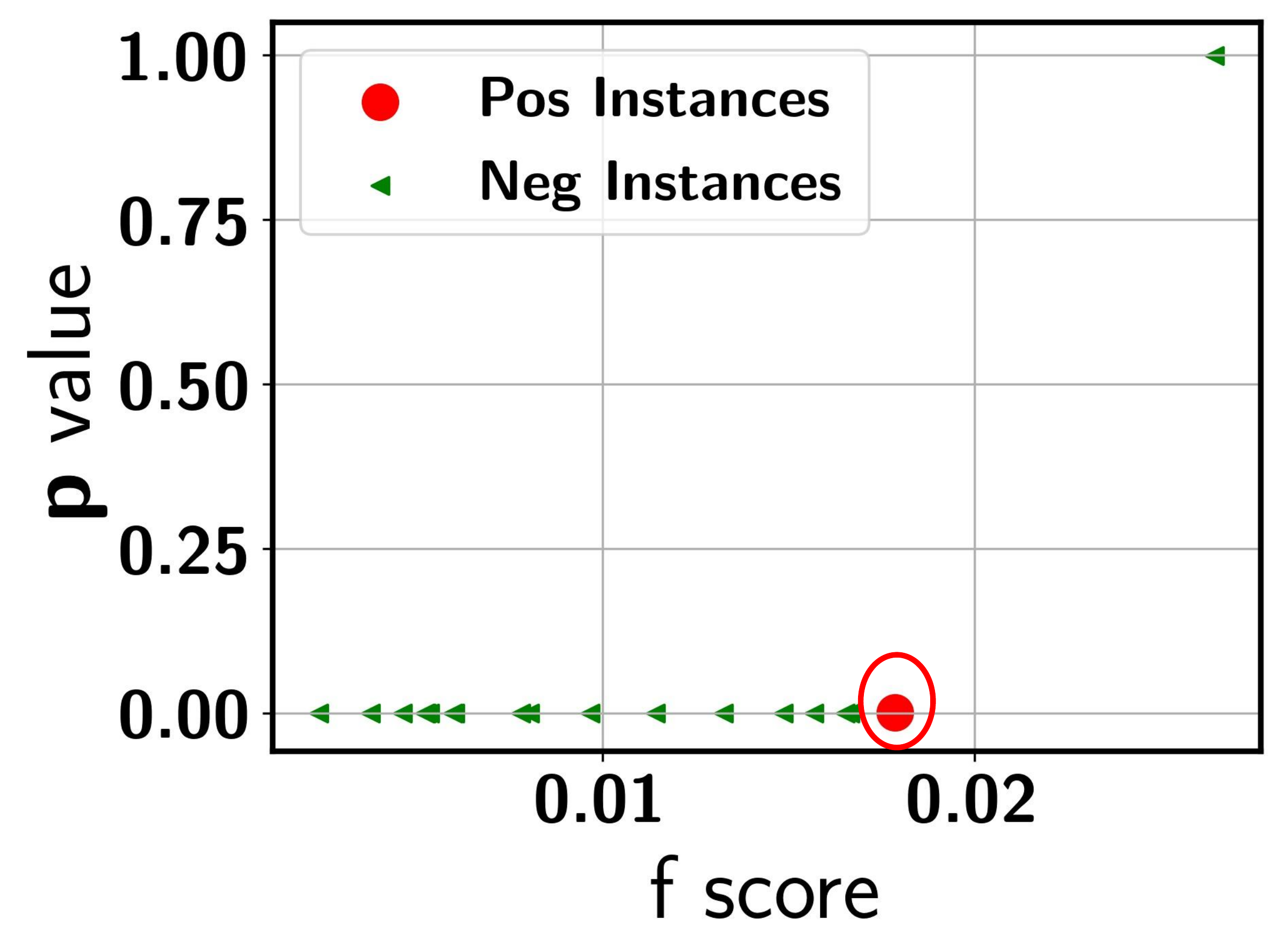}
  \vspace{-5mm}
  \caption{Step 8}
\end{subfigure}%
\begin{subfigure}{0.19\textwidth}
  \centering
  \includegraphics[width=\linewidth]{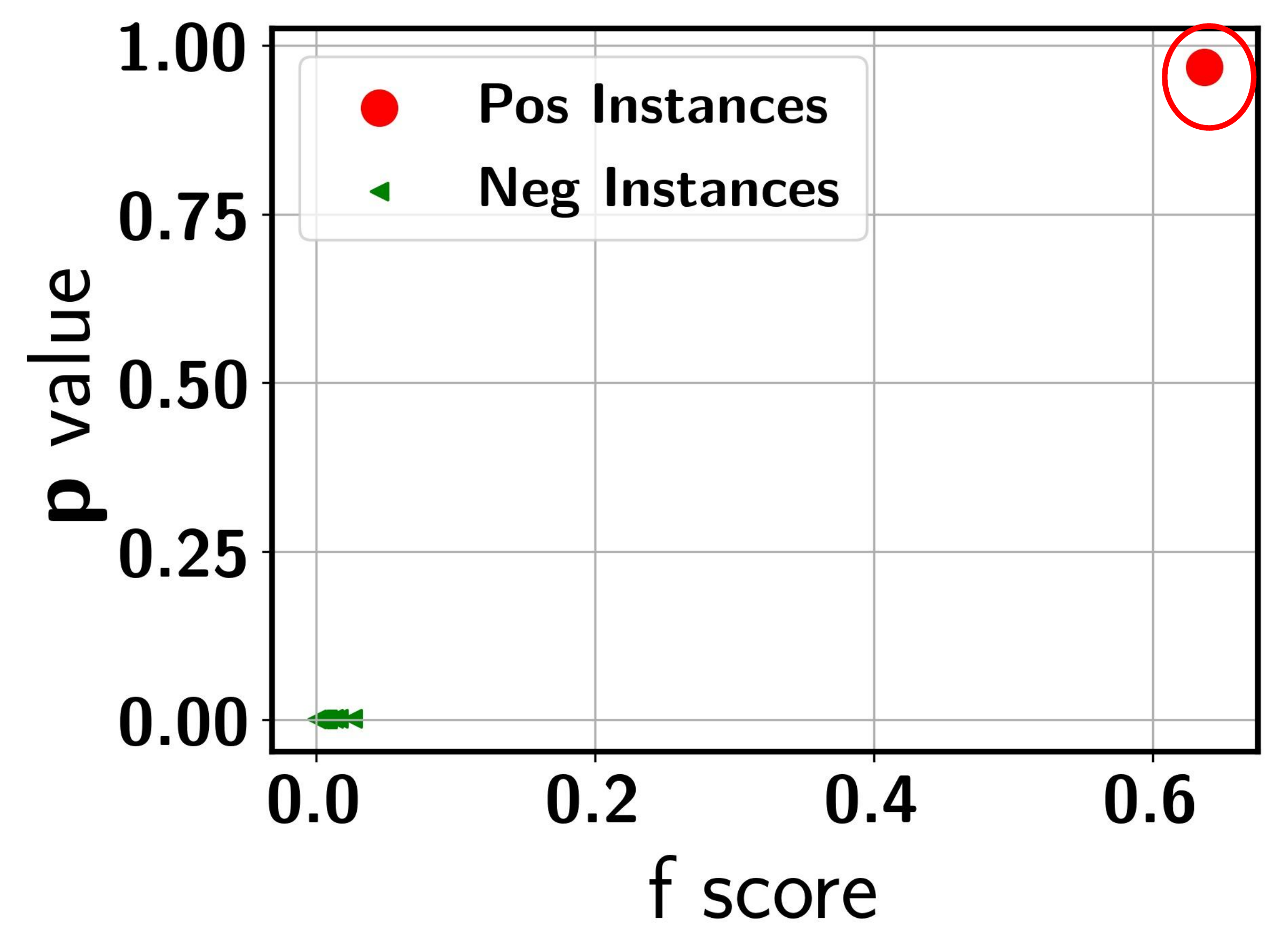}
  \vspace{-5mm}
  \caption{Step 10}
\end{subfigure}%
\vspace{-2mm}
\caption{(a) Example of a challenging bag; (b) MI-AL performance on instance-level predictions; (c)-(e) Prediction scores of instances in the bag in different MI-AL steps}
\label{fig: illustrative_examples_max_vs_proposed}
\vspace{-4mm}
\end{figure*}

The underlying reason for the less accurate instance-level prediction is due to the lack of instance labels. For positive instances that are relatively rare across bags, detecting them by only relying on bag labels is inherently challenging as the weakly supervised signal (\ie bag label) cannot be propagated to the instance level without sufficient statistical evidence. One promising direction to tackle this challenge is to augment MIL with active learning (AL). Multiple instance AL (or MI-AL) aims to select a small number of informative instances to improve the instance level prediction in MIL. In most MIL problems, the data is highly imbalanced at the instance level, where the positive ones are much more sparse. Since the positive instances usually carry more important information, a primary goal of MI-AL is to effectively sample the positive instances from a candidate pool dominated by the negative ones. If a true positive instance can be sampled and labeled, it can help to identify other similar positive instances in the same and different bags, which will significantly improve the instance-level predictions.

However, existing MIL models may easily miss some rare positive instances \cite{Sultani2018}. They may also focus on the wrongly identified negative instances due to their sensitivity to outliers or incapability of handling multimodal bags. Thus, the true positive instances may be assigned a low prediction score, indicating that they are predicted as negative with a high confidence. As a result, commonly used uncertainty based sampling will miss these important instances. Figure~\ref{fig: illustrative_examples_max_vs_proposed} (a) shows a challenging bag, which is an image that contains the shadow of a bird (as the positive class). The positive instances are patches that cover (part of) the bird shadow. Figure~\ref{fig: illustrative_examples_max_vs_proposed} (b) shows that combining uncertainty sampling with a maximum score based MIL model (the green curve) is not able to sample effectively so that instance-level prediction remains very low over the AL process. Figure~\ref{fig: illustrative_examples_max_vs_proposed} (c) further confirms that the initial prediction score (F-score) of the positive instance is close to 0, making it hard to be sampled.

We propose a novel MI-AL model for effective instance sampling to significantly boost the instance-level prediction in MIL. We design an unique variance regularized MIL loss that encourages a high variance of the prediction scores to address bags with a highly imbalanced instance distribution and/or those with outliers and multimodal scenarios. Since the variance regularizer is non-convex, we propose to optimize a distributionally robust bag likelihood (DRBL), which provides a good convex approximation of the variance based loss with a strong theoretical guarantee. The DRBL automatically adjusts the impact of the bag-level variance, making it more effective to detect potentially positive instances to support active sampling. It can also be naturally integrated with a deep architecture to support deep MIL model training using mini-batches of positive-negative bag pairs. Finally, a novel P-F sampling function is developed  that combines a probability vector (\ie $\bf p$) and predicted instance scores (\ie $\bf f$), obtained by optimizing the DRBL. By leveraging the key MIL assumption, the sampling function can explore the most challenging bags and effectively detect their positive instances for annotation, which significantly improves the instance-level prediction. Novel batch-mode sampling is developed to work seamlessly with the deep MIL, leading to a powerful active deep MIL (ADMIL) model to support sampling of high-dimensional data used in most MIL applications. Figure~\ref{fig: illustrative_examples_max_vs_proposed} (b) shows the proposed model (purple curve) that significantly improves instance predictions. Figures~\ref{fig: illustrative_examples_max_vs_proposed} (c)-(e) show P-F sampling dynamically updates the probability $\bf p$ and score $\bf f$ values to effectively sample the positive instance from the highly challenging bag in a few steps. 

Our main contribution includes: (i) an unique variance regularized MIL loss and its convex surrogate that address inherent MIL challenges to best support active sampling, (ii) a novel P-F sampling function to effectively explore most challenging bags with rare positive instances, (iii) mini-batch training and batch-mode active sampling to support ADMIL in broader MIL applications, and (iv) state-of-the-art instance prediction performance in MIL while maintaining low instance annotations.

\vspace{-4mm}
\section{Related Work}

%In this section, we will discuss the related work that is most relevant to ours, including active learning and multiple instance learning.  

{\bf Multiple Instance Learning (MIL).}
 %After being introduced by Dietterich et al.~\citep{Dietterich1997},  various approaches have been proposed to address MIL problems  \citep{Zhou2002, Zhang2017, Wei2014}. 
Existing supervised learning models have been leveraged to tackle MIL problems, including SVM \cite{Andrews2002}, boosting \cite{Xu2004}, graph-based models \cite{Zhou2009}, attention based \cite{Ilse2018, Hsu2020QueryDrivenML}, conditional random field  \cite{Deselaers2010} and Gaussian Processes  \cite{Haubmann2017, Kim2010}. Other approaches try to relax the MIL assumption, which allows positive instances in a negative bag to handle noisy bags \cite{Li2015}. As MIL is commonly applied to video anomaly detection and image segmentation that involve high dimensional data, deep neural networks (DNNs) have become a popular choice for training MIL models \cite{Ilse2018,Sultani2018,Hsu2020QueryDrivenML}. Despite the significant progress made so far, most existing models focus on improving the bag-level predictions. As a result, instance-level performance still falls short in meeting the high standard in critical applications~\cite{Sultani2018,Ilse2018,Haubmann2017}. The proposed ADMIL model aims to fill out this critical gap by augmenting MIL with novel active sampling strategies to significantly boost instance predictions using limited labeled instances to maintain a low annotation cost. 

\vspace{1mm}\noindent{\bf Active Learning (AL).} %A central component of AL is an acquisition function to sample informative data samples for annotation.  Design of acquisition functions depends on the specific machine learning task. 
%Design of AL models depends on the specific machine learning task. For binary classification problems, the primary goal is to estimate a data point's impact on the current decision boundary~\citep{tong2001support,seung1992query,schohn2000less,freund1997selective}.  For multi-class problems,  significant statistics of the data, such as 
Uncertainty and margin based measures are commonly leveraged in existing AL models to achieve efficient data sampling~\cite{settles2009active}. %~\cite{roy2001toward,holub2008entropy,rajan2008active,joshi2009multi}. 
Distributionally robust optimization has also been adopted in multi-class AL to address sampling bias and imbalanced data distribution~\cite{Zhu2019}. 
%For multi-label problems, AL mainly focuses on leveraging label correlation as an additional source to guide sampling in the label space~\citep{li2013active,reyes2018effective,vasisht2014active,yang2009effective,Kapoor:2012:MCU:2999325.2999430,vasisht2014active}. 
Deep learning (DL) models are good candidates for AL because of their high-dimensional data processing and automatic feature extraction capability. Existing models mainly target at improving uncertainty quantification of the network for reliable sampling~\cite{wang2016cost,gal2015bayesian,kendall2015bayesian,leibig2017leveraging}. Batch-mode sampling is commonly used in active DL to avoid frequent model re-training. It focuses on constructing representative batches to avoid redundant information given by similar instances~\cite{kirsch2019batchbald,sener2017active,ash2019deep}. AL in the MIL setting has been rarely investigated. One exception is the MI logistic model and its three uncertainty measures to simultaneously consider both instance and bag level uncertainty~\cite{settles2007multiple}. However, uncertainty sampling is ineffective to explore challenging bags, where all instances are confidently predicted as negative. In addition, the original model is a simple linear model, which does not provide sufficient capacity for high-dimensional data. There is no systematic way to support batch-mode sampling, either. A reinforcement learning based AL technique is developed in \cite{Casanova2020ReinforcedAL}, where segments are chosen to be labeled in each AL step . However, segmentation level annotations are required to compute the reward during the training process, which violates the assumption of MIL.  Another AL framework is developed for MIL tasks in \cite{Yuan2021}. However, sampling is conducted at the bag level (\ie choosing bags instead of instances).  Thus, it is essentially a multi-label AL model, aiming to improve the bag-level predictions with fewer annotated bags. This is fundamentally different from the design goal of ADMIL. 

\vspace{-2mm}
\section{Methodology}

%Based on a standard MIL assumption, we consider that for a positive bag, there is at least one positive instance whereas in case of negative bag, all instances are of negative type. Further, in our setting we consider that the number of instances may vary from one bag to another.
%Let ${\bf X} = \{{\bf x}_1, ..., {\bf x}_N\}$ denote a set of training bag instances with 
Let $\{{\bf x}_1,...,{\bf x}_n\}$ denote a set of instances associated with each bag $\mathcal{B}$, where each ${\bf x}_i\in \mathbb{R}^D$ is a feature vector. Let $t_{\mathcal{B}} \in \{+1,-1\}$ indicates the bag type.  
%All symbols are summarized by Table~\ref{tab: symbol_table}.
Following the standard MIL assumption discussed earlier, active sampling will focus on instances in the positive bags as all instances in a negative bag are negative. We also allow the number of instances to vary from one bag to another.

\vspace{-2mm}
\subsection{Variance Regularization}%\vspace{-1mm}
%\qi{Only bold the symbol not the subscript. Make sure the symbols are used consistently, e.g., bold for vectors.} 
Let ${\bf x}_i^+$ (or ${\bf x}_j^-$) be the $i^{th}$ (or $j^{th}$) instance in a positive bag $\mathcal{B}_{pos}$ (or a negative bag $\mathcal{B}_{neg}$). Following the MIL assumption, a commonly used loss function for training deep MIL models is to make the maximum prediction score of instances from a positive bag to be higher than a negative bag \cite{Sultani2018}. We define as
\begin{align}
     \mathcal{L}^\text{MS}=\left\{1-\max_{i\in \mathcal{B}_{pos}}[f({\bf x}_i^+;{\bf w})]+\max_{j\in \mathcal{B}_{neg}}[f({\bf x}_j^-;{\bf w})]\right\}_+\label{eq:max_mil}
\end{align}
where $f({\bf x};{\bf w}) \in [0,1]$ is the prediction score of instance ${\bf x}$ provided by a deep neural network parameterized by ${\bf w}$ and $[a]_+ = \max\{0, a\}$. We will omit ${\bf w}$ from $f({\bf x};{\bf w})$ to keep the notation uncluttered. The above objective function aims to maximize the gap between the maximum prediction score of instances from a positive bag and maximum score from a negative bag. Model training can be performed by sampling pairs of positive and negative bags $(\mathcal{B}_{pos}, \mathcal{B}_{neg})$, using their bag-level labels to evaluate the loss, and performing back-propagation. The maximum score based MIL (referred to as MS-MIL) models are designed primarily for bag label prediction as it aims to identify a single most positive instance from a positive bag and maximizes its prediction score. In this way, it fully leverages the MIL assumption (\ie at least one positive instance in $\mathcal{B}_{pos}$) and the weakly supervised signal (\ie bag-level label).

As discussed earlier, MS-MIL and its top-$k$ extensions suffer from key limitations that impact their instance-level prediction performance. Meanwhile, they provide inadequate support to sample the most informative instances to enhance the instance predictions. Inspired by the recent advances in learning theory to automatically balance bias and variance in risk minimization~\cite{Duci2019}, we propose a novel variance regularized MIL loss function to capture the inherent characteristics of  MIL, aiming to collectively address highly imbalanced instance distribution, existence of outliers, and multimodal scenarios. As a result, minimizing the new MIL loss can effectively improve the prediction scores of the positive instances, making them easier to be sampled for annotation by the proposed sampling function. In particular, the variance regularized loss introduces {\em two novel changes} to \eqref{eq:max_mil}, which are formalized below:
\begin{align}
 \mathcal{L}^\text{VAR} = \left\{1-\left[\frac{1}{n}\sum_{i=1}^nf({\bf x}_i^+)+C\sqrt{\frac{\text{Var}_n[f(X^+)]}{n}}\right]+\max_{j\in \mathcal{B}_{neg}}\left[f({\bf x}_j^-)\right]\right\}_+\label{eq:variance_regularized}
\end{align}
where $\forall i \in [1,n], {\bf x}_i^+\in \mathcal{B}_{pos}$, $n$ is the size of $\mathcal{B}_{pos}$, $\text{Var}_n$ is the empirical variance of $f(X^+)$ with $X^+$ being a random variable representing an instance from a positive bag, and parameter $C$ balances the mean score and the variance. 

The first key change is to use the mean score to replace the maximum score in \eqref{eq:max_mil}, which avoids the model to only focus on the most positive instance in a bag to make it robust to outliers and multimodal scenarios. Since positive bags are guaranteed to include positive instances and instances in a negative bag are all negative, it is desirable that the mean score for a positive bag should be high. Maximizing the mean score in a positive bag using a complex model (\eg a DNN) could effectively reduce the training loss (by reducing the bias) in estimating the bag-level labels. However, using the mean score alone is problematic as most instances in a positive bag are usually negative in a typical MIL setting. As a result, such a low bias model will lead to a very high false positive rate, which negatively impacts the overall instance-level prediction. The proposed loss function addresses this issue through the novel variance term, which effectively handles the highly imbalanced instance distribution. With only a small number of instances being truly positive, the empirical variance $\text{Var}_n$ for the bag should be high due to the large deviation of a small number of high scores from the majority of low scores. It is worth to note that the variance term in \eqref{eq:variance_regularized} plays a distinct role than risk minimization in standard supervised learning, where it is minimized to control the estimation error. In contrast, the variance in \eqref{eq:variance_regularized} is encouraged to be large to allow a small set of instances in a bag to be positive, aiming to precisely capture the imbalanced distribution. To our best knowledge, this is the first bias-variance formulation in the MIL setting. 

Conducting MI-AL using variance regularization still faces two  challenges. First, its effectiveness  hinges on an optimal balance between the mean score and the empirical variance, which is controlled by the hyperparameter $C$. Similar to the standard supervised learning, there lacks a systematic way of setting such a hyperparameter to achieve an optimal trade-off. Second, the variance term is non-convex with multiple local minima~\cite{Duci2019}, which makes model training much more difficult and time-consuming. Thus, it is not suitable for real-time interactions to support active sampling.

\vspace{-2mm}
\subsection{Distributionally Robust Bag Likelihood}\label{sec:drbl}%\vspace{-1mm}
To address the challenges as outlined above, we propose to formulate a distributionally robust bag level likelihood (DRBL) as a convex surrogate of the variance regularized loss in \eqref{eq:variance_regularized}. By extending the distributionally robust optimization framework developed for risk minimization in supervised learning~\cite{Namkoong2017,Duci2019}, we theoretically prove the equivalence between DRBL and variance regularization with high probability. Being convex, DRBL is easier to optimize that  facilitates MIL model training to support fast active sampling. Furthermore, by setting a proper uncertainty set as introduced next, we show that the  parameter $C$ is directly obtained when optimizing the DRBL, where the instance distribution in the bag is constrained by the uncertainty set. As a result, it achieves automatic trade-off between the mean prediction score and the variance. 

We first introduce a probability vector ${\bf p}=(p_1,...,p_n)^{\top}$, where $\sum_i p_i=1,p_i\ge 0, \forall i \in \{1,...,n\}$ and let $p_i$ denote the probability that instance ${\bf x}_i^+\in \mathcal{B}_{pos}$ can represent the bag. We further introduce a binary indicator vector ${\bf z}=(z_1,...,z_n)^{\top}$, where $p(z_i=1)=p_i$. Let $Y$ be a binary random variable that denotes the bag label. Conditioning on all the instances in the bag, the (conditional) bag likelihood for bag $\mathcal{B}_{pos}$ is given by $p(Y=1|{\bf z},{\bf f})=\prod_i f({\bf x}_i^+)^{z_i}$, where ${\bf f}=(f({\bf x}_1^+),...,f({\bf x}_n^+))^{\top}$. By integrating out the indicator variables, we have the marginal bag likelihood as $p(Y=1|{\bf p},{\bf f})=\sum_i p_if({\bf x}_i^+)$. Instead of letting a single most positive instance to determine the bag label, where $p(y=1|{\bf p},{\bf f})= f({\bf x}_k^+)$ with $k=\arg \max_i f({\bf x}_i^+)$, which is equivalent to MS-MIL, or assigning equal probability to each instance (\ie $p_i=1/n$), which is equivalent to the mean score, we introduce an uncertainty set $\mathcal{P}_n$ that allows ${\bf p}$ to deviate from a uniform distribution to some extent:
\begin{equation}
\mathcal{P}_n :=\left\{{\bf p}\in \mathbb{R}^n, {\bf p}^{\top}\mathbbm{1}=1, 0\leq {\bf p}, D_f\left({{\bf p}||\frac{\mathbbm{1}}{n}}\right)\leq \frac{\lambda}{n}  \right\}\label{eq:uncertainty_dro}
\end{equation}
where $D_{f}({\bf p}||{\bf q})$ is the $f$-divergence between two distributions $\bf p$ and $\bf q$, $\mathbbm{1}$ is a $n$-dimensional unit vector, and $\lambda$ controls the extent that ${\bf p}$ can deviate from a uniform vector, which essentially corresponds to the imbalanced instance distribution in the bag. Note that $\mathcal{P}_n$ only specifies a neighborhood that ${\bf p}$ may deviate from a uniform distribution. Since $\mathcal{P}_n$ is a convex set, an optimal ${\bf p}$ can be easily computed for each specific bag by optimizing the robust bag likelihood according to its specific imbalanced instance distribution. This is fundamentally more advantageous than a top-$k$ approach, where $k$ is discrete and hard to optimize. Next, we show that the optimal robust bag likelihood is equivalent to the variance regularized mean prediction score with high probability, which allows us to define a new MIL loss based on DRBL.

%To solve the limitations presented in the variance-regularized technique, we propose DRO based objective function that can approximate the variance regularized by high probability.

\begin{theorem}
Let $X^+$ be a random variable representing an instance from a positive bag, $f(X^+) \in [0, 1]$ is the score assigned to an instance, $\sigma^2=\text{Var}[f(X^+)]$ and $\text{Var}_n[f(X^+)]$ denote the population and sample variance of $f(X^+)$, respectively, and $D_{f}$ takes the form of $\chi^{2}$-divergence. For a fixed $\lambda$ and with $n\geq \max(2, \frac{\lambda}{\sigma^2}\max(8\sigma, 44))$, 
%$\theta \in \Omega$ and $D_{f}(p, q)$ be the squared Euclidean distance between two distributions $p$ and $q$. If $n\geq \max(2, \frac{\lambda}{\sigma^2}\max(8\sigma, 44)$ then with probability at least $1-\exp\left(-{\frac{3n\sigma^2}{10}}\right)$ we have
\begin{equation}
\max_{{\bf p}\in {\mathcal{P}_n}}\sum_{i=1}^np_if({\bf x}_i^+) = \frac{1}{n}\sum_{i=1}^nf({\bf x}_i^+)+\sqrt{\frac{\lambda Var_n[f(X^+)]}{n}} 
\label{eq:dro-variance-eqivalence}
\end{equation}
with probability at least $1-\exp\left(-{\frac{7n\sigma^2}{20}}\right)$, where $\mathcal{P}_n$ is an uncertainty set defined by \eqref{eq:uncertainty_dro}. 
\label{th:dro_variance_equi_formula}
\end{theorem}
It is worth to note that given the highly imbalanced positive instances in a typical MIL setting, the true variance $\sigma^2$ should be high. For a bag with a decent size, it guarantees the equivalence in \eqref{eq:dro-variance-eqivalence} with high probability. Furthermore, maximizing the robust bag likelihood given on the l.h.s. of \eqref{eq:dro-variance-eqivalence} assigns $C=\sqrt{\lambda}$, which automatically adjusts the impact of variance based on the uncertainty set. 
%In Theorem \ref{th:dro_variance_equi_formula}, $D_f$ takes the form of $\chi^{2}$-divergence between two distributions.
Theorem \ref{th:KL} below further generalizes this result to the KL-divergence.

\begin{theorem}
Let $X^+$ be a random variable representing an instance from a positive bag, $f(X^+) \in [0, 1]$ is the score assigned to an instance, $\sigma^2=\text{Var}[f(X^+)]$ and $\text{Var}_n[f(X^+)]$ denote the population and sample variance of $f(X^+)$, respectively, and $D_{f}$ takes the form of KL-divergence. We have
\begin{equation}
\begin{aligned}
\max_{{\bf p}\in {\mathcal{P}_n}} & \sum_{i=1}^np_if({\bf x}_i^+) = \frac{1}{n}\sum_{i=1}^nf({\bf x}_i^+)+\sqrt{\frac{2\lambda \text{Var}_n[f(X^+)]}{n}}+\epsilon\left(\frac{\lambda}{n}\right)\\
%\text{s.t.~} & \mathcal{P}_n :=\left\{{\bf p}\in \mathbb{R}^n, {\bf p}^{\top}\mathbbm{1}=1, 0\leq {\bf p}, D_{KL}\left({{\bf p}||{\bf p}_0}\right)\leq \frac{\lambda}{n} \right\}
    \label{eq:dro_kl_variance_equivalence}    
\end{aligned}
\end{equation}
\label{th:KL}
where %${\bf p}_0 = \frac{\mathbbm{1}}{n}$ is the uniform distribution indicating the center of the ball,
$\epsilon\left(\frac{\lambda}{n}\right)=\frac{\lambda}{3n}\frac{\mathcal{\kappa}_3(f(X^+))}{\text{Var}_n[f(X^+)]}+\mathcal{O}\left(\left(\frac{\lambda}{n}\right)^{3/2}\right)$ with $\kappa_3 = \mathbb{E}_0[(f(X^+)-\mathbb{E}_0[f(X^+)])^3]$ and $\mathbb{E}_0$ denotes the expectation taken over ${\bf p}_0$.
\end{theorem}
{\bf Remark}: Given a bag with a decent size $n\gg 1$ and since $\lambda$ is usually set to $\lambda \ll 1$ ($0.01$ is used in our experiments), we have $\epsilon\left(\frac{\lambda}{n}\right)\to 0$. When the empirical variance $\text{Var}_n[f(X^+)]$ is sufficiently large (which is true for MIL), the r.h.s. of \eqref{eq:dro_kl_variance_equivalence} is dominated by the first two terms, which implies 
\begin{equation}
\begin{aligned}
\max_{{\bf p}\in {\mathcal{P}_n}} \sum_{i=1}^np_if({\bf x}_i^+) \approx \frac{1}{n}\sum_{i=1}^nf({\bf x}_i^+)+\sqrt{\frac{2\lambda \text{Var}_n[f(X^+)]}{n}}
\end{aligned}
\end{equation}

Detailed proofs are given in Appendix~\ref{app:proof}.  %For a detailed proof of both Theorems, please refer to Appendix~\ref{app:proof}.    
Leveraging the above theoretical results, we formulate a DRBL-based MIL loss as 
%The corresponding loss after using DRO loss is given as
\begin{equation}
    \mathcal{L}^\text{DRBL} = \left\{1-\max_{{\bf p}\in \mathcal{P}_{n}}\left[\sum_{i=1}^np_if({\bf x}_i^+)\right]+\max_{j\in \mathcal{B}_{neg}}\left[f({\bf x}_j^-)\right]\right\}_+\label{eq:dro_loss_single_bag}
\end{equation}
The DRBL loss offers a very intuitive interpretation on the newly introduced probability vector $\bf p$. Since it can deviate from the uniform distribution as specified by the uncertainty set $\mathcal{P}_{n}$, each entry $p_i$ essentially corresponds to the contribution (or weight) of ${\bf x}_i^+$ to the bag likelihood (being positive). As a result, to maximize the robust bag likelihood, instances with a higher prediction score should receive a higher weight. Meanwhile, constrained by $\mathcal{P}_{n}$, multiple instances will contribute to the bag likelihood with a sizable weight as $\bf p$ cannot deviate too much from being uniform. Hence, their prediction scores will simultaneously be brought up by the model. This makes DRBL robust to the outlier and multimodal cases as it increases the chance for the true positive instances or multiple types of true positive instances to be assigned a high prediction score. This provides fundamental support to the proposed P-F active sampling function that combines the probability vector $\bf p$ and the prediction score $\bf f$ in a novel way to choose the most informative instances in a bag for annotation. 

\vspace{-2mm}
\subsection{P-F Active Sampling}%\vspace{-1mm}
Since we have the prediction score $f({\bf x}_i^+) \in [0,1]$, it can be naturally interpreted as the probability of instance ${\bf x}_i^+$ being positive. A straightforward way to perform uncertainty based instance sampling is to compute the $f$-score based entropy of the instances, referred to F-Entropy: 
\begin{align}
    {\bf x}_*&= \arg \max_{i \in \mathcal{B}_{pos}} H[f({\bf x}_i^+)],  %\quad \text{where}\\
    \label{eq:f-entropy}
\end{align}
where $ H[f] =-[f\log f+(1-f)\log (1-f)]$.
Since the sampled instance has the largest prediction uncertainty (according to F-Entropy), labeling such an instance can effectively improve the model's instance-level performance. Active sampling using \eqref{eq:f-entropy} is straightforward, which involves evaluating $H[f({\bf x}^+)]$ for all the instances from positive training bags (note that all the instances in a negative bag are negative). Since we consider a deep learning model to better accommodate high-dimensional data, 
%including images and videos, which are common for MIL problems,  
sampling one instance at a time requires frequent model training, which is computationally expensive. Instead, we sample a small batch of instances in each step based on their predicted F-Entropy. It is worth to note that, due to the highly imbalanced instance distribution, the majority of the prediction scores, including many positive instances, may be very low. The goal is to assign a relatively higher score to the potentially positive instances so that their entropy is not too low, indicating a confident negative prediction, which will be missed by the sampling function.

As discussed earlier, using the robust bag likelihood as the MIL loss can directly benefit instance sampling by increasing the chance to assign a higher prediction score to a positive instance so that it is more likely to be sampled. However, F-Entropy sampling still suffers from two major limitations. First, for some very difficult bags, such as the sample image shown in Figure~\ref{fig: illustrative_examples_max_vs_proposed} (a), identifying the positive instances (\eg the patch in the image containing the shadow of a bird) can be highly challenging. As a result, they may be assigned a very low $f$ score. In fact, as shown in Figure~\ref{fig: illustrative_examples_max_vs_proposed} (c), all the instances in this bag receive a very low score with the highest less than 0.01, leading to a very low entropy.  Some additional examples of challenging bags from the 20NewsGroup dataset are shown in Figure~\ref{fig:morebagexamples} of Appendix~\ref{app:examplebags}, where all the instances are predicted with a very low score. Hence, all these instances are predicted as negative with low uncertainty, making them less likely to be chosen by entropy based sampling. 
Second, since batch-mode sampling is adopted to reduce the training cost of a deep network, it is essential to diversify the selected instances in the same batch to minimize the annotation cost. However, choosing data instances solely based on their predicted entropy may lead to the annotation of similar instances, which is not cost-effective.

The proposed P-F active sampling overcomes the above two limitations simultaneously through effective bag exploration by combining the probability vector $\bf p$ and the prediction score $\bf f$ through a $\min\max$ function according to their distinct roles in a bag. The key design rationale of P-F sampling is rooted in the standard MIL assumption that ensures at least one positive instance in each positive bag to guide effective bag exploration. Both $\bf p$'s and $\bf f$'s along with the bag structure are dynamically updated during bag exploration to increase the chance of sampling the positive instances in an under-explored bag. A hybrid loss function further utilizes labels of  sampled negative instances in the same bag to boost the prediction scores of the positive instances. More specifically, let $B$ be the total number of positive training bags, P-F sampling will choose the following data instance:
%for a positive $\mathcal{B}_b$, where $b\in \{1,...,B\}$ with $B$ being the total number of positive training bags, we choose instance with the max entry from its probability vector ${\bf p}_b$: 
\begin{equation}
    {\bf x}^{PF}_*=\arg \min_{b\in \{1,...,B\}} f({\bf x}^+_{b_*}), \quad \text{and } b_*=\arg \max {\bf p}_b
    \label{eq:pfsampling}
\end{equation}
where ${\bf p}_b$ is the probability vector of bag $b$. For each bag, the sampling function first identifies the instance ${\bf x}^+_{b_*}$ with the largest $p$ value in each bag. Such an instance can be regarded as the most representative instance in the bag as it makes the largest contribution (according to ${\bf p}_b$) to the bag likelihood. According to the prediction score of ${\bf x}^+_{b_*}$, we can categorize bags into three groups: (1) easy bags, where $f({\bf x}^+_{b_*})$ takes a large value, indicating that the model makes confidently correct predictions, (2) confusing bags, where $f({\bf x}^+_{b_*})$ is reasonably large but uncertain, indicating the model is still confusing about its prediction, and (3) difficult bags, where $f({\bf x}^+_{b_*})$ is very low, indicating the model makes confidently wrong predictions. It is desirable to sample from both confusing and difficult bags as the model already makes accurate instance predictions for easy bags. Sampling instances from the confusing bags can be achieved through the proposed F-Entropy as the model makes uncertain predictions, which leads to a high entropy. Finally, sampling from the difficult bags is fundamentally more challenging due to low prediction scores for the entire bag. However, the MIL assumption provides a general direction for bag-level exploration of positive instances as there must be at least one positive instance in each positive bag. The P-F sampling function in \eqref{eq:pfsampling} chooses the representative instance from the bag with the lowest prediction score. Such an instance is guaranteed to be sampled from an under-explored (\ie difficult) bag as it has the lowest prediction score despite being predicted as the most positive instance in the bag. 

Extension to the batch-mode sampling is conducted in two directions, within a bag and across bags, for more effective exploration while ensuring diversity of the sampled instances. First, instead of only sampling the most positive instance from the identified under-explored bag, we propose to sample $k>1$ instances as the positive instances may be ranked lower than multiple negative instances in the bag according to the current prediction scores (see Figure~\ref{fig: illustrative_examples_max_vs_proposed} (c) for an example). This helps to more effectively explore very difficult bags. To ensure diversity among the sampled instances, we keep $k$ small but sample across multiple bags simultaneously. Only bags with a max prediction score $f({\bf x}^+_{b_*})$ less than a threshold (0.3 is used in our experiments) will be explored as these represent the difficult bags as discussed above. For bags with a larger $f({\bf x}^+_{b_*})$, they are either easy bags or confusing bags that can be effectively sampled using F-Entropy. Our overall P-F sampling function integrates bag exploration and F-Entropy and gives priority to the former to perform diversity-aware bag exploration first. As more bags are successfully explored along with MI-AL, less instances will be sampled by exploration and the focus will be naturally shifted to F-Entropy to perform model fine-tuning. The detailed sampling process is summarized by Algorithm~\ref{alg: sampling}. 

Similar to AL in standard supervised learning, the sampled annotated instances should be used to improve the model prediction performance. However, the MIL loss primarily focuses on the bag-level labels due to the lack of instance labels. To this end, we propose a hybrid loss function that integrates the bag and instance labels.  Let ${\bf X}^l=\{{\bf x}^l_1, {\bf x}^l_2,..., {\bf x}^l_m\}$ be the $m$ labeled instances queried by the proposed active learning function and ${\bf t}^l = \{t^l_1,  t^l_2,..., t^l_m\}$ with $t_l^i\in \{0, 1\}$ be the corresponding instance labels. We formulate a supervised binary cross-entropy (BCE) loss as
\begin{equation}
    L^\text{BCE} = -\frac{1}{m}\sum_{i=1}^m\left[t_i^l\log(f({\bf x}_i^l))+(1-t_i^l)\log(1-f({\bf x}_i^l))\right] \label{eq:bce_loss}
\end{equation}
It is clear that the sampled positive instances provide important supervised signals so that the model will predict a high score for similar positive instances, which will directly benefit instance-level prediction. In contrast, the sampled negative instances, especially those chosen from the under-explored bags, contribute less to improve the prediction performance as their original prediction scores are already low. However, they play a subtle but essential role to achieve more effective bag-level exploration. First, if a sampled instance is labeled as negative, it will be removed from the bag, which does not violate the MIL assumption. Meanwhile, since we have $\sum_ip_i=1$, the $p$ values will be redistributed and the chance for each remaining instance to be sampled is therefore increased. Furthermore, the BCE loss will further bring down the prediction scores of negative instances that are similar to the sampled one. This may help to improve the score of the positive instance so that it can have a higher chance to be sampled in the future. Finally, the hybrid loss that combines the MIL loss and the supervised loss is used to retrain the model after a new batch of instances are queried:
\begin{equation}
\mathcal{L}^\text{Hybrid}= \mathcal{L}^{DRBL}(\mathcal{B}_{pos}, \mathcal{B}_{neg})+\beta \mathcal{L}^{BCE}({\bf X}^l, {\bf t}^l)
    \label{eq:total_loss}
    %\vspace{-2mm}
\end{equation}
where $\beta$ is used to trade-off bag- and instance-level losses.
\vspace{-2mm}

\begin{algorithm}

\DontPrintSemicolon
  
  \KwInput{${\bf p}_{{\mathcal B}_{pos}}$, ${\mathcal Q}_{prev}$, $Th_{PF}$, $Th_{H}$, $BSize$, $k$}
  \KwOutput{$\mathcal{Q}$}
  \KwData{$B$ positive training bags \tcp{Feature vector for each bag}} 
 {\bf Initialization:}  $\mathcal{U}_B = \{\}$, count = 0, $\mathcal{Q}_{P-F} =\{\}$, $\mathcal{Q}_{F} =\{\}$ \;
  \For{$b \in [B]$}{
  ${\bf p}_b\leftarrow {\bf p}_{\mathcal{B}_{pos}}[b]$,
  $b_{*}\leftarrow \argmax {\bf p}_{b} \setminus \mathcal{Q}_{prev}[b]$ \;
  \If{$f({\bf x}_{b_*}^+)\leq Th_{PF}$ } 
    {
        $\mathcal{U}_{B} \leftarrow b_*$ \;
    }
  
  }
  \tcc{Adding instances from unexplored bags}
  $\mathcal{U}_{B} = \argsortAsc_{b_* \in {\mathcal{U}_{B}}}f({\bf x}_{b_*}^+)$ \;
  
  \For{$b_*$ $\in$ $\mathcal{U}_{B}$}{
  \If{$b_{*}$ $\in$ $\mathcal{Q}_{prev}$}
    {
       \If{positive ins $\in$ $\mathcal{Q}_{prev}[b]$}
        {continue}
        }

    \Else{
    $\mathcal{X}^{PF} = \argsortDesc_{b_{*}} \left(f({\bf x}_{b_*}^+) \setminus \mathcal{Q}_{prev}[b_{*}]\right)$[:$k$] \;
    \For{${\bf x}_i \in \mathcal{X}^{PF}$}
    {
    \If{count$\geq$ BSize}
        {break}
        $\mathcal{Q}_{P-F}[b_{*}] \leftarrow {\bf x}_i$ \;
        count $\leftarrow$ count+1\;
    }

    }
  
  }

$\mathcal{Q}_{prev} = \mathcal{Q}_{prev} \cup \mathcal{Q}_{P-F}$ \;

\tcc{Adding instances with highest F-Entropy; $H[f({\bf x}_i^+)] = -\left[f({\bf x}_i^+\log f({\bf x}_i^+))+(1-f({\bf x}_i^+))\log(1-f({\bf x}_i^+))\right]$}

$\mathcal{C}_{idx} = \argsortDesc_{i} \left( H[f({\bf x}_i^+)]\geq Th_H\right)$ \;
 \For{$i \in \mathcal{C}_{idx}$}
    {
    \If{count$\geq$ BSize}
        {break}
        
    \If{${\bf x}^+_i \in \mathcal{Q}_{prev}[b_i]$}
        {break}
        
        $\mathcal{Q}_{F}[b_{i}] \leftarrow {\bf x}^+_i$ \;
        count $\leftarrow$ count+1\;
    }

$\mathcal{Q} = \mathcal{Q}_{prev} \cup Q_{F}$
 
\caption{{\bf P-F Active Sampling \label{alg: sampling}}}
\vspace{-1mm}
\end{algorithm}
%\vspace{-1mm}
\section{Experiments}
We conduct extensive experimentation  over multiple real-world MIL datasets 
to justify the effectiveness of the proposed  ADMIL model.  The purpose of our experiments is to demonstrate: (i) the state-of-the-art instance prediction performance  by comparing with existing competitive baselines, (ii) effectiveness of the proposed P-F active sampling function through comparison with other sampling mechanisms, (iii) impact of key model parameters through a detailed ablation study, and (iv) qualitative evaluation through concrete examples to provide deeper and intuitive insights on the working rationale of the proposed model.

\begin{table*}[t!]
    \caption{Number of positive and negative bags  on different datasets}
\vspace{-4mm}   
%\footnotesize

\begin{center}
\resizebox{1.2\columnwidth}{!}{%
\begin{tabular}{|c|c|c|c|c|c|c|c|c|}
\hline
\textbf{Split}&\multicolumn{2}{|c|}{\textbf{20NewsGroup}} &\multicolumn{2}{|c|}{\textbf{Cifar10}} &\multicolumn{2}{|c|}{\textbf{Cifar100}} &\multicolumn{2}{|c|}{\textbf{Pascal VOC}}\\
\cline{2-9} 
\textbf{} & \textbf{\textit{Positive}}& \textbf{\textit{Negative}}& \textbf{\textit{Positive}} & \textbf{\textit{Negative}} & \textbf{\textit{Positive}} & \textbf{\textit{Negative}} & \textbf{\textit{Positive}} & \textbf{\textit{Negative}}\\
\hline
Train & $30$ & $30$ & $500$ & $500$ & $500$ & $500$ & $124$ & $124$  \\
\hline
Test & $20$ & $20$ & $100$ & $100$ & $100$ & $100$ & $84$ & $84$ \\
\hline
\end{tabular}
\label{tab: bag_level_distribution}
}
\end{center}

\vspace{-15mm}
\end{table*}

\vspace{-2mm}
\subsection{Experimental Setup}\vspace{-1mm}

\vspace{1mm}\noindent{\bf Datasets.} Our experiments involve four datasets covering both textual and image data: 20NewGroup~\cite{Zhou2009}, Cifar10~\cite{Krizhevsky2009LearningML},  Cifar100~\cite{Krizhevsky2009LearningML}, and Pascal VOC \cite{Everingham15}.
The detailed description of each dataset is given below and bag level statistics is summarized in the Table \ref{tab: bag_level_distribution}

%For 20NewsGroup, the dataset is already available in the MIL setting, which consists of 20 topics where each topic contains 50 positive and 50 negative bags. For Cifar10 and Cifar100 datasets, bags are constructed by treating each image as an instance. For Cifar10, images corresponding to 'automobile', 'bird', and 'dog' are regarded as a positive instance otherwise negative. In case of Cifar100, images in superclass flowers are treated as positive and the rest as negative. In Pascal VOC, we perform image segmentation so each image is regarded as a bag and corresponding patches cropped from the image are treated as instances. In our experiments, images containing birds as a positive bags and others as negative. Table \ref{tab: bag_level_distribution}  summarizes the bag statistics.

\begin{itemize}[leftmargin=*]
    \item {\bf 20NewsGroup:}  In this dataset, an instance refers to a post from a particular topic. % represented by a 200 dimensional TF-IDF feature vector \citep{Zhou2009}. 
    For each topic, a bag is considered as positive if it contains at least one instance from that topic and negative otherwise. This dataset is particularly challenging because of the severe imbalance where there are very few ($\approx 3\%$) positive instances in each positive bag. While number of instances per bag may vary, on average there are around 40 instances per bag.
    \item {\bf Cifar10:} In the original dataset, there are 50,000 training and 10,000 testing images with 10 classes indicating different images. The bags are constructed as follows. First, we pick `automobile', `bird', and `dog' related images as positive instances and the rest as negative. To construct a positive bag, we choose a random number from 1 to 3 and pick the positive instances equal to the randomly generated number. The rest of the instances are selected from a negative instances pool. For negative bags, all instances are selected from the negative instance pool. For each bag, we consider 32 instances. %For training purpose, we construct bags from original training set whereas, for validation purpose we construct bags from original testing set. %The detailed bag level distribution is presented in the Table \ref{tab: bag_level_distribution}. 
    \item {\bf Cifar100:} The dataset consists of 50,000 training and 10,000 testing images with 20 different superclasses indicating different species. Bag construction is similar to Cifar10, where images in superclass flowers are treated as positive and the rest as negative.  
    
    \item {\bf Pascal VOC:} This dataset consists of 2,913 images, where images are used for segmentation. Each image is treated as a bag and instances are obtained as follows. We define a grid size of $60\times 75$ and partition the images. Depending on the image size, the number of instances may vary. We treat an instance as positive if at least $5\%$ of the total pixels in a given instance are related to the object of interest otherwise negative. In our case, we considerthe  bird as the object of interest. All the images consisting of bird are regarded as positive bags and others as negative. %We randomly split positive bags into 60:40 ratio to yield training and validation set. We select the same number of negative bags that is equal to positive bags. The detailed bag level statistics is presented in the Table \ref{tab: bag_level_distribution}.
\end{itemize}

\begin{figure*}[t!]
\centering
\begin{subfigure}{0.24\textwidth}
  \centering
  \includegraphics[width=1.0\linewidth]{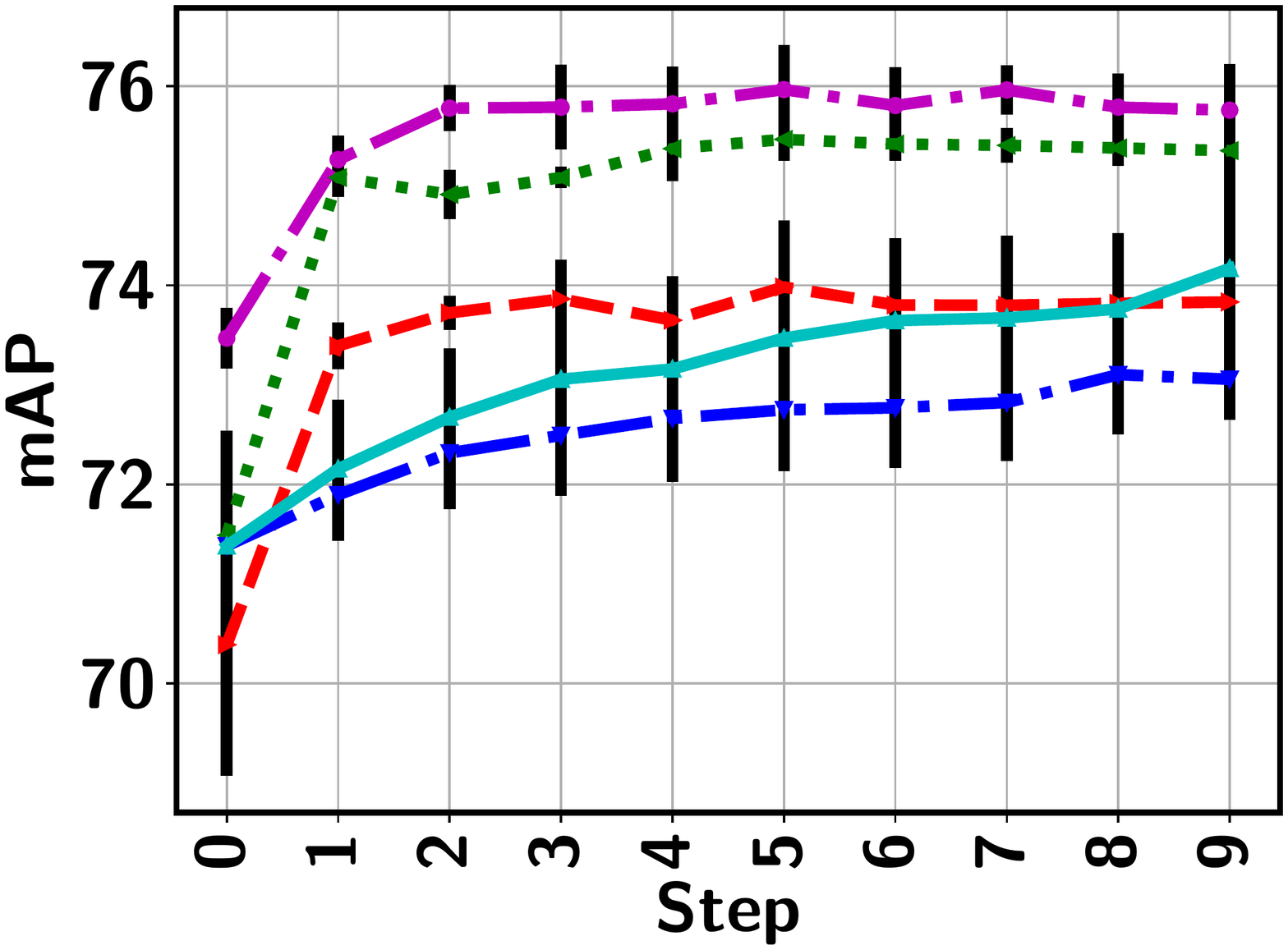}
  \vspace{-18mm}
  \caption{20NewsGroup}
  
\end{subfigure}
\begin{subfigure}{0.24\textwidth}
  \centering
  \includegraphics[width=1.0\linewidth]{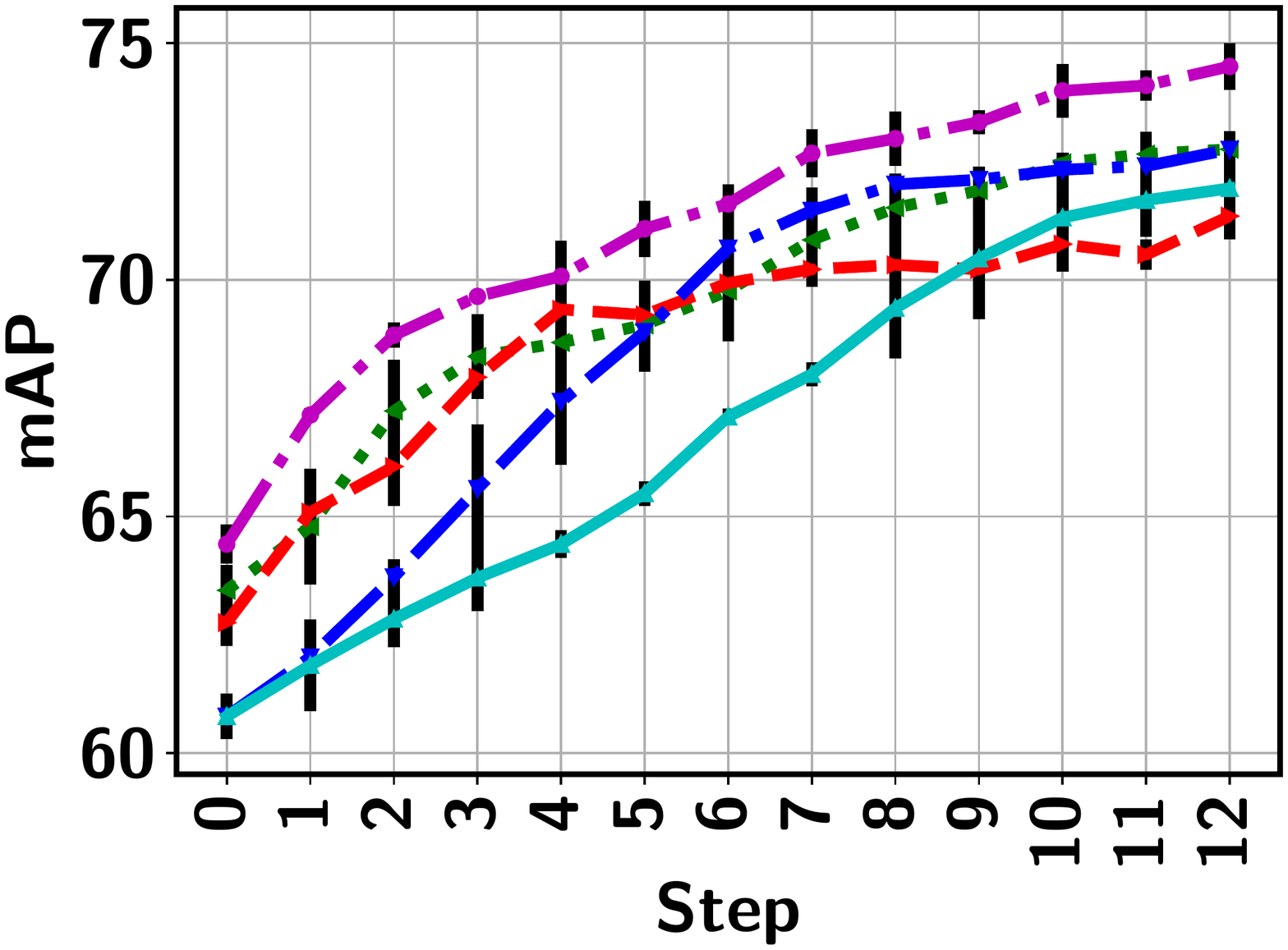}
  \vspace{-18mm}
  \caption{Cifar10}
\end{subfigure}
\begin{subfigure}{0.24\textwidth}
  \centering
  \includegraphics[width=1.0\linewidth]{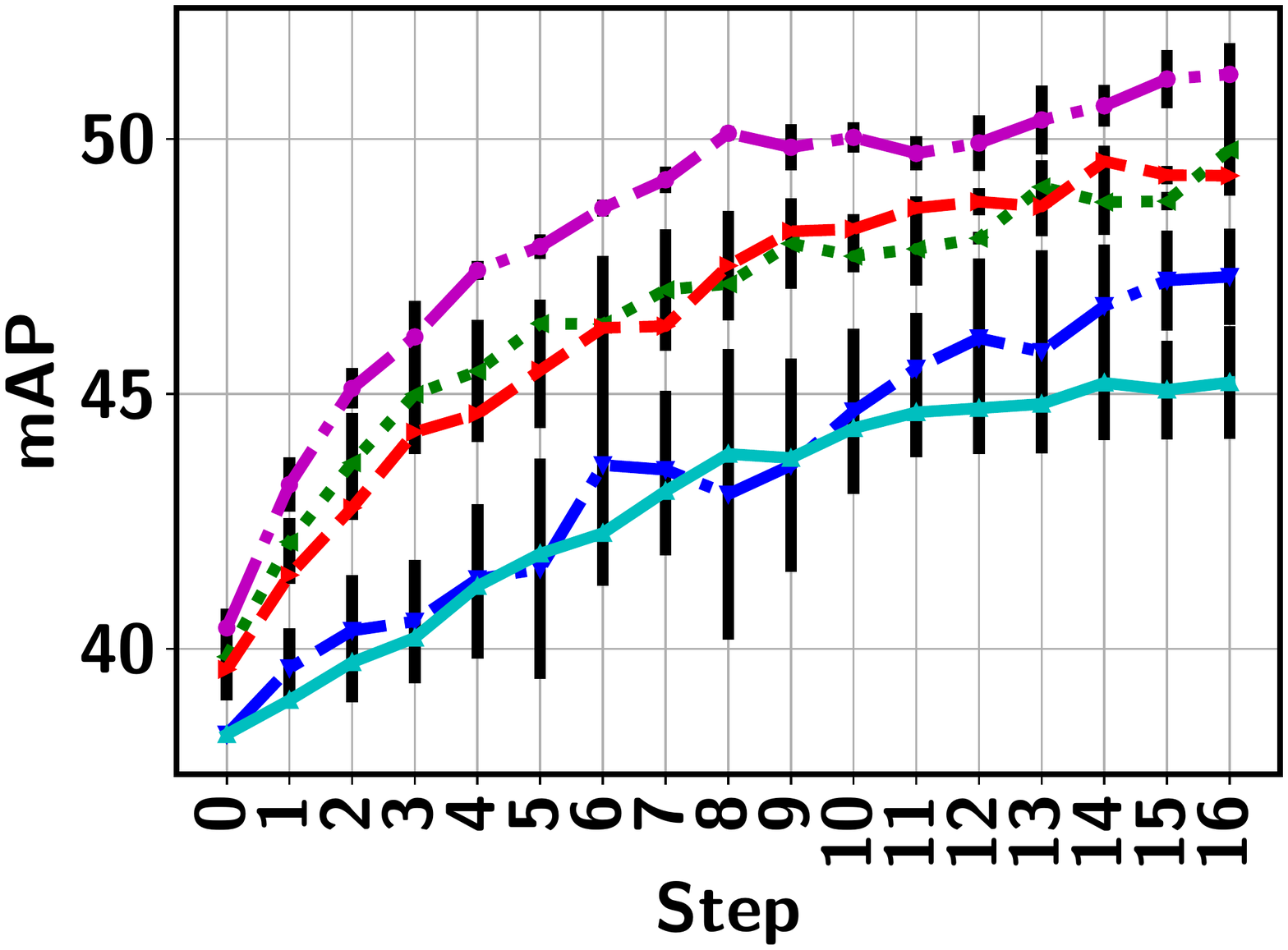}
  \vspace{-18mm}
  \caption{Cifar100}
\end{subfigure}%
\begin{subfigure}{0.24\textwidth}
  \centering
  \includegraphics[width=1.0\linewidth]{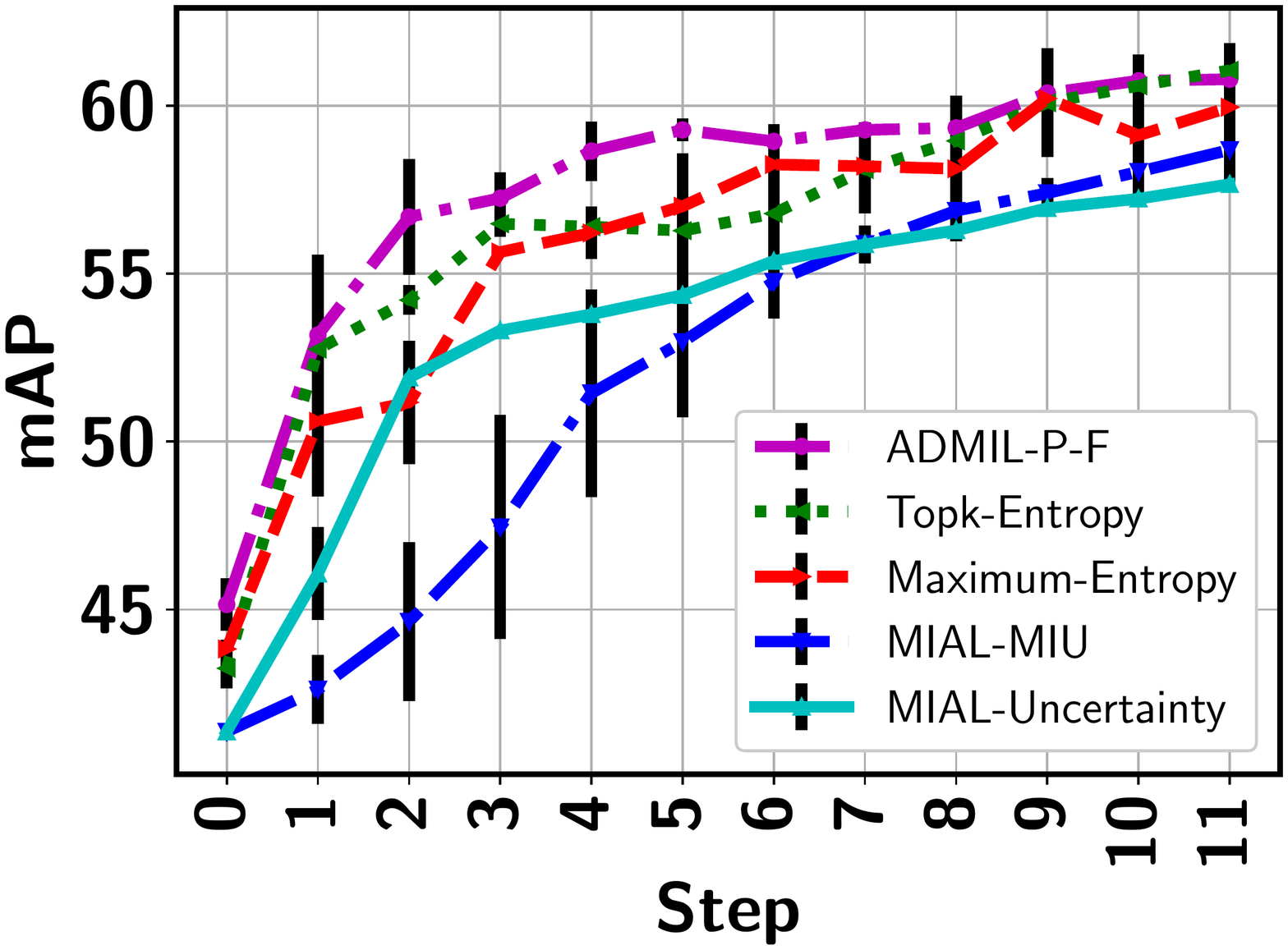}
  \vspace{-18mm}
  \caption{Pascal VOC}
\end{subfigure}
\vspace{-3mm}
\caption{MI-AL performance}
\label{fig: al_curve_comparison_mean_sd}
\vspace{-14mm}
\end{figure*} 

\begin{figure*}[t!]
\centering
\begin{subfigure}{0.24\textwidth}
  \centering
  \includegraphics[width=1.0\linewidth]{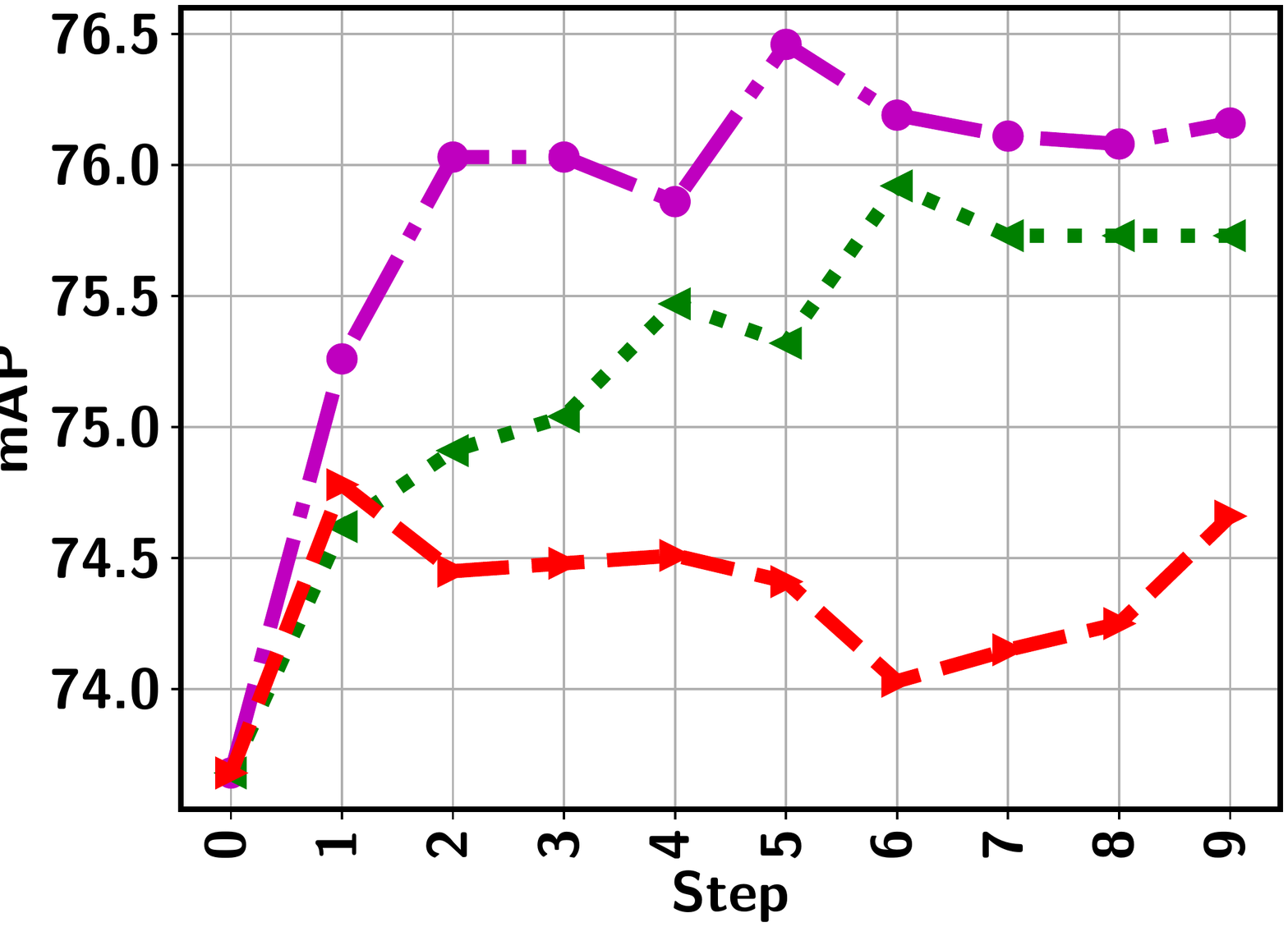}
  \vspace{-18mm}
  \caption{20NewsGroup}
  
\end{subfigure}
\begin{subfigure}{0.24\textwidth}
  \centering
  \includegraphics[width=1.0\linewidth]{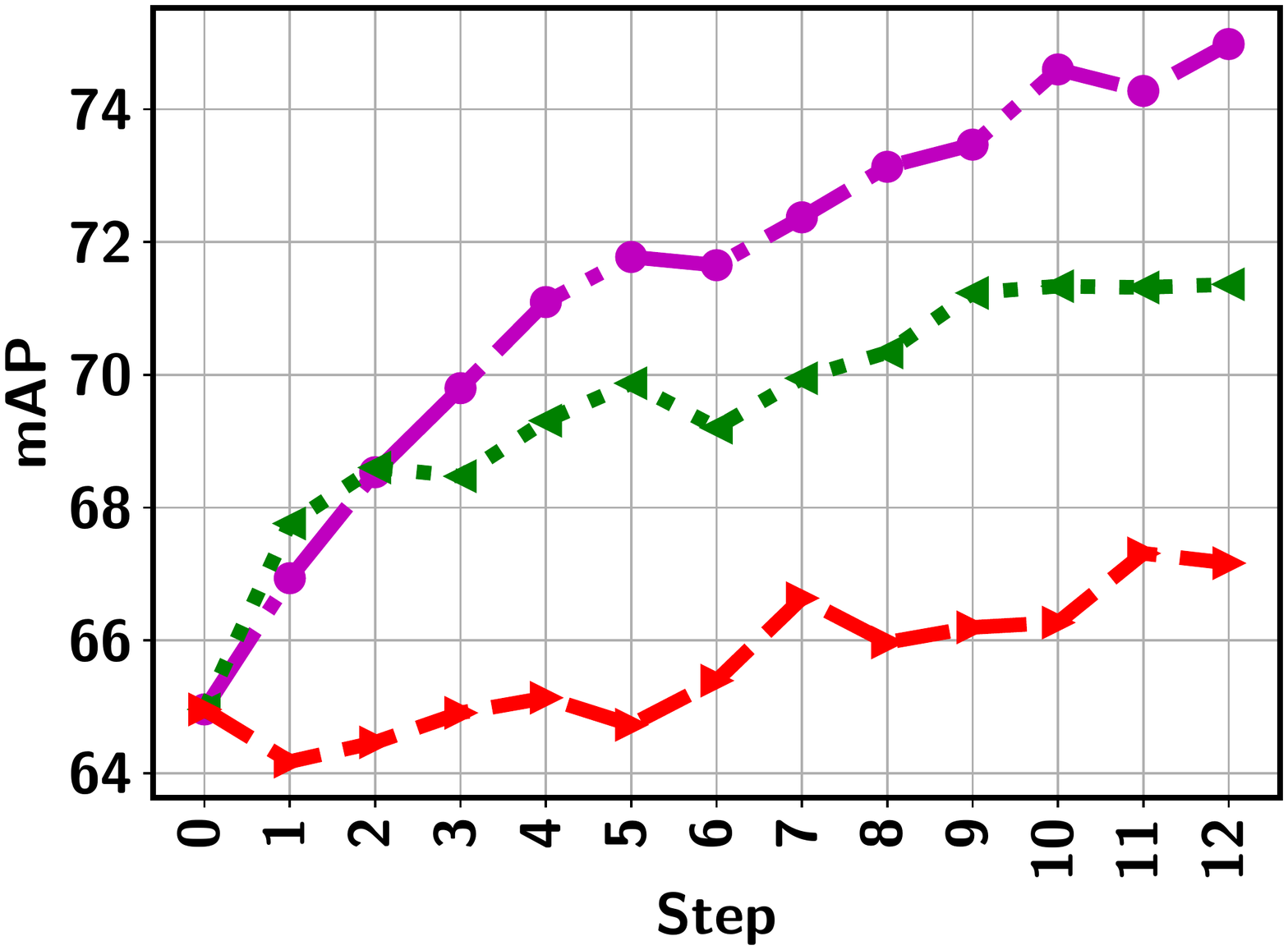}
  \vspace{-18mm}
  \caption{Cifar10}
\end{subfigure}
\begin{subfigure}{0.24\textwidth}
  \centering
  \includegraphics[width=1.0\linewidth]{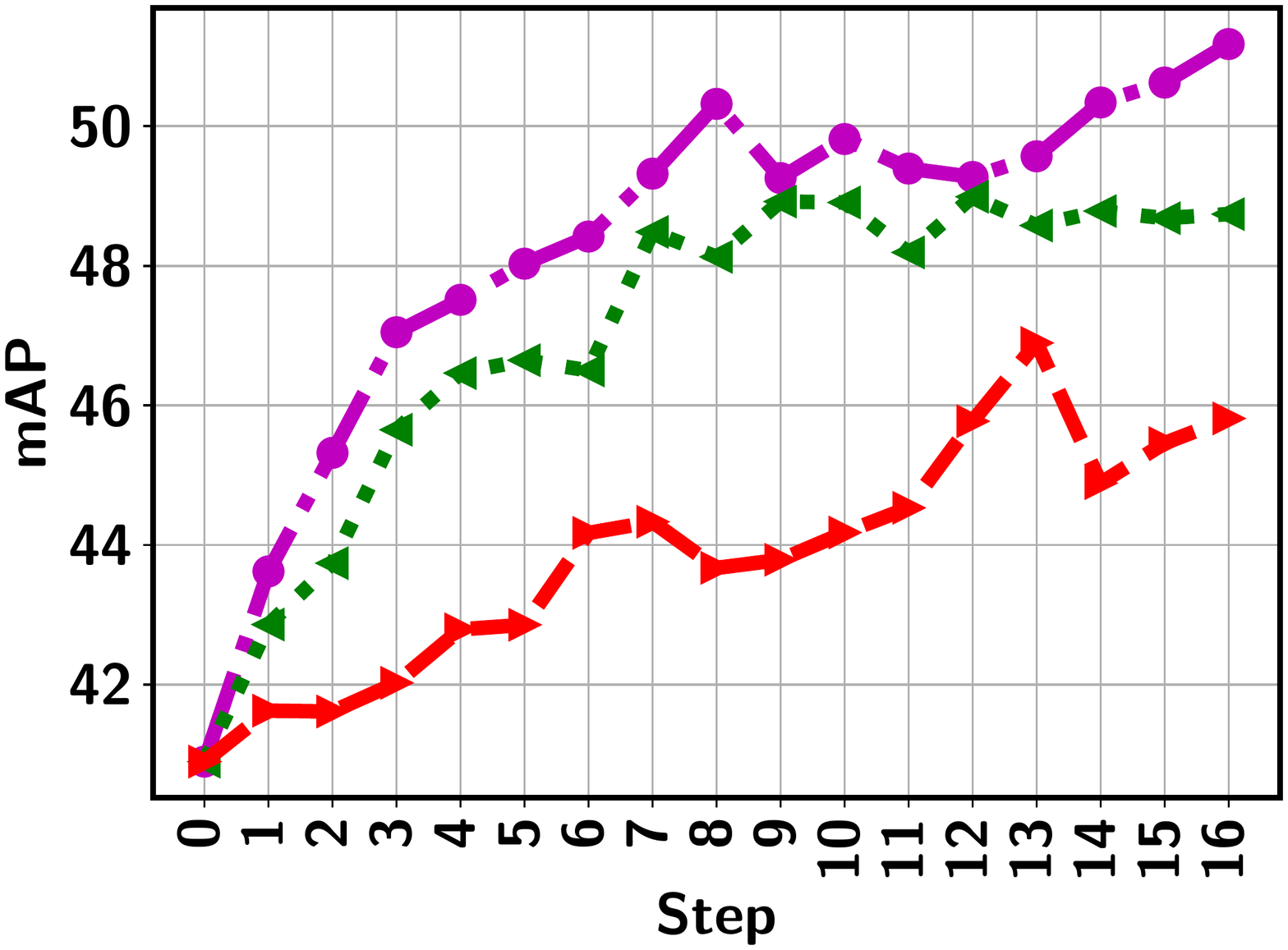}
  \vspace{-18mm}
  \caption{Cifar100}
\end{subfigure}%
\begin{subfigure}{0.24\textwidth}
  \centering
  \includegraphics[width=1.0\linewidth]{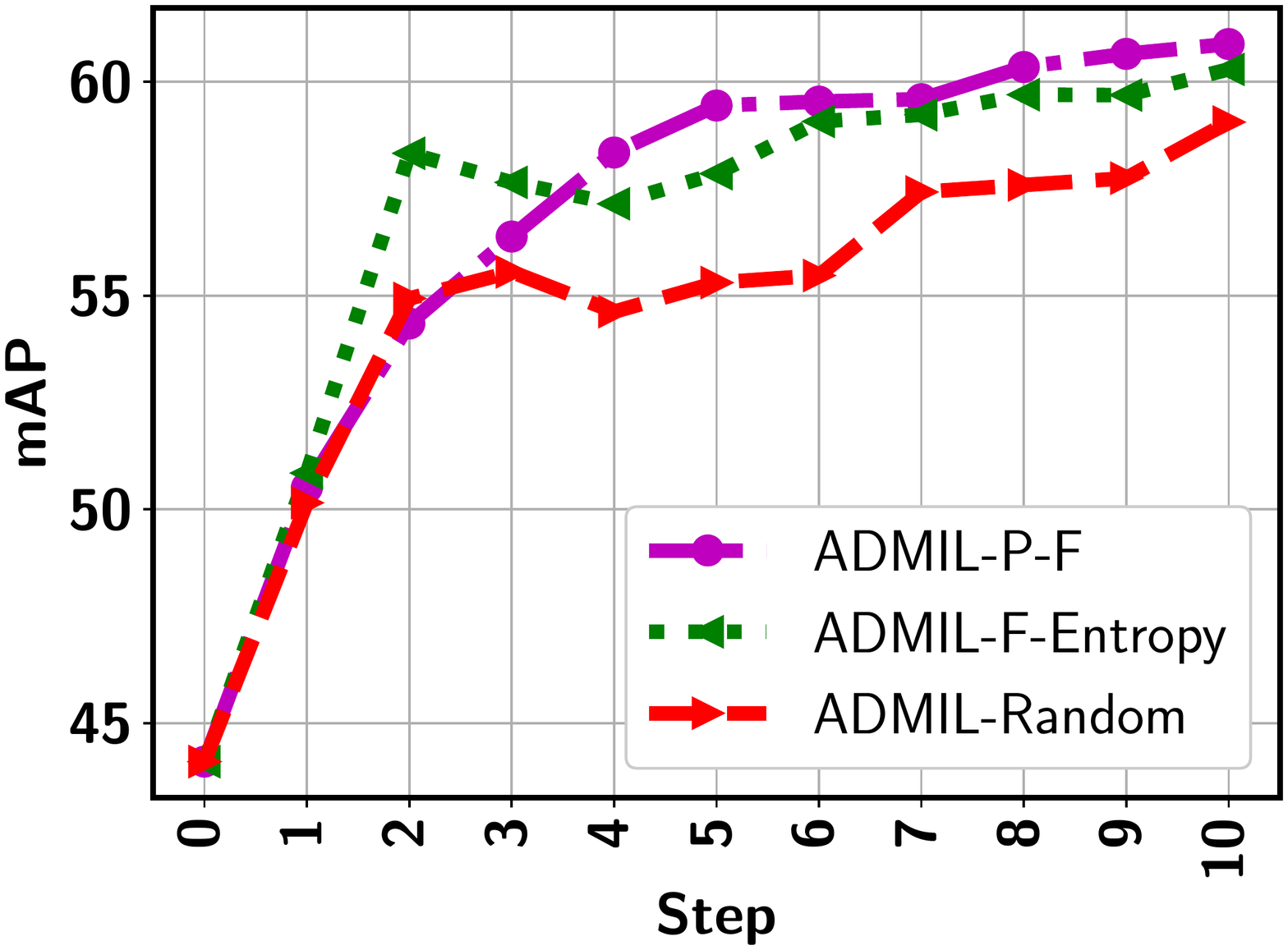}
  \vspace{-18mm}
  \caption{Pascal VOC}
\end{subfigure}
\vspace{-3mm}
\caption{Effectiveness of P-F active sampling}
\label{fig: al_curve_ablation}
\vspace{-4mm}
\end{figure*} 
%\vspace{-2mm}
\noindent{\bf Evaluation metric and model training.}
To assess the model performance, we report the instance-level mean average precision (mAP) score, which summarizes a precision-recall curve as a weighted mean of precision achieved at each threshold, with the increase in recall from the previous threshold as the weight. mAP explicitly places much stronger emphasis on the correctness of the few top ranked instances than other metrics (\eg AUC)~\cite{su2015relationship}. This makes it particularly suitable for instance prediction evaluation as a small subset of instances with the highest prediction scores will eventually be identified as positive for further inspection (by human experts) with the rest being ignored. For Cifar10, Cifar100, and Pascal VOC datasets, we extract the visual features from the second-to-the last layer of a VGG16 network pre-trained using the imagenet dataset, yielding a 4,096 dimensional feature vector for each instance. For 20NewsGroup, we use the available 200-dimensional feature vector. In terms of network architecture, we use a 3-layer FC neural network. The first layer has 32 units followed by 16 units and 1 unit FC layers. We adopt $60\%$ dropout between FC layers. ReLU and sigmoid activations are used for the first and last FC layers. Learning rate 0.01 is used for all dataset except for 20NewsGroup which is 0.1.

\vspace{-2mm}
\subsection{Performance Comparison}%\vspace{-1mm}
%{\bf Performance comparison.} 
To demonstrate the instance prediction performance achieved by the proposed ADMIL model, we compare it with competitive baselines. First, the two MI-AL sampling strategies: MIAL-Uncertainty and MIAL-MIU \cite{settles2007multiple}, from the MI logistic model are included. Since our datasets involve high-dimensional data, we replace the original linear model by the exact DNN model used in our ADMIL so we can focus on comparing MI active sampling. The EGL sampling technique in \cite{settles2007multiple} was not included due to the prohibitive computational cost to evaluate the gradient of each instance output with respect to the large number of DNN parameters. We also implement an MS-MIL model and its top-$k$ variant with uncertainty sampling using entropy. Given the different sizes of the datasets, we query maximum 15 instances per step in 20NewsGroup, 30 instances in Pascal VOC, and 150 instances in Cifar10 and Cifar100. Figure~\ref{fig: al_curve_comparison_mean_sd} shows the MI-AL curves with one standard deviation (computed over three runs) represented by vertical black line for all four datasets. ADMIL achieves the best performance in all cases. For most datasets, it shows a much better initial performance, which results from the proposed DRBL-based MIL loss that significantly benefits MIL performance in passive learning. Overall the entire MI-AL process, ADMIL consistently stays the best and converges to a higher point in the end for all datasets. For the Pascal VOC, the top-$k$ MIL model with entropy sampling achieves closer performance towards the end, which is mainly due to the limited positive instances in this dataset. Hence, no testing bags contain similar positive instances in the challenging bags that are explored by P-F sampling. While ADMIL achieves much better instance predictions in those bags, the advantage does not transfer to the testing bags. For reference, we also compare ADMIL with two recently developed MIL models, including Ilse et al. \cite{Ilse2018} and  Hsu et al. \cite{Hsu2020QueryDrivenML}, under the passive setting. 
As shown in Table~\ref{tab: dro_passive_performance}, ADMIL achieves better or at least comparable performance as compared with these competitive baselines. 
%comparable (better) to the state-of-the-art techniques which 
This clearly justifies of using ADMIL as a base model for active sampling. After labeling a small set of actively sampled instances, the performance is significantly boosted (as shown in the parenthesis), which further justifies the benefits of combining AL with MIL. Our qualitative study will provide a more detailed analysis on this.

\begin{table}[t!]
    \caption{MIL Performance in Passive Setting}
\vspace{-4mm}   
%\footnotesize
\begin{center}
\resizebox{1\columnwidth}{!}{%
\begin{tabular}{|c|c|c|c|c|c|c|c|c|}
\hline
\textbf{Approach} & {\textbf{20NewsGroup}} & {\textbf{Cifar10}} & {\textbf{Cifar100}} & {\textbf{Pascal VOC}}\\
\hline
Ilse et al. \cite{Ilse2018} & $60.85$ & $65.16$ & $40.15$ & $40.15$  \\
\hline
Hsu et al. \cite{Hsu2020QueryDrivenML} & $42.08$ & $63.84$ & $41.57$ & $34.83$ \\
\hline
ADMIL & $73.47 ({\bf 75.42})$ & $64.41 ({\bf 74.50})$ & $40.41 ({\bf 51.26})$ & $45.15 ({\bf 60.79})$ \\
\hline
\end{tabular}}
\label{tab: dro_passive_performance}
\end{center}
\vspace{-6mm}
\end{table}

\begin{figure*}[t!]
\vspace{-5mm}
\centering
\begin{subfigure}{0.24\textwidth}
  \centering
  \includegraphics[width=1.0\linewidth]{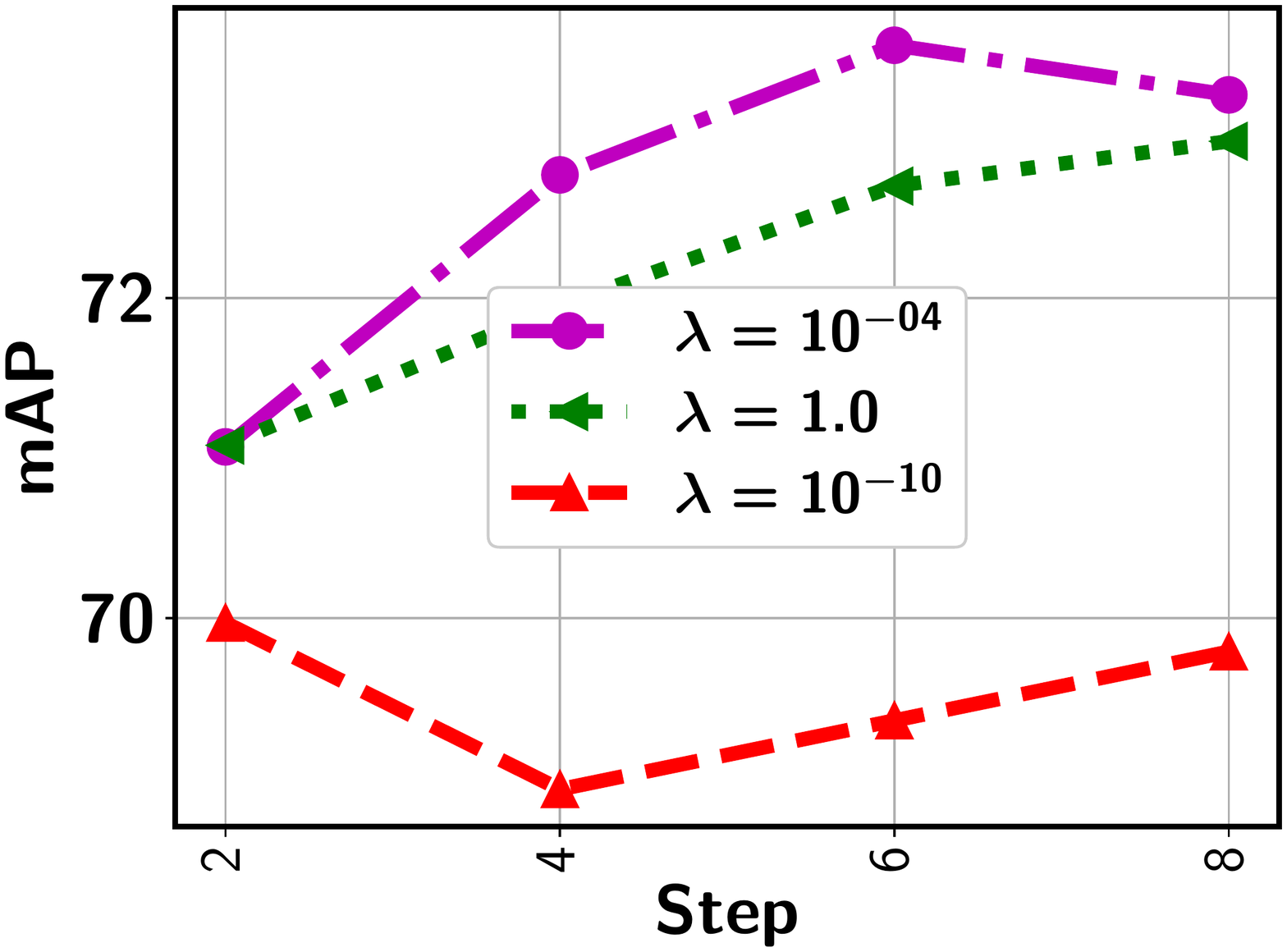}
  \vspace{-18mm}
  \caption{20NewsGroup}
\end{subfigure}%
\begin{subfigure}{0.24\textwidth}
  \centering
  \includegraphics[width=1.0\linewidth]{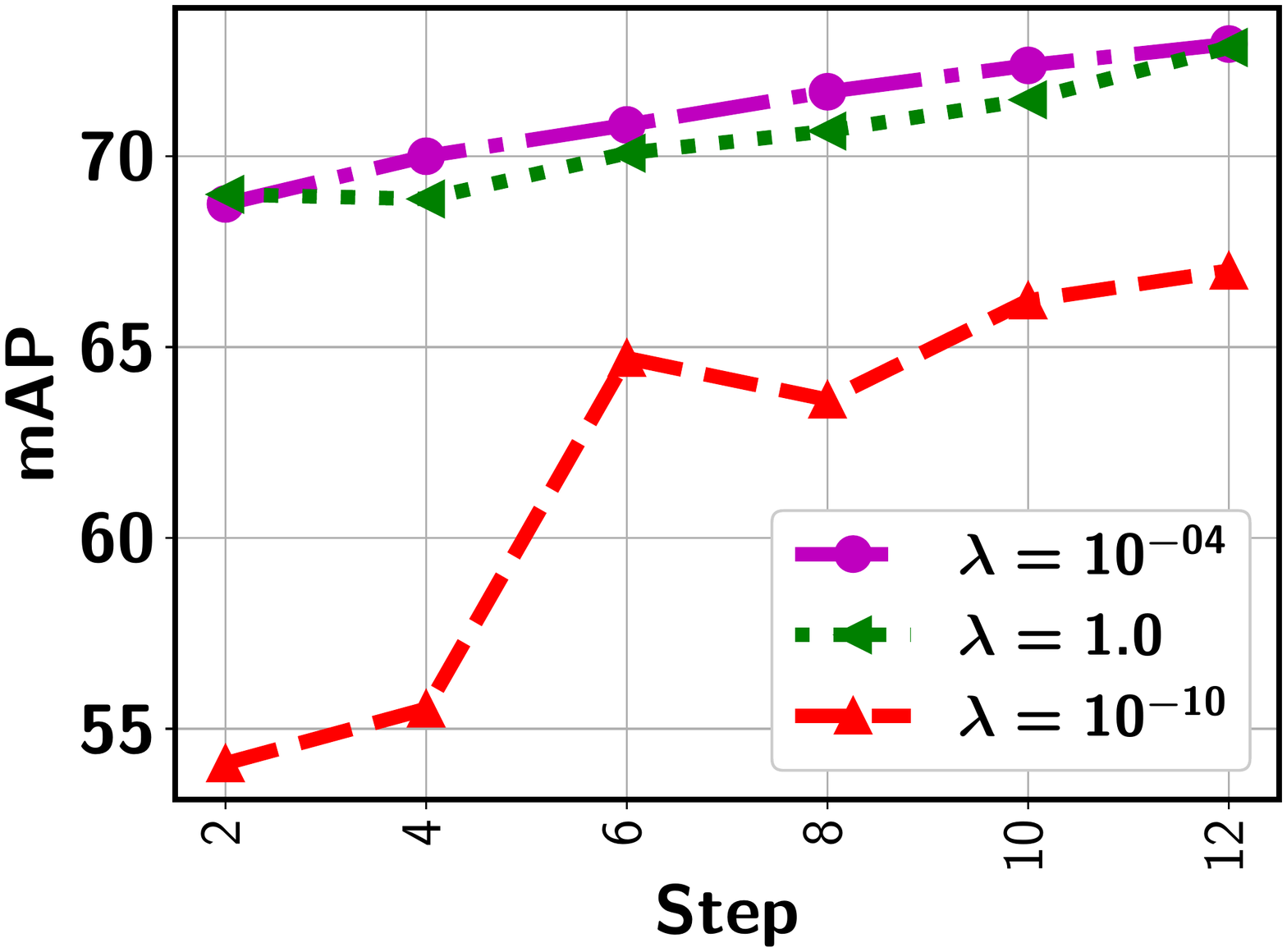}
\vspace{-18mm}
  \caption{Cifar10}
\end{subfigure}%
\begin{subfigure}{0.24\textwidth}
  \centering
  \includegraphics[width=1.0\linewidth]{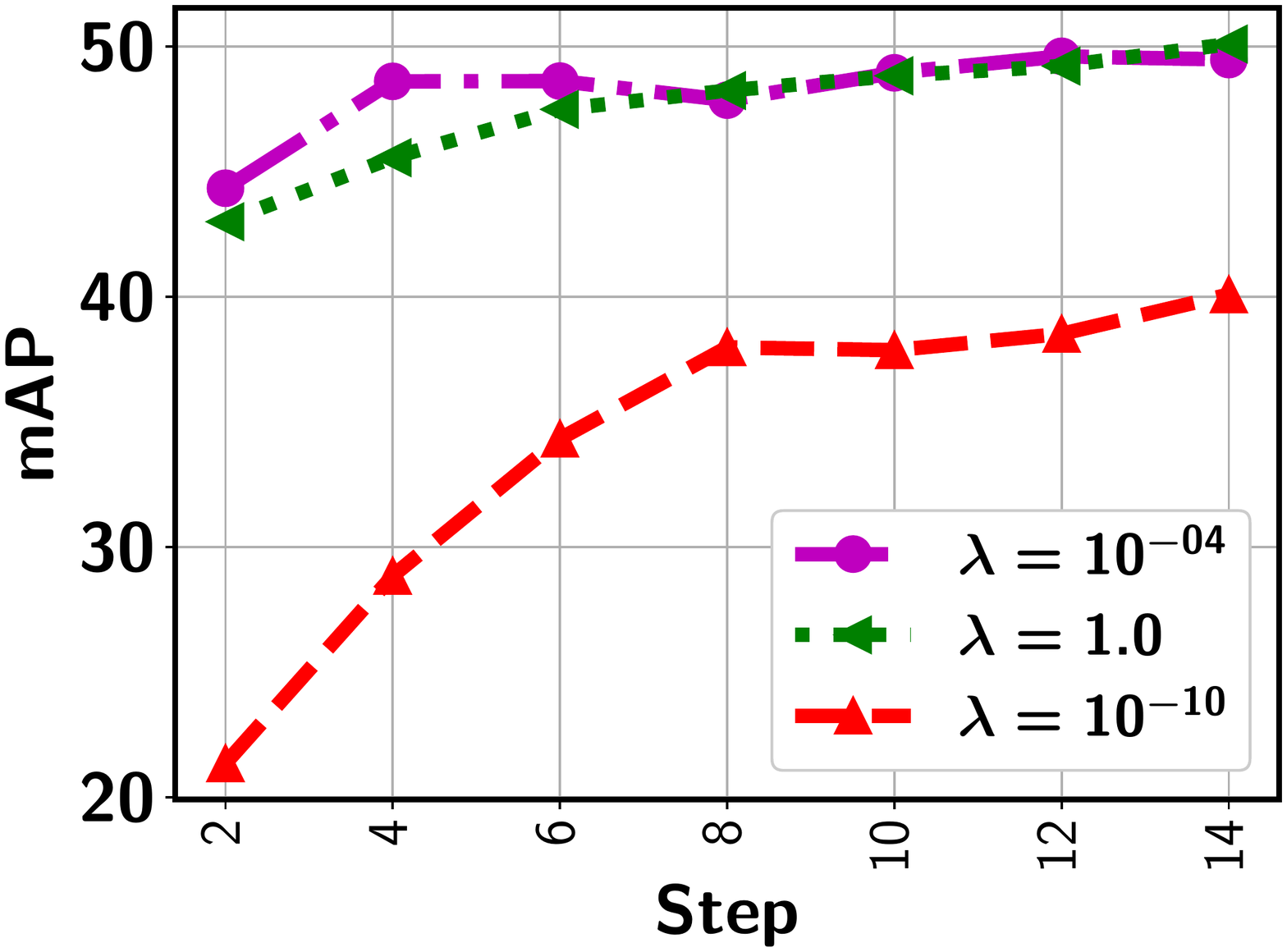}
  \vspace{-18mm}
  \caption{Cifar100}
\end{subfigure}%
\begin{subfigure}{0.24\textwidth}
  \centering
  \includegraphics[width=1.0\linewidth]{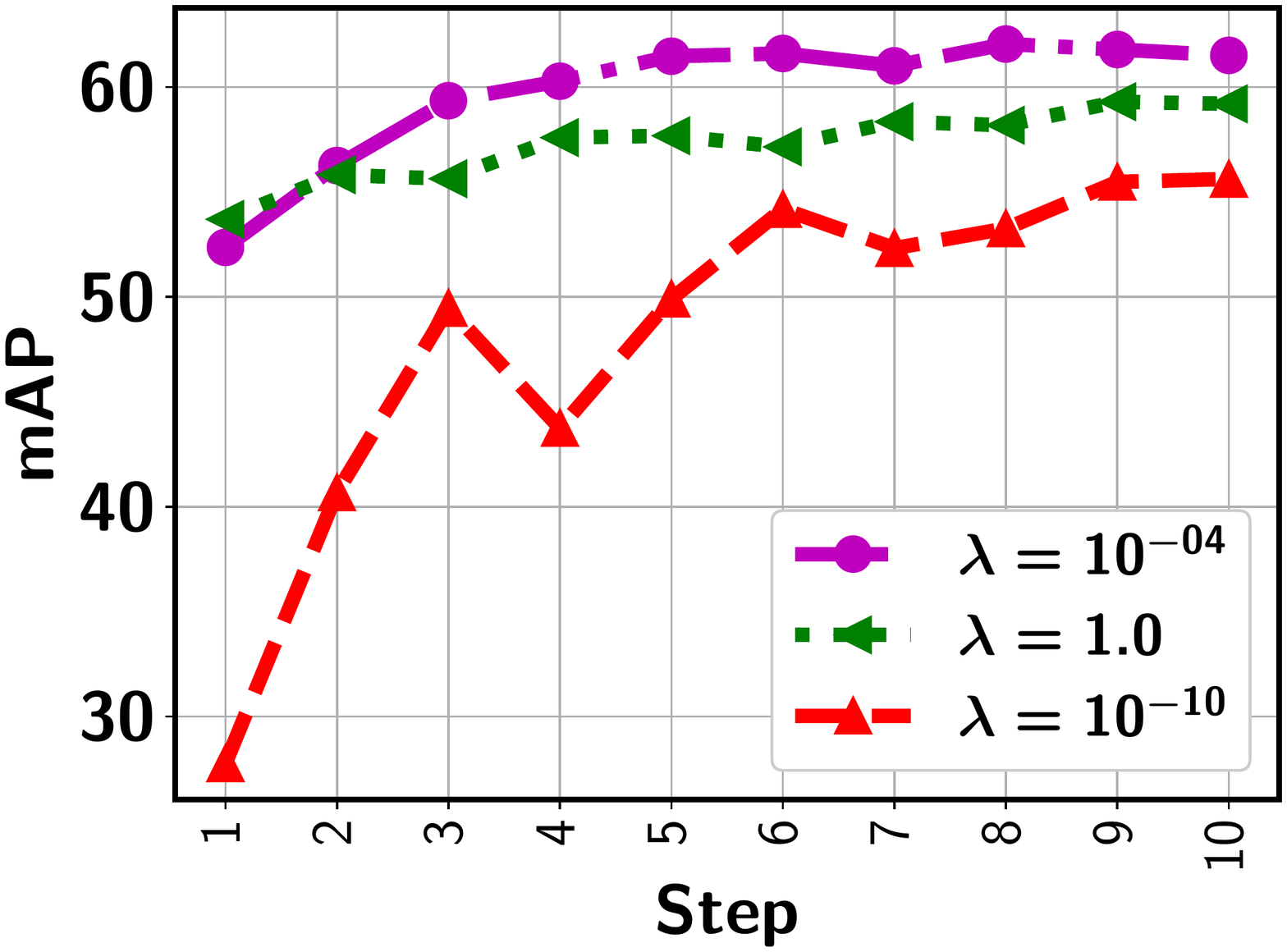}
\vspace{-18mm}
  \caption{Pascal VOC}
\end{subfigure}%
\vspace{-4mm}
\caption{Impact of model parameter $\lambda$}
\label{fig: ablation_lambda}
\vspace{-14mm}
\end{figure*}

\begin{figure*}[t!]
\centering
\begin{subfigure}{0.24\textwidth}
  \centering
  \includegraphics[width=1.0\linewidth]{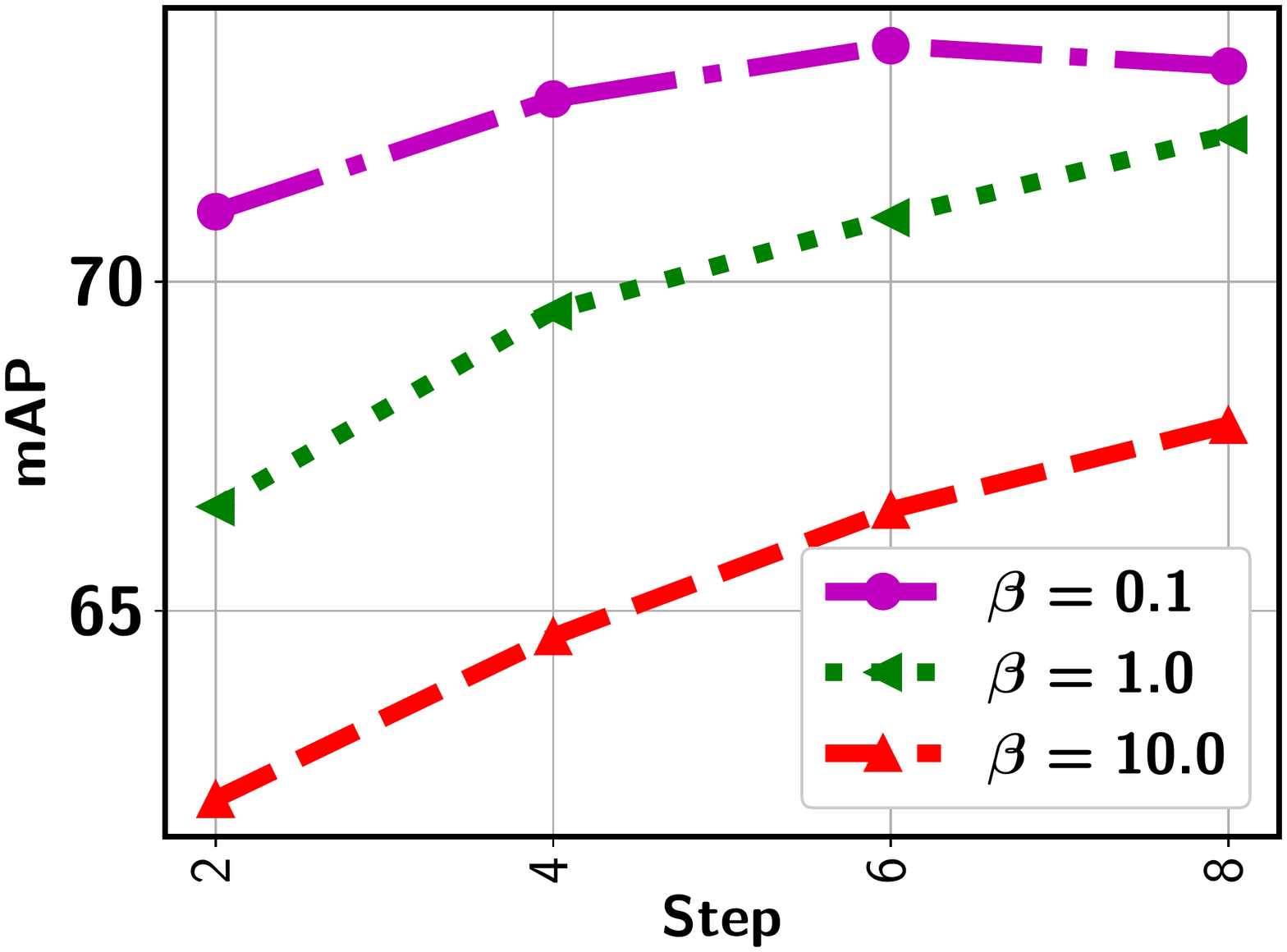}
   \vspace{-18mm}
  \caption{20NewsGroup }
\end{subfigure}%
\begin{subfigure}{0.24\textwidth}
  \centering
  \includegraphics[width=1.0\linewidth]{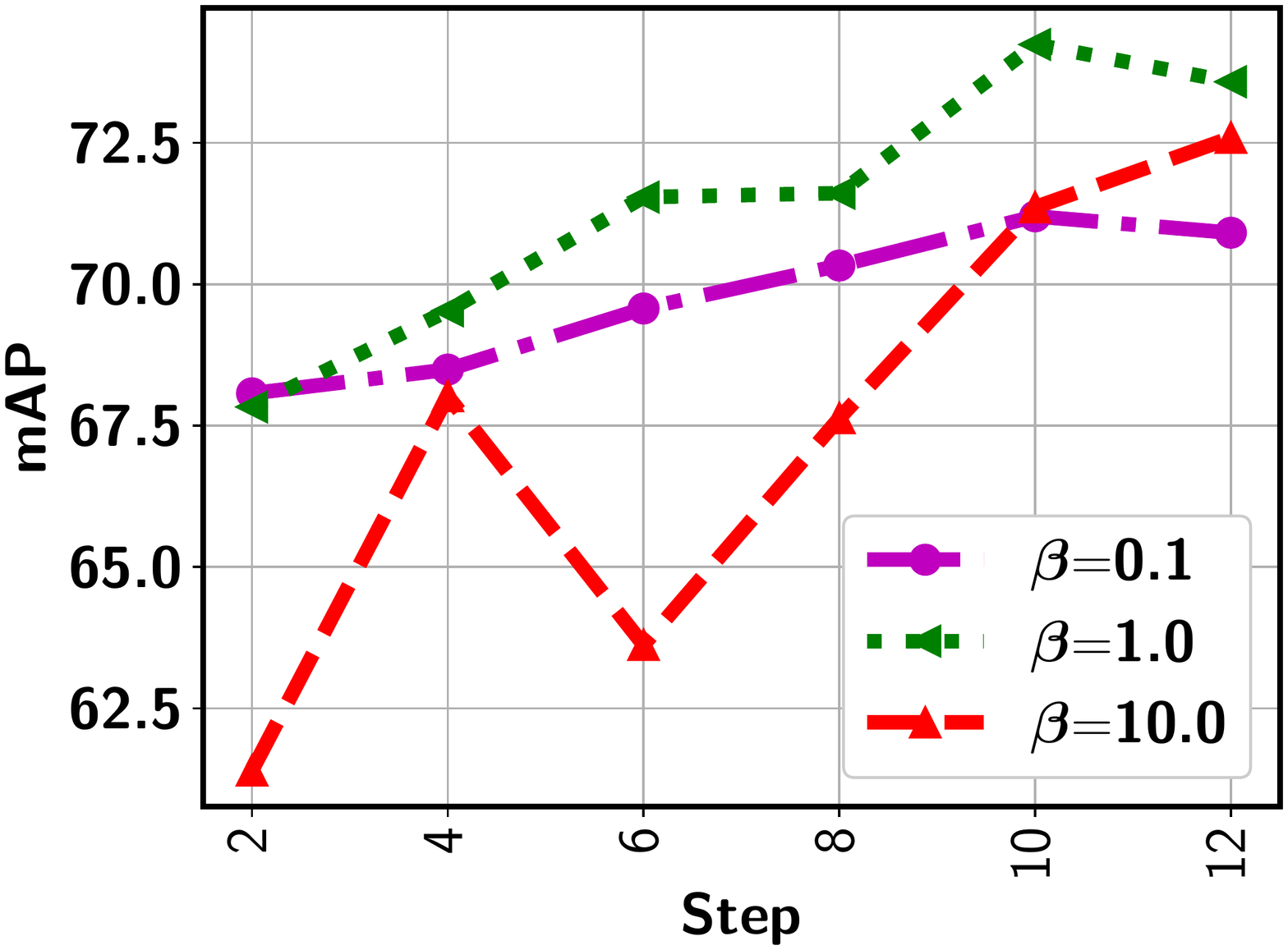}
  \vspace{-18mm}
  \caption{Cifar10}
\end{subfigure}%
\begin{subfigure}{0.24\textwidth}
  \centering
  \includegraphics[width=1.0\linewidth]{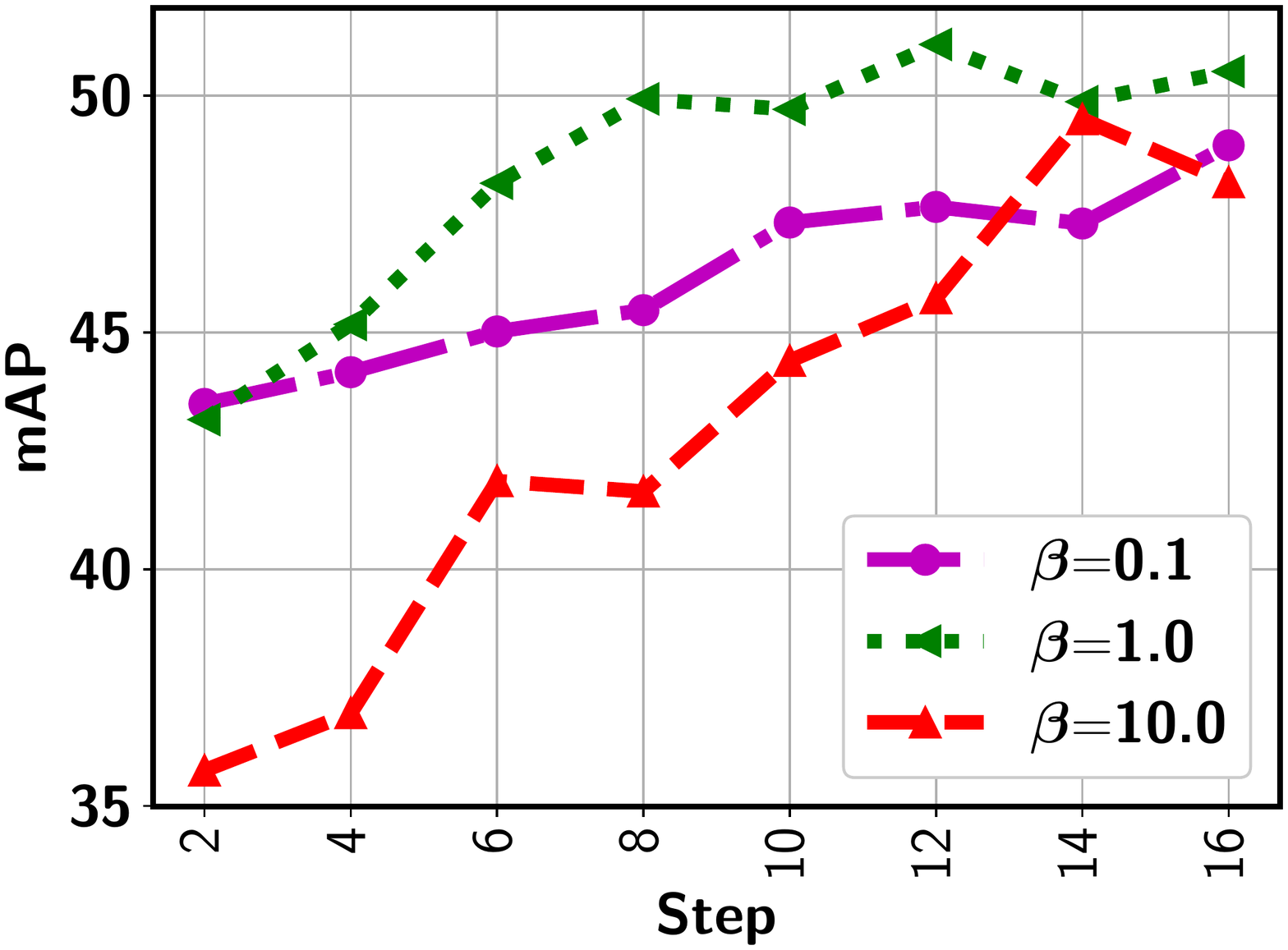}
   \vspace{-18mm}
  \caption{Cifar100}
\end{subfigure}%
\begin{subfigure}{0.24\textwidth}
  \centering
  \includegraphics[width=1.0\linewidth]{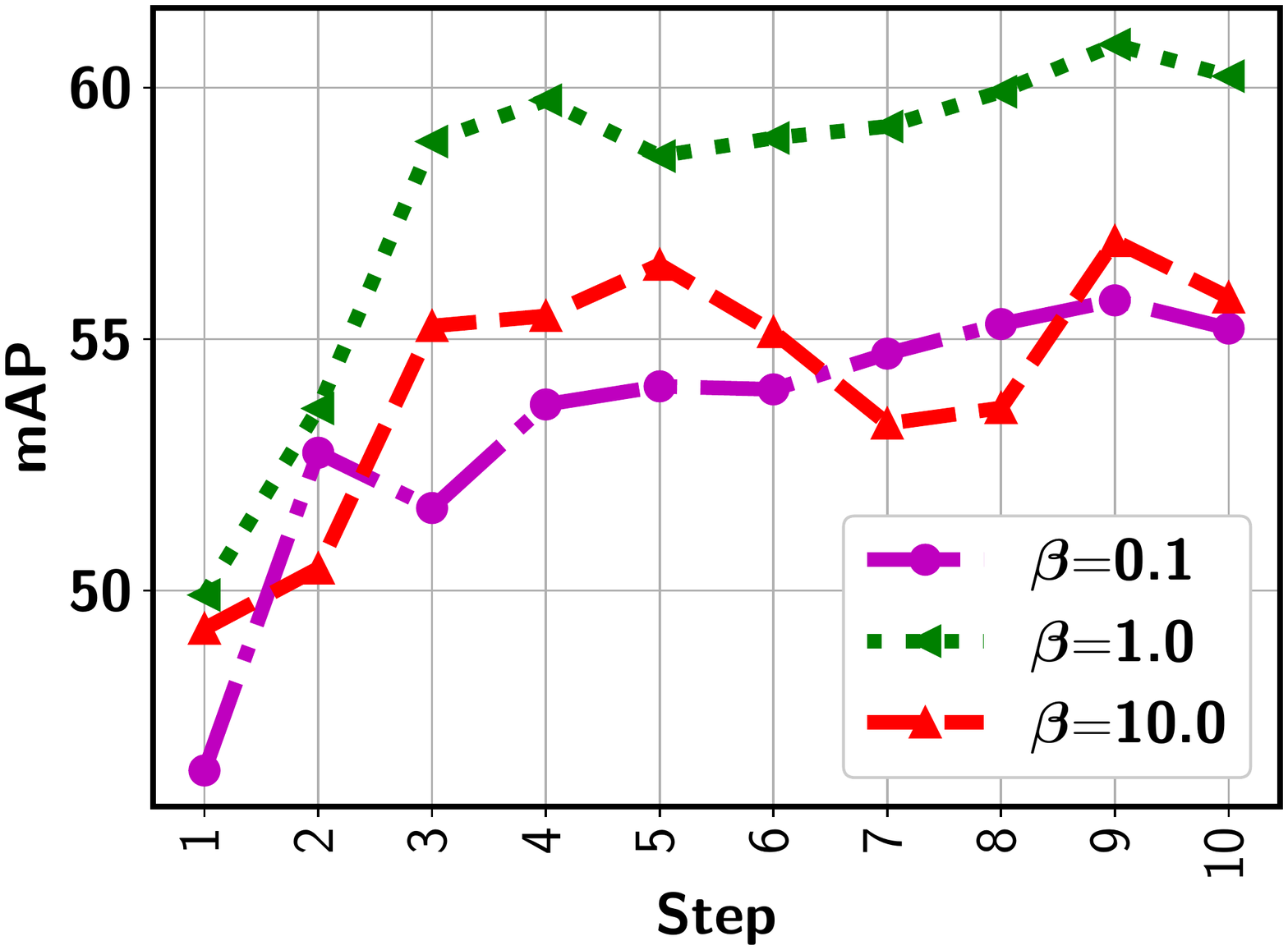}
  \vspace{-18mm}
  \caption{Pascal VOC}
\end{subfigure}%
\vspace{-4mm}
\caption{Impact of model parameter $\beta$}
\label{fig: ablation_beta}
\vspace{-14mm}
\end{figure*} 

\begin{figure*}[t!]
\centering
\begin{subfigure}{0.24\textwidth}
  \centering
  \includegraphics[width=1.0\linewidth]{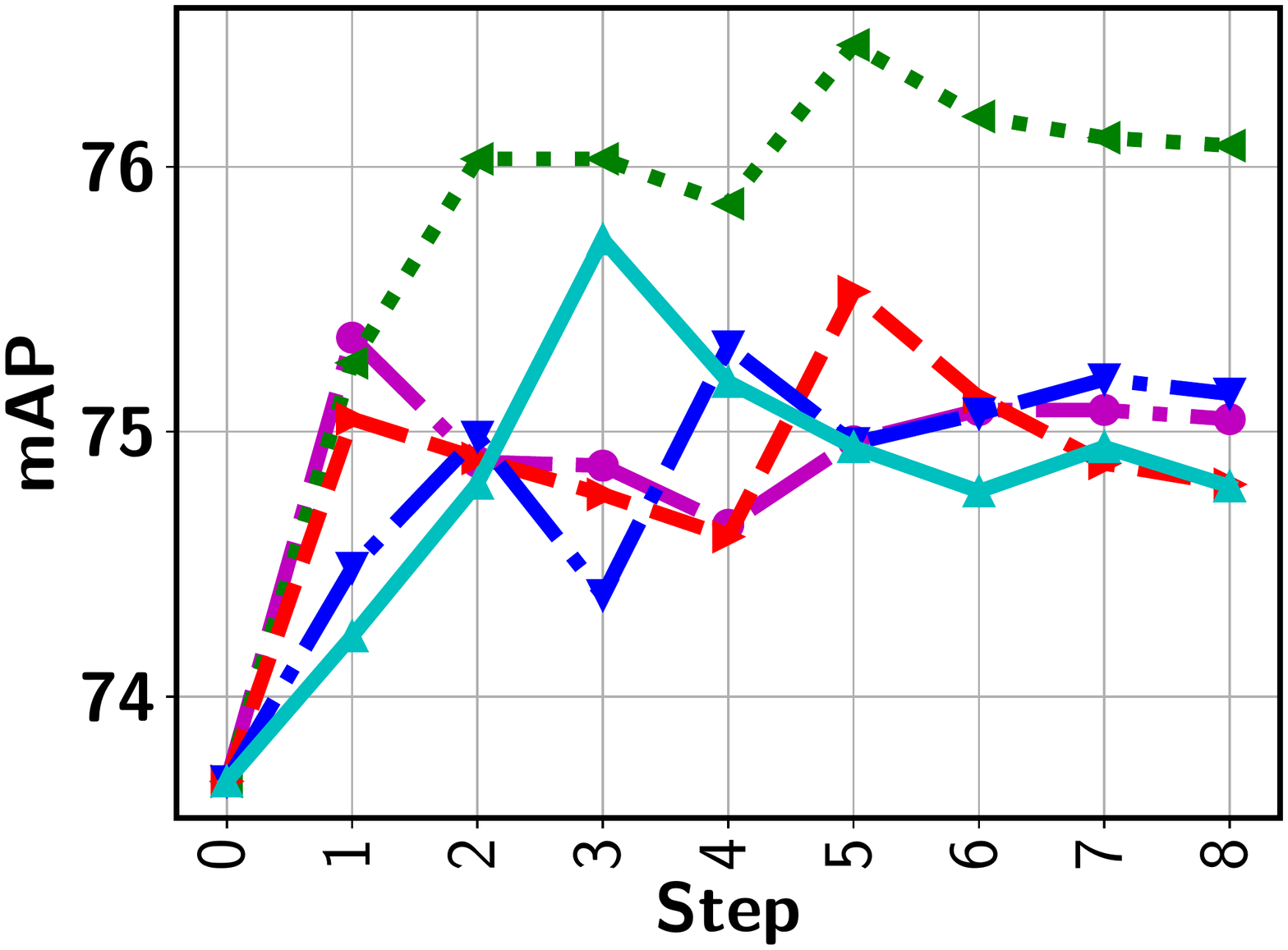}
  \vspace{-18mm}
  \caption{20NewsGroup}
  
\end{subfigure}
\begin{subfigure}{0.24\textwidth}
  \centering
  \includegraphics[width=1.0\linewidth]{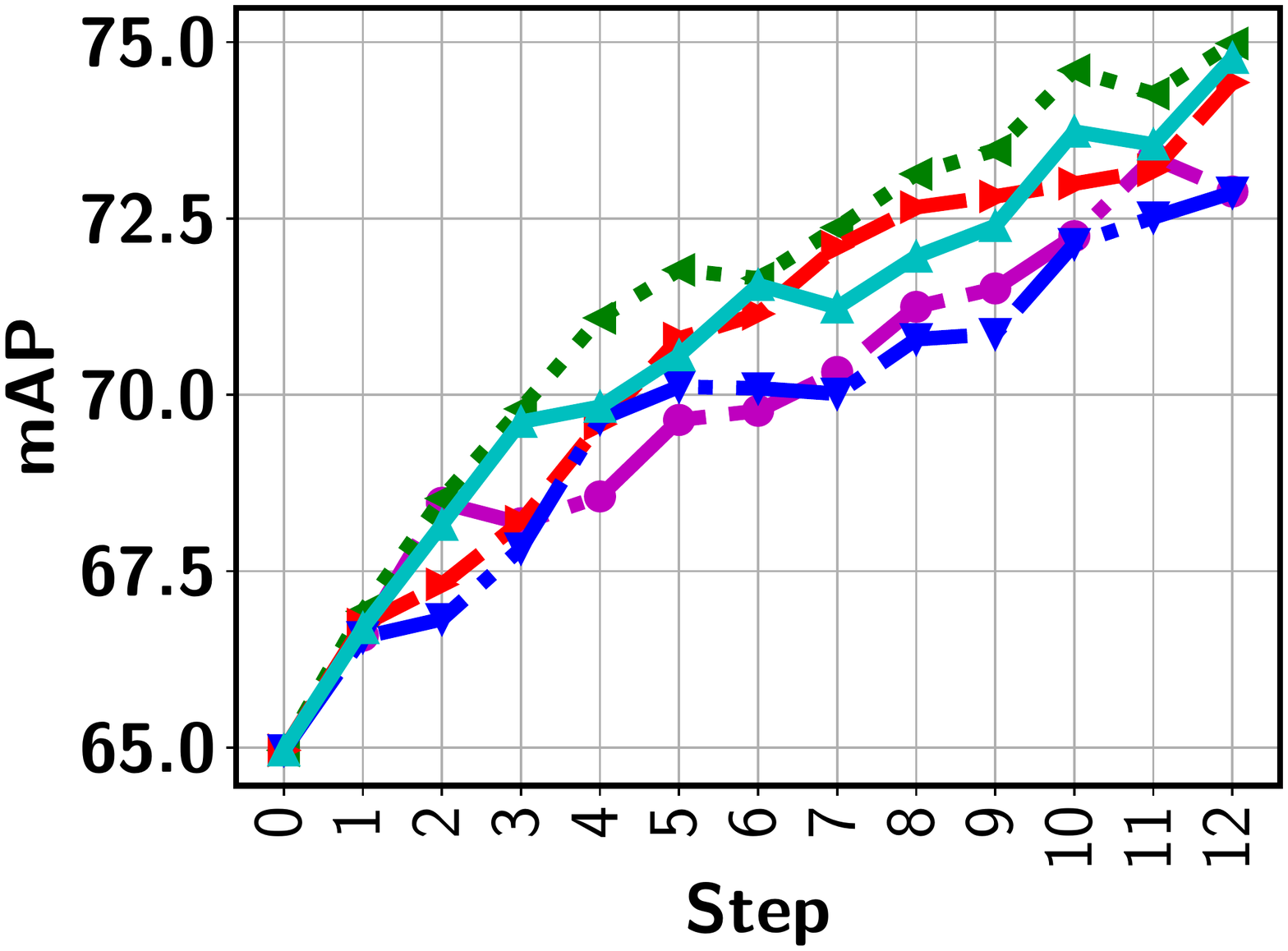}
  \vspace{-18mm}
  \caption{Cifar10}
\end{subfigure}
\begin{subfigure}{0.24\textwidth}
  \centering
  \includegraphics[width=1.0\linewidth]{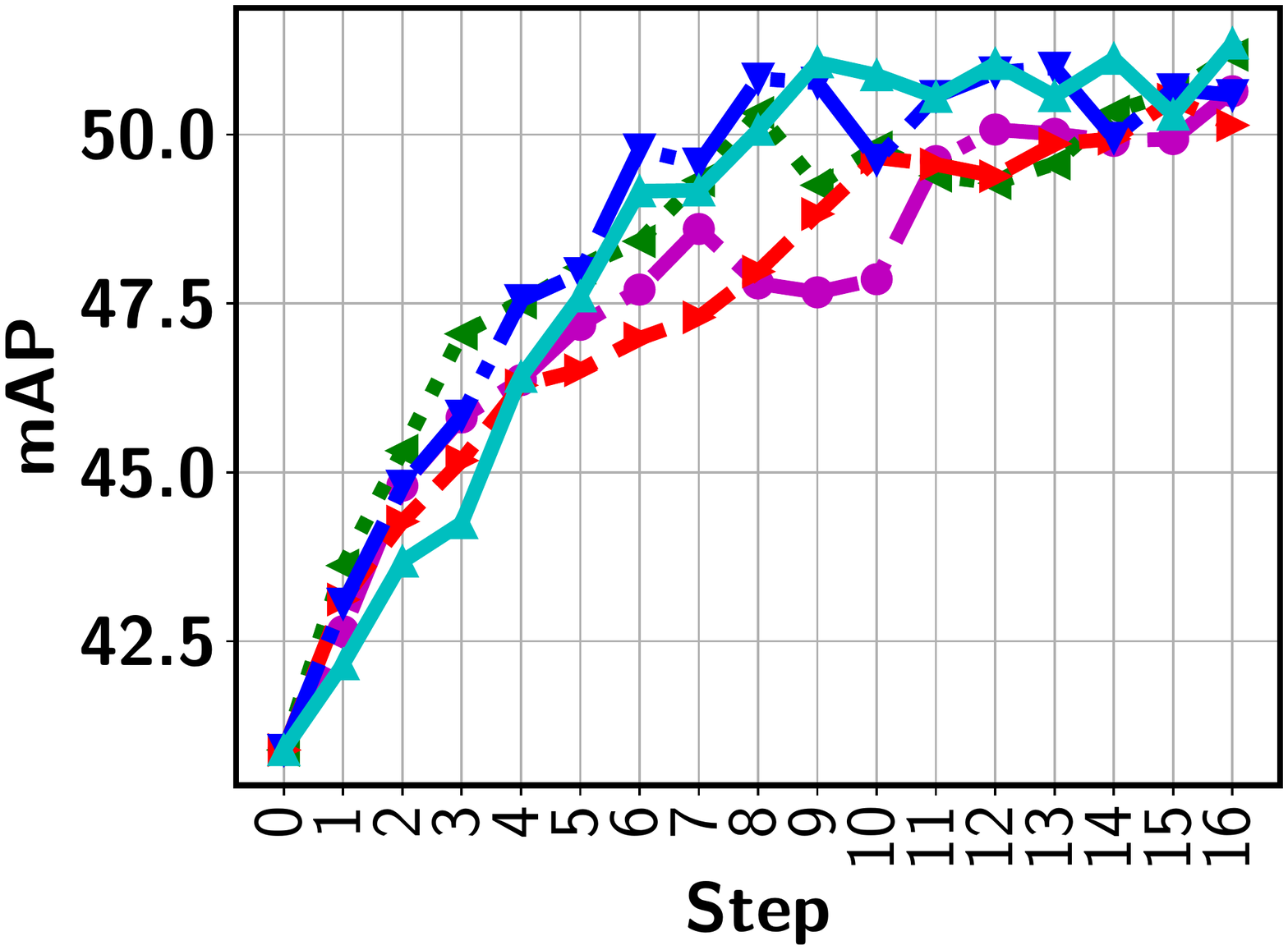}
  \vspace{-18mm}
  \caption{Cifar100}
\end{subfigure}%
\begin{subfigure}{0.24\textwidth}
  \centering
  \includegraphics[width=1.0\linewidth]{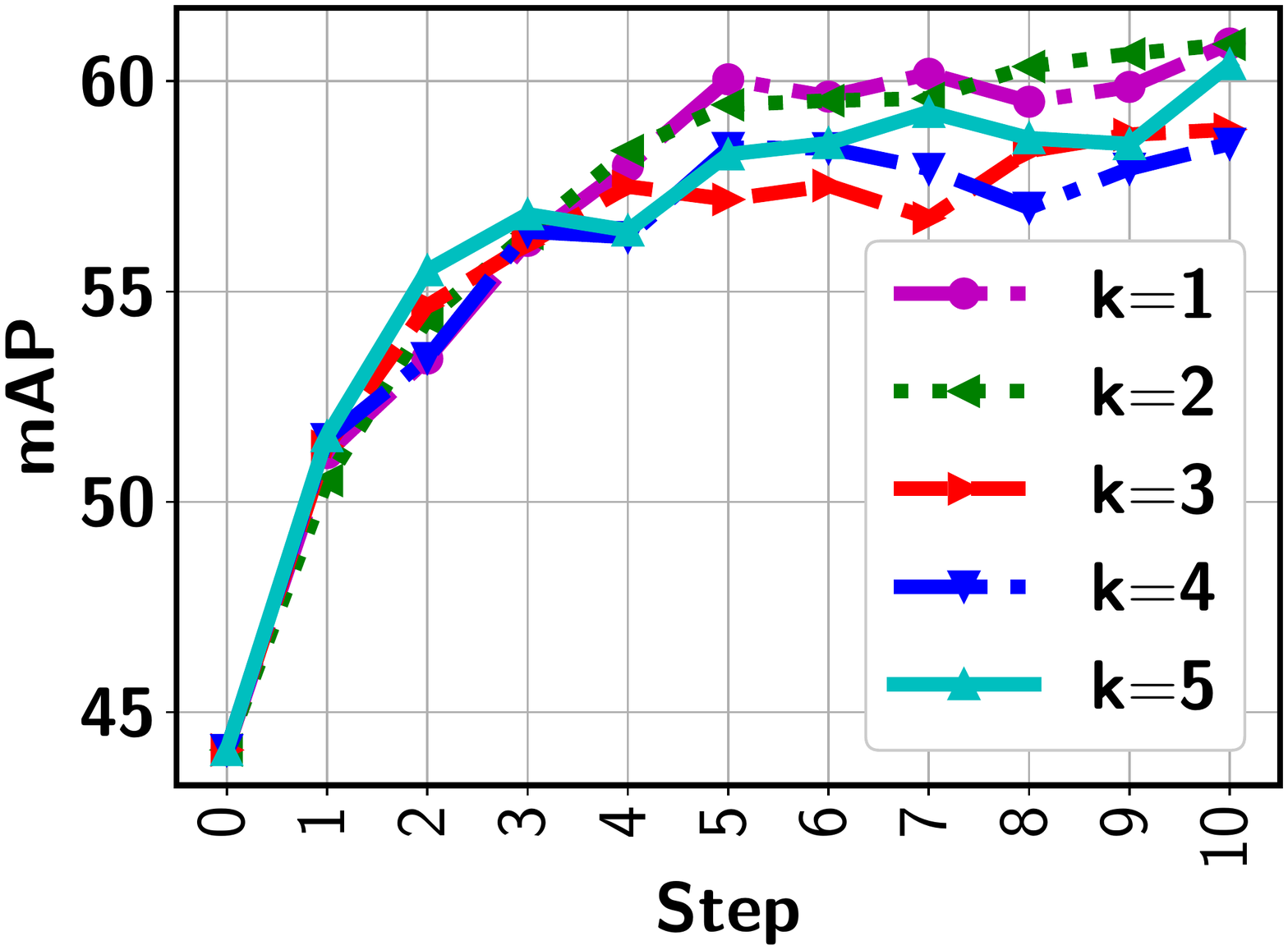}
  \vspace{-18mm}
  \caption{Pascal VOC}
\end{subfigure}
\vspace{-4mm}
\caption{Impact of hyperparameter $k$}
\label{fig: al_curve_ablation_k}
\vspace{-5mm}
\end{figure*}

\vspace{1mm}\noindent{\bf Effectiveness of active sampling.} To demonstrate the effectiveness of the proposed P-F active sampling function, 
we compare it with two other sampling methods, F-Entropy and random sampling, while keeping all other parts of the model the same. As shown in Figure \ref{fig: al_curve_ablation}, P-F sampling clearly outperforms others with a large margin in the first three datasets. It's advantage over F-Entropy is smaller on Pascal VOC due to the same reason as explained above. The performance gain is mainly attributed to the effective exploration of P-F sampling over the most challenging bags.

\vspace{-3mm}
\subsection{Ablation Study}\vspace{-1mm}

\vspace{1mm}\noindent{\bf Impact of $\lambda$ and $\beta$:}
Figures~\ref{fig: ablation_lambda} and ~\ref{fig: ablation_beta} demonstrate the impact of  $\lambda$ (with $\beta = 1$) and $\beta$ (with $\lambda$ = 0.01) to the model performance. In particular, $\lambda$ can be set according to the imbalanced instance distribution within bags, where a larger $\lambda$ corresponds to a higher imbalanced distribution. We vary $\lambda$ in $[10^{-10},1]$ and since most bags in the MIL setting are highly imbalanced,  relatively higher $\lambda$ value  gives very good performance in general. Figure~\ref{fig: ablation_lambda}  shows that $\lambda=0.0001$ clearly outperforms too large (or small) $\lambda$ values. As for $\beta$, placing less emphasis on an instance level loss (small $\beta$), we may not fully leverage labels of queried instances. Meanwhile, with too much emphasis on the instance level loss (large $\beta$), the model overly focuses on the limited queried instances with less attention to the bag labels. Therefore, a good balance  results in an optimal performance, shown in Figures~\ref{fig: ablation_beta}.  %Figure~\ref{fig: ablation_hyperparameters_cifar10_20newsgroup} show the impact of $\lambda$ and $\beta$ on Cifar10 and 20NewsGroup datasets. Similar to the findings on the other two datasets, Figures~\ref{fig: ablation_hyperparameters_cifar10_20newsgroup} (b) and (d)  demonstrate that a $\lambda$ in the middle range outperforms too large (or small) $\lambda$ values. In case of $\beta$, placing too less  (or too much) emphasis may result in overly (or poorly) leveraging annotated instances. Therefore, a good balance between the bag-level and instance-level losses achieves the best result, as shown in Figures~\ref{fig: ablation_hyperparameters_cifar10_20newsgroup} (a) and (c). 

\vspace{1mm}\noindent{\bf Impact of $k$:}
Figure~\ref{fig: al_curve_ablation_k} shows the impact of the hyperparameter $k$, which is the number of instances queried in each unexplored bag, on model performance. As can be seen, $k=2$ achieves a generally decent performance across all the datasets. For datasets with a larger size (\eg Cifar100), a larger $k$ leads to a slightly better performance.

%Figure~\ref{fig: al_curve_comparison_mean_sd} reports the performance comparison with one standard deviation (computed over three runs), which is represented by the vertical black line. As discussed in the main paper, the mean MI-AL curve of ADMIL clearly outperforms other competitive baselines. Meanwhile, the standard deviation of the proposed ADMIL model is also relatively small, which indicates its overall stable MI-AL performance across different datasets in multiple runs.

\begin{figure*}[t!]
\vspace{2mm}
\centering
\begin{subfigure}{0.24\textwidth}
  \centering
    \vspace{-3mm}
  \includegraphics[width=1\linewidth]{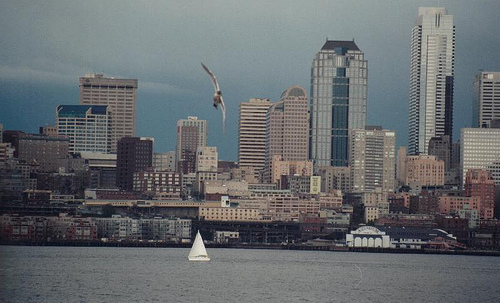}
  \vspace{-1mm}
  \caption{{\sc Sample bag B$_2$}}
  
\end{subfigure}%
~
\begin{subfigure}{0.24\textwidth}
  \centering
    \vspace{-2mm}
  \includegraphics[width=\linewidth, height = 2.5cm]{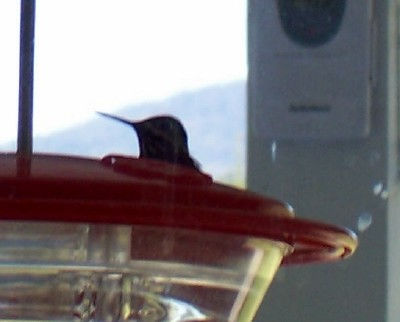}
  \vspace{-1mm}
  \caption{{\sc Sample bag B$_3$}}
\end{subfigure}%
\begin{subfigure}{0.24\textwidth}
  \centering  
  \vspace{-2mm}
\scalebox{.9}{
\begin{tabular}{|c|c|c|}
\hline
\textbf{Bag} & \textbf{P-F} & \textbf{F-Entropy} \\
\hline
{\sc B$_1$} & $1.00$  & $0.04$  \\
\hline
{\sc B$_2$} & $0.53$ &  $0.07$  \\
\hline
{\sc B$_3$}  & $0.64$  & $0.55$  \\
\hline
{\sc B$_1$} &\multicolumn{2}{| l |}{Shadow of a bird} \\ \hline
{\sc B$_2$} &\multicolumn{2}{| l |}{Side view of a bird} \\ \hline
{\sc B$_3$} &\multicolumn{2}{| l |}{Part of a bird} \\ \hline
\end{tabular}}
  \vspace{3mm}
  \caption{{\sc mAP scores}}
\end{subfigure}%
~
\begin{subfigure}{0.24\textwidth}
\centering
\vspace{-15mm}
    \includegraphics[width = 1.\textwidth]{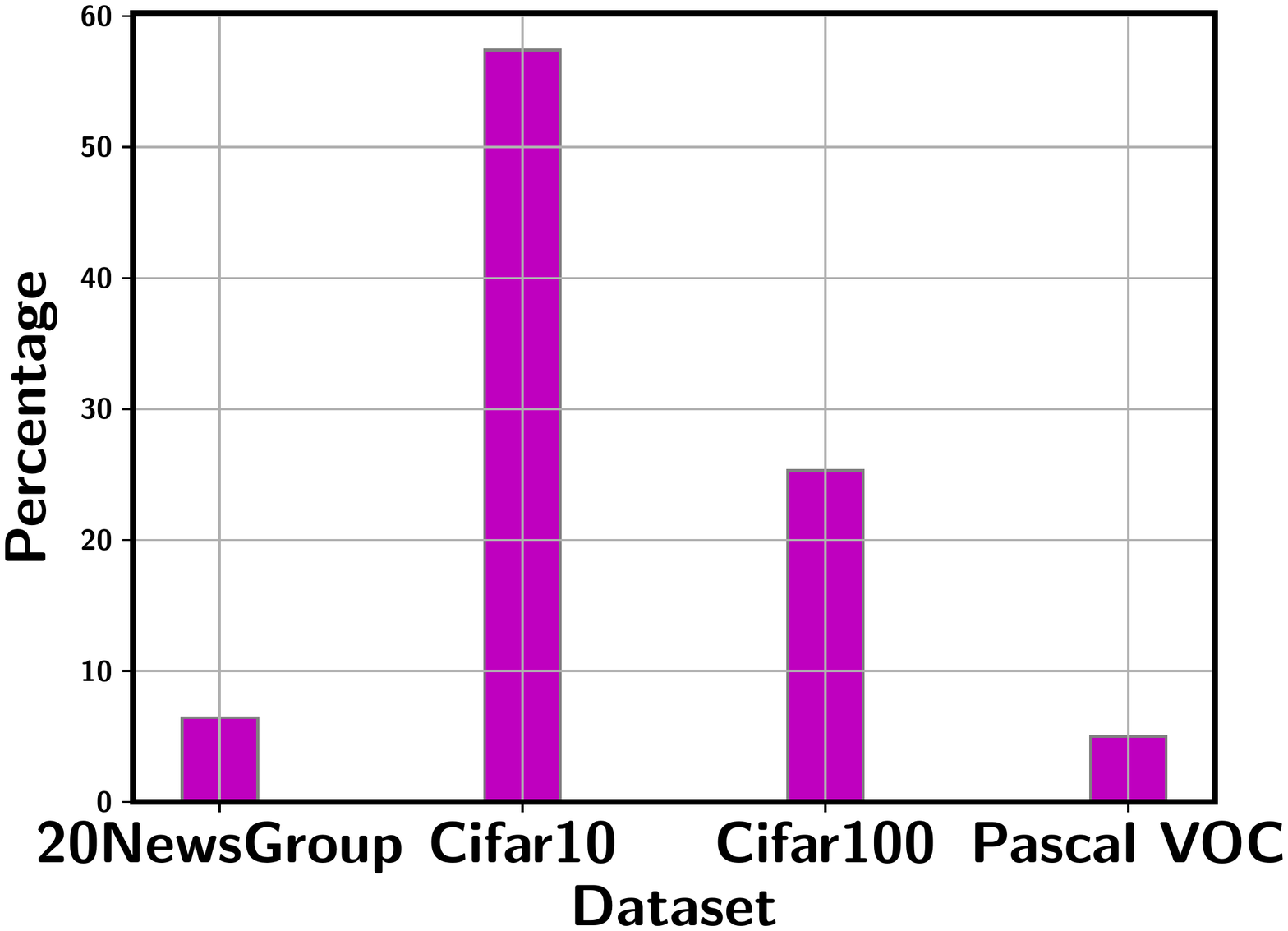}
    \vspace{-18mm}
    \caption{Percentage of TP bags}
 \label{fig: tp_bags}
\end{subfigure}
\vspace{-2mm}
\caption{(a-b) Poorly explored bags in Pascal VOC;  (c) Description of these bags and their mAP scores; (d) Additional true positive bags successfully explored by P-F sampling}
\label{fig: poorly_explored_bags}
\vspace{-3mm}
\end{figure*}

\vspace{-2mm}
\subsection{Qualitative analysis}
 To further justify why the proposed ADMIL model and its P-F sampling function work better than other baselines, we provide a few illustrative examples to offer deeper insights on its good performance. First, we show two challenging bags in addition to the one shown in Figure~\ref{fig: illustrative_examples_max_vs_proposed} (a). As shown in Figure~\ref{fig: poorly_explored_bags} (a-b), B$_2$ presents a side view of a bird while only a small portion of the bird is visible in B$_3$. For those difficult cases, the model originally predicts all instances as a negative with high confidence. However, by coupling the P-F sampling and the hybrid loss in \eqref{eq:total_loss}, the positive instances from those bags are successfully queried. Figure \ref{fig: poorly_explored_bags} (c) shows a clear advantage in the mAP scores between P-F sampling and F-Entropy. As a further evidence, we investigate the number of true positive (TP) bags being explored by both P-F sampling and F-Entropy. TP bags refer to those that the model is being able to query at least one true positive instance. Instead of reporting the actual number of bags, which is affected by the size of the dataset, we show the additional percentage TP bags being explored by P-F sampling in Figure~\ref{fig: poorly_explored_bags} (d). It is worth to note that neither method tries to query the easy bags as their positive instances are correctly predicted with high confidence. The major difference is from the challenging bags and the percentage of these bags varies among different datasets. Nevertheless, P-F sampling consistently explores more effectively than F-Entropy across all datasets. 
\vspace{-2mm}
\section{Conclusion}
To tackle the low instance-level prediction performance of existing MIL models that is essential for many critical applications, 
we develop a novel MI-AL model to sample a small number of most informative instances, especially those from confusing and challenging bags, to enhance the instance-level prediction while keeping a low annotation cost. 
%Considering the inherent challenges in the MIL setting, including imbalanced data distribution and existence of noises and multimodal scenarios, 
We propose to optimize a robust bag likelihood as a convex surrogate of a variance regularized MIL loss to identify a subset of potentially positive instances. Active sampling is conducted by properly balancing between exploring the challenging bags (through P-F sampling) and refining the model by sampling the most confusing instances (through F-Entropy). The design of the loss function naturally supports mini-batch training, which coupled with the batch-mode sampling, makes the MI-AL model work seamlessly with a deep neural network to support broader MIL applications that involve high-dimensional data. Our extensive experiments conducted on multiple MIL datasets show clear advantage over existing baselines. 
%\vspace{-2mm}
\section*{Acknowledgement}
This research was supported in part by an NSF IIS award IIS-1814450 and an ONR award N00014-18-1-2875. The views and conclusions contained in this paper are those of the authors and should not be interpreted as representing any funding agency. 
%\vspace{-2mm}
\bibliographystyle{ACM-Reference-Format}
\bibliography{ref}

%%% -*-BibTeX-*-
%%% Do NOT edit. File created by BibTeX with style
%%% ACM-Reference-Format-Journals [18-Jan-2012].

\begin{thebibliography}{32}

%%% ====================================================================
%%% NOTE TO THE USER: you can override these defaults by providing
%%% customized versions of any of these macros before the \bibliography
%%% command.  Each of them MUST provide its own final punctuation,
%%% except for \shownote{}, \showDOI{}, and \showURL{}.  The latter two
%%% do not use final punctuation, in order to avoid confusing it with
%%% the Web address.
%%%
%%% To suppress output of a particular field, define its macro to expand
%%% to an empty string, or better, \unskip, like this:
%%%
%%% \newcommand{\showDOI}[1]{\unskip}   % LaTeX syntax
%%%
%%% \def \showDOI #1{\unskip}           % plain TeX syntax
%%%
%%% ====================================================================

\ifx \showCODEN    \undefined \def \showCODEN     #1{\unskip}     \fi
\ifx \showDOI      \undefined \def \showDOI       #1{#1}\fi
\ifx \showISBNx    \undefined \def \showISBNx     #1{\unskip}     \fi
\ifx \showISBNxiii \undefined \def \showISBNxiii  #1{\unskip}     \fi
\ifx \showISSN     \undefined \def \showISSN      #1{\unskip}     \fi
\ifx \showLCCN     \undefined \def \showLCCN      #1{\unskip}     \fi
\ifx \shownote     \undefined \def \shownote      #1{#1}          \fi
\ifx \showarticletitle \undefined \def \showarticletitle #1{#1}   \fi
\ifx \showURL      \undefined \def \showURL       {\relax}        \fi
% The following commands are used for tagged output and should be
% invisible to TeX
\providecommand\bibfield[2]{#2}
\providecommand\bibinfo[2]{#2}
\providecommand\natexlab[1]{#1}
\providecommand\showeprint[2][]{arXiv:#2}

\bibitem[Andrews et~al\mbox{.}(2002)]%
        {Andrews2002}
\bibfield{author}{\bibinfo{person}{Stuart Andrews}, \bibinfo{person}{Ioannis
  Tsochantaridis}, {and} \bibinfo{person}{Thomas Hofmann}.}
  \bibinfo{year}{2002}\natexlab{}.
\newblock \showarticletitle{Support Vector Machines for Multiple-Instance
  Learning}. In \bibinfo{booktitle}{\emph{NIPS}}.
\newblock


\bibitem[Ash et~al\mbox{.}(2019)]%
        {ash2019deep}
\bibfield{author}{\bibinfo{person}{Jordan~T Ash}, \bibinfo{person}{Chicheng
  Zhang}, \bibinfo{person}{Akshay Krishnamurthy}, \bibinfo{person}{John
  Langford}, {and} \bibinfo{person}{Alekh Agarwal}.}
  \bibinfo{year}{2019}\natexlab{}.
\newblock \showarticletitle{Deep batch active learning by diverse, uncertain
  gradient lower bounds}.
\newblock \bibinfo{journal}{\emph{arXiv}} (\bibinfo{year}{2019}).
\newblock


\bibitem[Carbonneau et~al\mbox{.}(2018)]%
        {Carbonneau2018MultipleIL}
\bibfield{author}{\bibinfo{person}{Marc-Andr{\'e} Carbonneau},
  \bibinfo{person}{V. Cheplygina}, \bibinfo{person}{Eric Granger}, {and}
  \bibinfo{person}{Ghyslain Gagnon}.} \bibinfo{year}{2018}\natexlab{}.
\newblock \showarticletitle{Multiple instance learning: A survey of problem
  characteristics and applications}.
\newblock \bibinfo{journal}{\emph{Pattern Recognit.}}  \bibinfo{volume}{77}
  (\bibinfo{year}{2018}), \bibinfo{pages}{329--353}.
\newblock


\bibitem[Casanova et~al\mbox{.}(2020)]%
        {Casanova2020ReinforcedAL}
\bibfield{author}{\bibinfo{person}{Arantxa Casanova}, \bibinfo{person}{Pedro
  H.~O. Pinheiro}, \bibinfo{person}{Negar Rostamzadeh}, {and}
  \bibinfo{person}{Christopher~Joseph Pal}.} \bibinfo{year}{2020}\natexlab{}.
\newblock \showarticletitle{Reinforced active learning for image segmentation}.
\newblock \bibinfo{journal}{\emph{ArXiv}}  \bibinfo{volume}{abs/2002.06583}
  (\bibinfo{year}{2020}).
\newblock


\bibitem[Deselaers and Ferrari(2010)]%
        {Deselaers2010}
\bibfield{author}{\bibinfo{person}{Thomas Deselaers} {and}
  \bibinfo{person}{Vittorio Ferrari}.} \bibinfo{year}{2010}\natexlab{}.
\newblock \showarticletitle{A Conditional Random Field for Multiple-Instance
  Learning}. In \bibinfo{booktitle}{\emph{ICML}}. \bibinfo{pages}{287–294}.
\newblock
\showISBNx{9781605589077}


\bibitem[Dietterich et~al\mbox{.}(1997)]%
        {Dietterich1997}
\bibfield{author}{\bibinfo{person}{Thomas~G. Dietterich}, \bibinfo{person}{R.
  Lathrop}, {and} \bibinfo{person}{Tomas Lozano-Perez}.}
  \bibinfo{year}{1997}\natexlab{}.
\newblock \showarticletitle{Solving the Multiple Instance Problem with
  Axis-Parallel Rectangles}.
\newblock \bibinfo{journal}{\emph{Artif. Intell.}}  \bibinfo{volume}{89}
  (\bibinfo{year}{1997}), \bibinfo{pages}{31--71}.
\newblock


\bibitem[Duchi and Namkoong(2019)]%
        {Duci2019}
\bibfield{author}{\bibinfo{person}{John Duchi} {and} \bibinfo{person}{Hongseok
  Namkoong}.} \bibinfo{year}{2019}\natexlab{}.
\newblock \showarticletitle{Variance-based Regularization with Convex
  Objectives}.
\newblock \bibinfo{journal}{\emph{Journal of Machine Learning Research}}
  \bibinfo{volume}{20}, \bibinfo{number}{68} (\bibinfo{year}{2019}),
  \bibinfo{pages}{1--55}.
\newblock


\bibitem[Everingham et~al\mbox{.}(2015)]%
        {Everingham15}
\bibfield{author}{\bibinfo{person}{M. Everingham}, \bibinfo{person}{S.~M.~A.
  Eslami}, \bibinfo{person}{L. Van~Gool}, \bibinfo{person}{C.~K.~I. Williams},
  \bibinfo{person}{J. Winn}, {and} \bibinfo{person}{A. Zisserman}.}
  \bibinfo{year}{2015}\natexlab{}.
\newblock \showarticletitle{The Pascal Visual Object Classes Challenge: A
  Retrospective}.
\newblock \bibinfo{journal}{\emph{IJCV}} \bibinfo{volume}{111},
  \bibinfo{number}{1} (\bibinfo{year}{2015}), \bibinfo{pages}{98--136}.
\newblock


\bibitem[Gal and Ghahramani(2015)]%
        {gal2015bayesian}
\bibfield{author}{\bibinfo{person}{Yarin Gal} {and} \bibinfo{person}{Zoubin
  Ghahramani}.} \bibinfo{year}{2015}\natexlab{}.
\newblock \showarticletitle{Bayesian convolutional neural networks with
  Bernoulli approximate variational inference}.
\newblock \bibinfo{journal}{\emph{arXiv}} (\bibinfo{year}{2015}).
\newblock


\bibitem[Haußmann et~al\mbox{.}(2017)]%
        {Haubmann2017}
\bibfield{author}{\bibinfo{person}{Manuel Haußmann}, \bibinfo{person}{Fred~A.
  Hamprecht}, {and} \bibinfo{person}{Melih Kandemir}.}
  \bibinfo{year}{2017}\natexlab{}.
\newblock \showarticletitle{Variational Bayesian Multiple Instance Learning
  with Gaussian Processes}. In \bibinfo{booktitle}{\emph{CVPR}}.
  \bibinfo{pages}{810--819}.
\newblock


\bibitem[Hsu et~al\mbox{.}(2020)]%
        {Hsu2020QueryDrivenML}
\bibfield{author}{\bibinfo{person}{Yen-Chi Hsu}, \bibinfo{person}{Cheng-Yao
  Hong}, \bibinfo{person}{Ming-Sui Lee}, {and} \bibinfo{person}{Tyng-Luh Liu}.}
  \bibinfo{year}{2020}\natexlab{}.
\newblock \showarticletitle{Query-Driven Multi-Instance Learning}. In
  \bibinfo{booktitle}{\emph{AAAI}}.
\newblock


\bibitem[Ilse et~al\mbox{.}(2018)]%
        {Ilse2018}
\bibfield{author}{\bibinfo{person}{Maximilian Ilse}, \bibinfo{person}{Jakub~M.
  Tomczak}, {and} \bibinfo{person}{Max Welling}.}
  \bibinfo{year}{2018}\natexlab{}.
\newblock \showarticletitle{Attention-based Deep Multiple Instance Learning}.
  In \bibinfo{booktitle}{\emph{ICML}}.
\newblock


\bibitem[Kendall et~al\mbox{.}(2015)]%
        {kendall2015bayesian}
\bibfield{author}{\bibinfo{person}{Alex Kendall}, \bibinfo{person}{Vijay
  Badrinarayanan}, {and} \bibinfo{person}{Roberto Cipolla}.}
  \bibinfo{year}{2015}\natexlab{}.
\newblock \showarticletitle{Bayesian segnet: Model uncertainty in deep
  convolutional encoder-decoder architectures for scene understanding}.
\newblock \bibinfo{journal}{\emph{arXiv}} (\bibinfo{year}{2015}).
\newblock


\bibitem[Kim and Torre(2010)]%
        {Kim2010}
\bibfield{author}{\bibinfo{person}{Minyoung Kim} {and} \bibinfo{person}{F.
  Torre}.} \bibinfo{year}{2010}\natexlab{}.
\newblock \showarticletitle{Gaussian Processes Multiple Instance Learning}. In
  \bibinfo{booktitle}{\emph{ICML}}.
\newblock


\bibitem[Kirsch et~al\mbox{.}(2019)]%
        {kirsch2019batchbald}
\bibfield{author}{\bibinfo{person}{Andreas Kirsch}, \bibinfo{person}{Joost
  Van~Amersfoort}, {and} \bibinfo{person}{Yarin Gal}.}
  \bibinfo{year}{2019}\natexlab{}.
\newblock \showarticletitle{Batchbald: Efficient and diverse batch acquisition
  for deep bayesian active learning}.
\newblock \bibinfo{journal}{\emph{arXiv}} (\bibinfo{year}{2019}).
\newblock


\bibitem[Krizhevsky(2009)]%
        {Krizhevsky2009LearningML}
\bibfield{author}{\bibinfo{person}{A. Krizhevsky}.}
  \bibinfo{year}{2009}\natexlab{}.
\newblock \showarticletitle{Learning Multiple Layers of Features from Tiny
  Images}.
\newblock \bibinfo{journal}{\emph{University of Toronto}}
  (\bibinfo{year}{2009}).
\newblock


\bibitem[Lam(2016)]%
        {Lam2016RobustSA}
\bibfield{author}{\bibinfo{person}{Henry Lam}.}
  \bibinfo{year}{2016}\natexlab{}.
\newblock \showarticletitle{Robust Sensitivity Analysis for Stochastic
  Systems}.
\newblock \bibinfo{journal}{\emph{Math. Oper. Res.}}  \bibinfo{volume}{41}
  (\bibinfo{year}{2016}), \bibinfo{pages}{1248--1275}.
\newblock


\bibitem[Leibig et~al\mbox{.}(2017)]%
        {leibig2017leveraging}
\bibfield{author}{\bibinfo{person}{Christian Leibig}, \bibinfo{person}{Vaneeda
  Allken}, \bibinfo{person}{Murat~Se{\c{c}}kin Ayhan}, \bibinfo{person}{Philipp
  Berens}, {and} \bibinfo{person}{Siegfried Wahl}.}
  \bibinfo{year}{2017}\natexlab{}.
\newblock \showarticletitle{Leveraging uncertainty information from deep neural
  networks for disease detection}.
\newblock \bibinfo{journal}{\emph{Scientific reports}} \bibinfo{volume}{7},
  \bibinfo{number}{1} (\bibinfo{year}{2017}), \bibinfo{pages}{1--14}.
\newblock


\bibitem[Li and Vasconcelos(2015)]%
        {Li2015}
\bibfield{author}{\bibinfo{person}{Weixin Li} {and} \bibinfo{person}{N.
  Vasconcelos}.} \bibinfo{year}{2015}\natexlab{}.
\newblock \showarticletitle{Multiple instance learning for soft bags via top
  instances}.
\newblock \bibinfo{journal}{\emph{2015 IEEE Conference on Computer Vision and
  Pattern Recognition (CVPR)}} (\bibinfo{year}{2015}),
  \bibinfo{pages}{4277--4285}.
\newblock


\bibitem[Namkoong and Duchi(2017)]%
        {Namkoong2017}
\bibfield{author}{\bibinfo{person}{Hongseok Namkoong} {and}
  \bibinfo{person}{John~C Duchi}.} \bibinfo{year}{2017}\natexlab{}.
\newblock \showarticletitle{Variance-based Regularization with Convex
  Objectives}. In \bibinfo{booktitle}{\emph{Advances in Neural Information
  Processing Systems}}, \bibfield{editor}{\bibinfo{person}{I.~Guyon},
  \bibinfo{person}{U.~V. Luxburg}, \bibinfo{person}{S.~Bengio},
  \bibinfo{person}{H.~Wallach}, \bibinfo{person}{R.~Fergus},
  \bibinfo{person}{S.~Vishwanathan}, {and} \bibinfo{person}{R.~Garnett}}
  (Eds.), Vol.~\bibinfo{volume}{30}. \bibinfo{publisher}{Curran Associates,
  Inc.}
\newblock


\bibitem[Petersen et~al\mbox{.}(2000)]%
        {Petersen}
\bibfield{author}{\bibinfo{person}{I.R. Petersen}, \bibinfo{person}{M.R.
  James}, {and} \bibinfo{person}{P. Dupuis}.} \bibinfo{year}{2000}\natexlab{}.
\newblock \showarticletitle{Minimax optimal control of stochastic uncertain
  systems with relative entropy constraints}.
\newblock \bibinfo{journal}{\emph{IEEE Trans. Automat. Control}}
  \bibinfo{volume}{45}, \bibinfo{number}{3} (\bibinfo{year}{2000}),
  \bibinfo{pages}{398--412}.
\newblock


\bibitem[Sener and Savarese(2017)]%
        {sener2017active}
\bibfield{author}{\bibinfo{person}{Ozan Sener} {and} \bibinfo{person}{Silvio
  Savarese}.} \bibinfo{year}{2017}\natexlab{}.
\newblock \showarticletitle{Active learning for convolutional neural networks:
  A core-set approach}.
\newblock \bibinfo{journal}{\emph{arXiv}} (\bibinfo{year}{2017}).
\newblock


\bibitem[Settles(2009)]%
        {settles2009active}
\bibfield{author}{\bibinfo{person}{Burr Settles}.}
  \bibinfo{year}{2009}\natexlab{}.
\newblock \showarticletitle{Active learning literature survey}.
\newblock  (\bibinfo{year}{2009}).
\newblock


\bibitem[Settles et~al\mbox{.}(2007)]%
        {settles2007multiple}
\bibfield{author}{\bibinfo{person}{Burr Settles}, \bibinfo{person}{Mark
  Craven}, {and} \bibinfo{person}{Soumya Ray}.}
  \bibinfo{year}{2007}\natexlab{}.
\newblock \showarticletitle{Multiple-instance active learning}.
\newblock \bibinfo{journal}{\emph{NIPS}}  \bibinfo{volume}{20}
  (\bibinfo{year}{2007}), \bibinfo{pages}{1289--1296}.
\newblock


\bibitem[Settles et~al\mbox{.}(2008)]%
        {Settles2008}
\bibfield{author}{\bibinfo{person}{Burr Settles}, \bibinfo{person}{Mark
  Craven}, {and} \bibinfo{person}{Soumya Ray}.}
  \bibinfo{year}{2008}\natexlab{}.
\newblock \showarticletitle{Multiple-Instance Active Learning}. In
  \bibinfo{booktitle}{\emph{NIPS}},
  \bibfield{editor}{\bibinfo{person}{J.~Platt}, \bibinfo{person}{D.~Koller},
  \bibinfo{person}{Y.~Singer}, {and} \bibinfo{person}{S.~Roweis}} (Eds.),
  Vol.~\bibinfo{volume}{20}.
\newblock


\bibitem[Su et~al\mbox{.}(2015)]%
        {su2015relationship}
\bibfield{author}{\bibinfo{person}{Wanhua Su}, \bibinfo{person}{Yan Yuan},
  {and} \bibinfo{person}{Mu Zhu}.} \bibinfo{year}{2015}\natexlab{}.
\newblock \showarticletitle{A relationship between the average precision and
  the area under the ROC curve}. In \bibinfo{booktitle}{\emph{ICTIR}}.
  \bibinfo{pages}{349--352}.
\newblock


\bibitem[Sultani et~al\mbox{.}(2018)]%
        {Sultani2018}
\bibfield{author}{\bibinfo{person}{Waqas Sultani}, \bibinfo{person}{Chen Chen},
  {and} \bibinfo{person}{Mubarak Shah}.} \bibinfo{year}{2018}\natexlab{}.
\newblock \showarticletitle{Real-World Anomaly Detection in Surveillance
  Videos}. In \bibinfo{booktitle}{\emph{CVPR}}.
\newblock


\bibitem[Wang et~al\mbox{.}(2016)]%
        {wang2016cost}
\bibfield{author}{\bibinfo{person}{Keze Wang}, \bibinfo{person}{Dongyu Zhang},
  \bibinfo{person}{Ya Li}, \bibinfo{person}{Ruimao Zhang}, {and}
  \bibinfo{person}{Liang Lin}.} \bibinfo{year}{2016}\natexlab{}.
\newblock \showarticletitle{Cost-effective active learning for deep image
  classification}.
\newblock \bibinfo{journal}{\emph{IEEE Transactions on Circuits and Systems for
  Video Technology}} \bibinfo{volume}{27}, \bibinfo{number}{12}
  (\bibinfo{year}{2016}), \bibinfo{pages}{2591--2600}.
\newblock


\bibitem[Xu and Frank(2004)]%
        {Xu2004}
\bibfield{author}{\bibinfo{person}{Xin Xu} {and} \bibinfo{person}{Eibe Frank}.}
  \bibinfo{year}{2004}\natexlab{}.
\newblock \showarticletitle{Logistic Regression and Boosting for Labeled Bags
  of Instances}. In \bibinfo{booktitle}{\emph{PAKDD}}.
\newblock


\bibitem[Yuan et~al\mbox{.}(2021)]%
        {Yuan2021}
\bibfield{author}{\bibinfo{person}{Tianning Yuan}, \bibinfo{person}{Fang Wan},
  \bibinfo{person}{Mengying Fu}, \bibinfo{person}{Jianzhuang Liu},
  \bibinfo{person}{Songcen Xu}, \bibinfo{person}{Xiangyang Ji}, {and}
  \bibinfo{person}{Qixiang Ye}.} \bibinfo{year}{2021}\natexlab{}.
\newblock \showarticletitle{Multiple instance active learning for object
  detection}.
\newblock \bibinfo{journal}{\emph{CoRR}}  \bibinfo{volume}{abs/2104.02324}
  (\bibinfo{year}{2021}).
\newblock
\showeprint[arxiv]{2104.02324}


\bibitem[Zhou et~al\mbox{.}(2009)]%
        {Zhou2009}
\bibfield{author}{\bibinfo{person}{Zhi-Hua Zhou}, \bibinfo{person}{Yu-Yin Sun},
  {and} \bibinfo{person}{Yu-Feng Li}.} \bibinfo{year}{2009}\natexlab{}.
\newblock \showarticletitle{Multi-Instance Learning by Treating Instances as
  Non-I.I.D. Samples}. In \bibinfo{booktitle}{\emph{ICML}}.
  \bibinfo{pages}{1249–1256}.
\newblock
\showISBNx{9781605585161}


\bibitem[Zhu et~al\mbox{.}(2019)]%
        {Zhu2019}
\bibfield{author}{\bibinfo{person}{Dixian Zhu}, \bibinfo{person}{Zhe Li},
  \bibinfo{person}{Xiaoyu Wang}, \bibinfo{person}{Boqing Gong}, {and}
  \bibinfo{person}{Tianbao Yang}.} \bibinfo{year}{2019}\natexlab{}.
\newblock \showarticletitle{A Robust Zero-Sum Game Framework for Pool-based
  Active Learning}. In \bibinfo{booktitle}{\emph{AISTATS}},
  Vol.~\bibinfo{volume}{89}. \bibinfo{pages}{517--526}.
\newblock


\end{thebibliography}
\newpage
\appendix

%Finally, we provide the link to the source code in Appendix~\ref{app:sourcecode}.

\section{Proofs of Theorems} \label{app:proof}
In this section, we provide the detailed proofs for both theorems. 

%{\bf Proof of Theorem 1.} 
\paragraph{Proof of Theorem 1} 
%\begin{proof}
Our proof of Theorem 1 is adapted from~\cite{Duci2019} by making extensions that fit the unique design of the distributionally robust bag likelihood (DRBL). We start by introducing the following lemma, which will later be used in our proof.

\begin{lemma}[Maurer and Pontil Theorem 10]\label{le:variance}
Let $Y$ be a random variable taking values in [0, L]. Let $\sigma^2=\text{Var}[Y]$ and $\text{Var}_n[Y]=\frac{1}{n}\sum_{i=1}^nY_i^2-(\frac{1}{n}\sum_{i=1}^nY_i)^2$ be the population and sample variance of $Y$, respectively. Then for for $n\geq 2$,
\begin{align}
\label{eq:lemma}
P(\sigma-t\leq \sqrt{Var_n[Y]}\leq\sigma+t)\geq 1-\exp\left({-\frac{nt^2}{2L^2}}\right)    
\end{align}
%$$P(\sigma-t\leq \sqrt{\text{Var}_n[f(X^+)]}\leq\sigma+t)\geq 1-\exp\left({-\frac{nt^2}{2M^2}}\right)$$
\end{lemma}
The distributionally robust bag likelihood (DRBL), \ie the l.h.s. of \eqref{eq:dro-variance-eqivalence}, can be formulated as the following constrained optimization problem:
\begin{equation}
\begin{aligned}
\max_{{\bf p}\in {\mathcal{P}_n}} & \sum_{i=1}^np_if({\bf x}_i^+) \\
\text{s.t.~} & \mathcal{P}_n :=\left\{{\bf p}\in \mathbb{R}^n, {\bf p}^{\top}\mathbbm{1}=1, 0\leq {\bf p}, D_f\left({{\bf p}||\frac{\mathbbm{1}}{n}}\right)\leq \frac{\lambda}{n} \right\}
    \label{eq:dro_optimization_function}    
\end{aligned}
\end{equation}
%\max_{{\bf p}}\sum_{i=1}^n p_if({\bf x}_i^+) \quad \text{s.t.~} \mathcal{P}_n:=\{{\bf p}\in R^n_+, ||n{\bf p}-1||_2^2\leq\lambda, {\bf p}^T\mathbbm{1} = 1\}
Since the $D_f ({\bf p}||{\bf q})$ is assumed to be the $\chi^2$-divergence and ${\bf q}$ follows the uniform distribution, $D_f ({\bf p}||{\bf q})$ is reduced to the squared Euclidean distance.
%\textbf{Note:} Since the second term in $D_f\left({{\bf p}||\frac{\mathbbm{1}}{n}}\right)$ is the uniform distribution and in this case, the $\chi^2$ divergence becomes proportional to the square Euclidean distance. Therefore, in our proof, we proceed using the squared Euclidean distance metric. 
We first introduce the mean of $f({\bf x}_i^+)$'s, which is denoted as $\bar{f} = \frac{1}{n}\sum_{i=1}^nf({\bf x}_i^+)$. Also, recall we denote the score vector by ${\bf f}=(f({\bf x}_1^+),...,f({\bf x}_n^+))^{\top}$ in Section~\ref{sec:drbl}. Thus, the empirical variance of $f(X^+)$ is given by $\text{Var}_n[f(X^+)] = \frac{1}{n}||{\bf f}||_2^2-\bar{f}^2 = \frac{1}{n}||{\bf f}-\bar{f}\mathbbm{1}||_2^2$. We further introduce  ${\bf u} = {\bf p}-\frac{\mathbbm{1}}{n}$, so the objective in \eqref{eq:dro_optimization_function} can be transformed as 
\begin{equation}
{\bf p}^{\top}{\bf f} = ({\bf u}+\frac{\mathbbm{1}}{n})^{\top}{\bf f}=\bar{f}+{\bf u}^{\top}{\bf f} = \bar{f}+{\bf u}^{\top}({\bf f}-\bar{f}\mathbbm{1})   
\end{equation}
where the last equality holds because ${\bf u}^{\top}\mathbbm{1} = 0$. Thus, the optimization problem in \eqref{eq:dro_optimization_function} can be further transformed into
%indicates the mean. \qi{Please update the notation in the proof. We introduced $f(X^+)=(f({\bf x}_1^+),...,f({\bf x}_n^+))^{\top}$ to represent a vector of $f({\bf x}_i^+)$'s in the main paper. Only bold vectors not scalars. Use $^{\top}$ for transpose not $^T$.} Further to keep notations uncluttered, let $f(X^+)=(f({\bf x}_1^+),...,f({\bf x}_n^+))^{\top}$. Further, consider $\text{Var}_n[f(X^+)] = \frac{1}{n}||f(X^+)||_2^2-\bar{f}^2 = \frac{1}{n}||f(X^+)-\bar{f}||_2^2$ denotes the empirical variance of the vector $\bf f$.
%Then by introducing the variable ${\bf u} = {\bf p}-\frac{\mathbbm{1}}{n}$ then the objective in problem \ref{eq:dro_optimization_function}, satisfies ${\bf p}^{\top}f(X^+) = \bar{f}+{\bf u}^{\top}f(X^+) = \bar{f}+{\bf u}^{\top}(f(X^+)-\bar{f})$ with ${\bf u}^{\top}\mathbbm{1} = 0$. Thus the above problem is equivalent to solving
\begin{equation}
\max_{{\bf u}\in {\mathbb R}^n}\bar{f}+{\bf u}^{\top}({\bf f}-\bar{f}\mathbbm{1}) \quad \text{s.t.} \quad ||{\bf u}||_2^2\leq\frac{\lambda}{n^2}, {\bf u}^{\top}\mathbbm{1} = 0, {\bf u}\geq -\frac{1}{n}
    \label{eq:dro_optimization_u}
\end{equation}
where the first constraint is derived by replacing $D_f$ with the $\chi^2$-divergence. Now, using the Cauchy-Schwarz inequality, which states that ${\bf u}^{\top}{\bf v}\leq ||{\bf u}||_2||{\bf v}||_2$, gives the following condition
\begin{equation}
    {\bf u}^{\top}({\bf f}-\bar{f}\mathbbm{1}) \leq \frac{\sqrt{\lambda}}{n}||{\bf f}-\bar{f}\mathbbm{1}||_2 =\sqrt\frac{{\lambda \text{Var}_n[f(X^+)]}}{n}
    \label{eq:cauchy_schwarz}
\end{equation}
where the equality holds if and only if 
\begin{equation}
u_i=\frac{\sqrt{\lambda} (f({\bf x}_i^+)-\bar{f})}{n||{\bf f}-\bar{f}\mathbbm{1}||_2} = \frac{\sqrt{\lambda} (f({\bf x}_i^+)-\bar{f})}{n\sqrt{n\text{Var}_n[f(X^+)]}}
    \label{eq:cauchy_schwarz_equality}
\end{equation}
Since we also have a constraint ${\bf u}\geq -\frac{1}{n}$, which satisfies if and only if 
\begin{equation}
\min_{i\in [n]}\frac{\sqrt{\lambda}(f({\bf x}_i^+)-\bar{f})}{\sqrt{n\text{Var}_n[f(X^+)]}} \geq -1
    \label{eq:equality_condition}
\end{equation}

Thus, if inequality \eqref{eq:equality_condition} holds for vector $\bf f$, we have
\begin{equation}
\max_{{\bf p}\in \mathcal{P}_n}{\bf p}^{\top}{\bf f} = \bar{f}+\sqrt{\frac{\lambda \text{Var}_n[f(X^+)]}{n}}
    \label{eq:dro_variance}
\end{equation}
which will prove the Theorem given in \eqref{eq:dro-variance-eqivalence}. 

What remains is to prove inequality \eqref{eq:equality_condition} holds with a high probability. To show this, we leverage the concentration inequality given by Lemma~\ref{le:variance}. Since $f({\bf x}_i^+) \in [0,1]$, we have $|f({\bf x}_i^+)-\bar{f}| \leq 1$. To satisfy inequality \eqref{eq:equality_condition}, it is sufficient to have
\begin{align}
  \frac{\lambda}{n\text{Var}_n[f(X^+)]}\leq 1\quad  
    &\text{or} \quad \text{Var}_n[f(X^+)]\geq \frac{\lambda}{n} \label{eq:sufficient}
\end{align}
%Now with the probability $1-\exp\left(-{\frac{77n\sigma^2}{125}}\right)$, we show that above inequality holds.
Let us define the following event 
\begin{equation}
\epsilon_n:=\left\{\text{Var}_n[f(X^+)]\geq \frac{1}{43}\sigma^2\right\}
    \label{eq:event}
\end{equation}
In Theorem 1, we suppose $n\geq\frac{4\lambda}{\sigma^2}\max\{2\sigma, 11\})$. Then, on event $\epsilon_n$, we have $n\geq \frac{44\lambda}{\sigma^2}\geq \frac{\lambda}{\text{Var}_n[f(X^+)]}$, so that the sufficient condition \eqref{eq:sufficient} holds and the \eqref{eq:dro_variance} becomes true. 

Now we find the probability of holding the above event in \eqref{eq:event} using Lemma~\ref{le:variance}. 
%\begin{lemma}[Maurer and Pontil Theorem 10]
%Let $f({\bf x}_i^+)$ be i.i.d. random variable taking the values in [0, M] and $\text{Var}_n[f(X^+)] = \frac{1}{n}||f(X^+)||_2^2-\bar{f}^2$. Then for for $n\geq 2$
%$$P(\sigma-t\leq \sqrt{\text{Var}_n[f(X^+)]}\leq\sigma+t)\geq 1-\exp\left({-\frac{nt^2}{2M^2}}\right)$$
%\end{lemma}
%In our case M = 1 and 
First, $L=1$ in our case, which gives
$$P(\sigma-t\leq \sqrt{\text{Var}_n[f(X^+)]}\leq\sigma+t)\geq 1-\exp\left({-\frac{nt^2}{2}}\right)$$
The following also holds true:
$$P\left(\sigma-t\leq \sqrt{\text{Var}_n[f(X^+)]}\right)\geq P(\sigma-t\leq \sqrt{\text{Var}_n[f(X^+)]}\leq\sigma+t)$$
$$\geq 1-\exp\left({-\frac{nt^2}{2}}\right)$$
Let $t = \left(1-\sqrt{\frac{1}{43}}\right)\sigma$, which gives $\sigma-t = \sqrt{\frac{1}{43}}\sigma$. Substituting this to \eqref{eq:lemma} leads to
$$P\left(\sqrt{\frac{1}{43}}\sigma\leq \sqrt{\text{Var}_n[f(X^+)]}\right)\geq 1-\exp\left({-\frac{nt^2}{2}}\right); P(\epsilon_n)\geq 1-\exp\left({-\frac{nt^2}{2}}\right)$$
Further substituting the value of $t = \left(1-\sqrt{\frac{1}{43}}\right)\sigma$ gives rise to
$$P(\epsilon_n)\geq 1-\exp\left({-0.359n\sigma^2}\right) \geq 1-\exp\left({-\frac{7n\sigma^2}{20}}\right)$$

This completes the proof of Theorem 1.
\paragraph{Proof of Theorem 2}

%\begin{proof}
In order to prove this theorem, we consider two assumptions, which both hold true for our MIL setting.  

%The first assumption is related to the finiteness which is as follow

 \begin{figure*}[t!]
\centering
\vspace{-14mm}
\begin{subfigure}{0.25\textwidth}
  \centering
  \includegraphics[width=\linewidth]{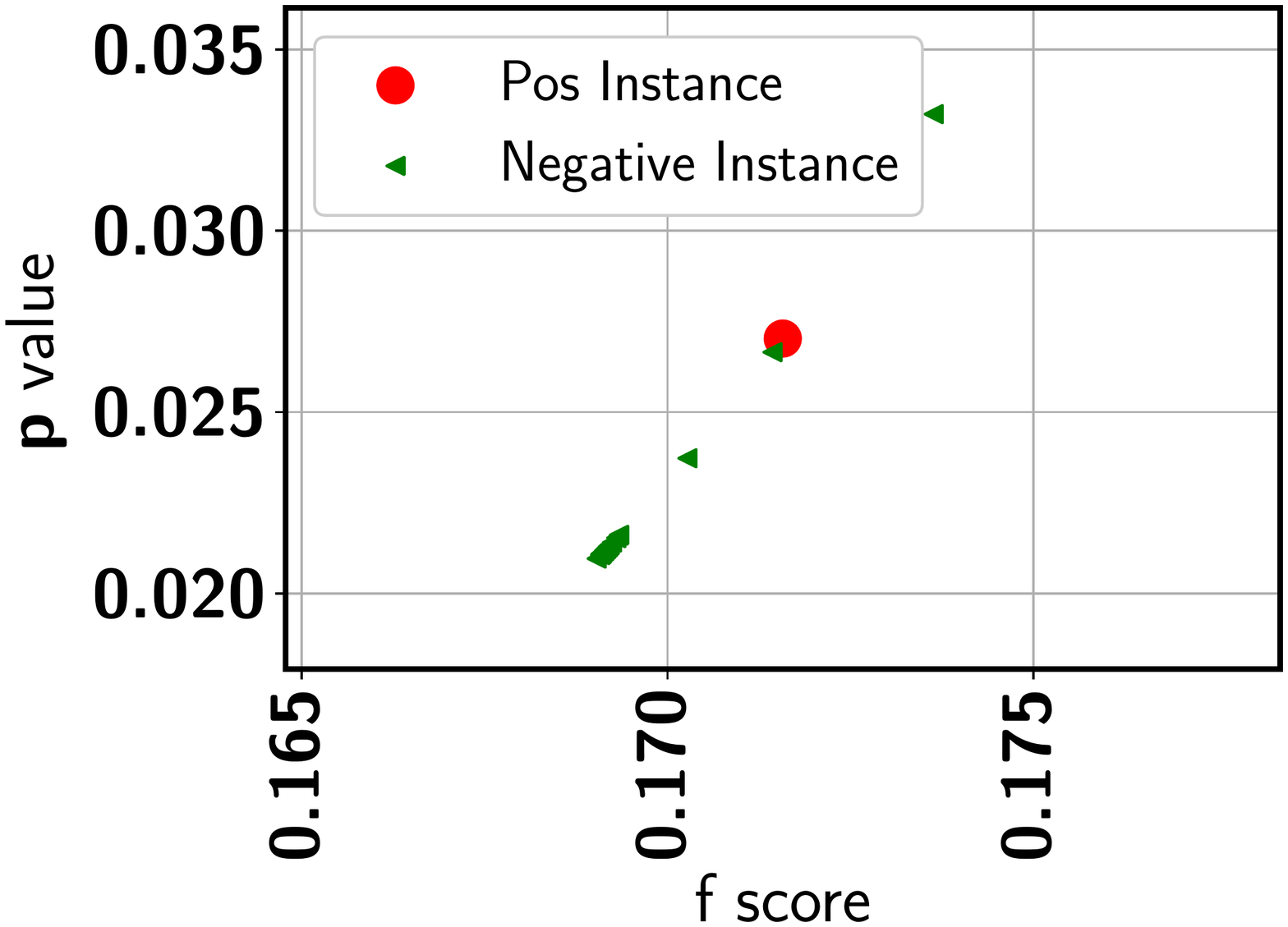}
  %\vspace{-4mm}
  %\caption{{\sc comp.graphics B-8}}
  
\end{subfigure}%
~
\begin{subfigure}{0.25\textwidth}
  \centering
  \includegraphics[width=\linewidth]{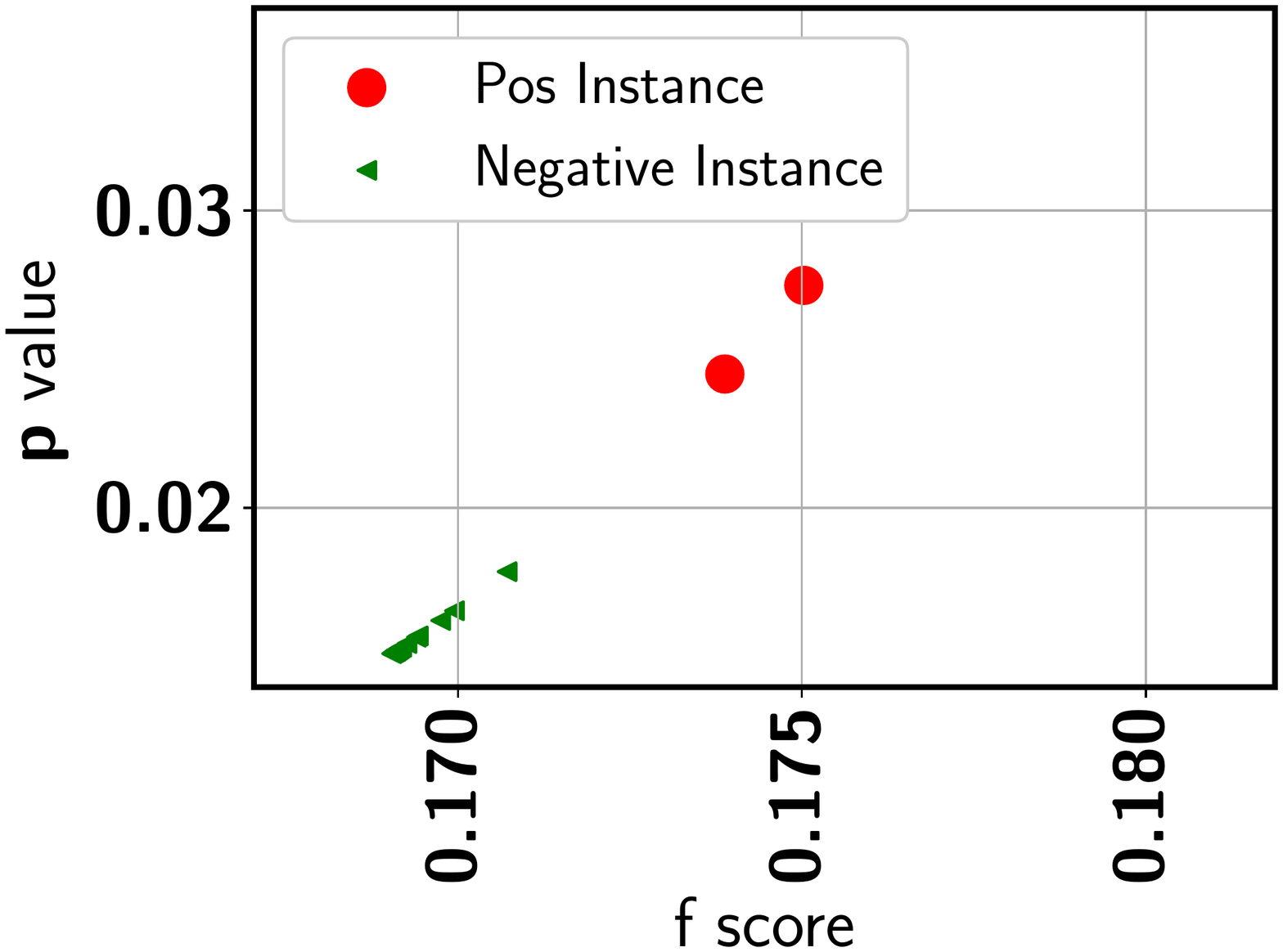}
 % \vspace{-4mm}
  %\caption{{\sc rec.motorcycles B-2}}
\end{subfigure}%
~
\begin{subfigure}{0.25\textwidth}
  \centering
  \includegraphics[width=\linewidth]{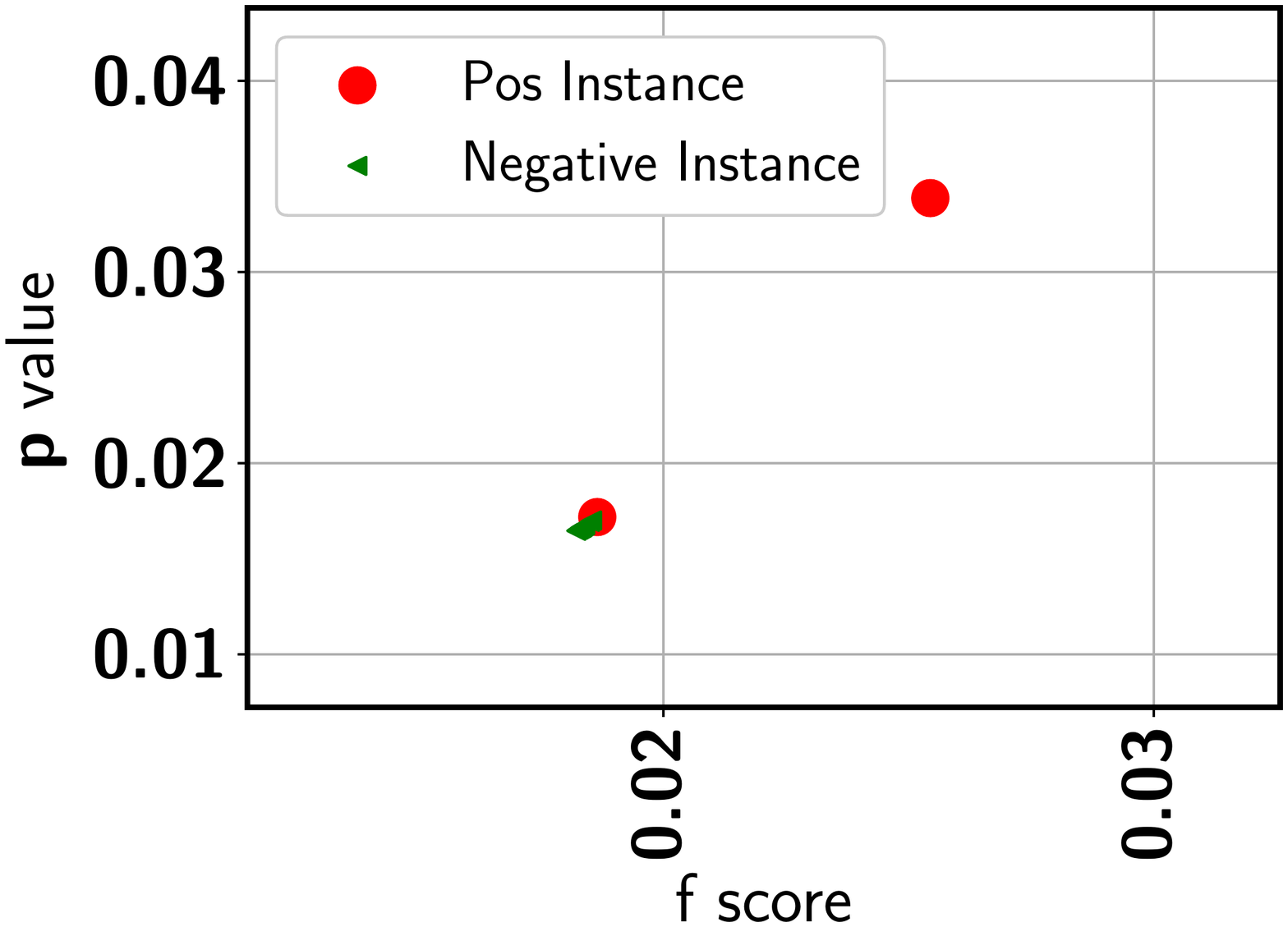}
  %\vspace{-4mm}
  %\caption{{\sc talk.politics.misc B-13}}
\end{subfigure}
\vspace{-16mm}
\caption{\label{fig:morebagexamples}Example of challenging bags from different topics in 20NewsGroup}
\vspace{-2mm}
\end{figure*}

{\bf Assumption 1:} Random variable $f(X^+)$ has a finite exponential moment in a neighborhood of 0 under the distribution ${\bf p}_0$ \ie $\mathbb{E}_0[\exp(\tau f(X^+))]<\infty$ for $\tau \in [-\tau_0, \tau_0]$ for some $\tau_0>0$.

%The second assumption is related to the non-degeneracy condition as specified below.
{\bf Assumption 2:} Random variable $f(X^+)$ is non-constant under ${\bf p}_0$. 

Assumption 1 is true in our case as $f(X^+)$ is bounded in [0, 1]; Assumption 2 also empirically holds true as there are both positive and negative instances in a positive bag so the output scores are distinct over different instances in a bag. The second assumption ensures that the uniform distribution ${\bf p}_0$ is not a locally optimum, which means there exists an opportunity to upgrade the value by re-balancing the probability between positive and negative instances in a positive bag.

Consider ${\bf p}$ that is absolutely continuous with respect to ${\bf p}_0$ and therefore the likelihood ratio $g=\frac{d{\bf p}}{d{\bf p}_0}$ (a.k.a., Radon–Nikodym derivative) exists. Using a change of measure, the optimization problem in the l.h.s. of \eqref{eq:dro_kl_variance_equivalence} can be written as

\begin{equation}
\begin{aligned}
    \max_{g \in \mathcal{L}_1({\bf p}_0)} \mathbb{E}_0[g f(X^+)]
    \; \text{s.t.} \left\{\mathbb{E}_0[g\log g]\leq \frac{\lambda}{n}, \mathbb{E}_0[g] = 1, {g\geq 0}\right\} 
  \end{aligned}  
  \label{eq: optimization_likelihood_space}
\end{equation}
%where $\mathcal{L}: = \left\{g \in \mathcal{L}_1({\bf p}_0)\right\}$ and we denote ${\mathcal L}_1({\bf p}_0)$ as $\mathcal{L}_1$-space with respect to the measure ${\bf p}_0$. The key in the above optimization is to find the optimal solution $g^*$ first.
where ${\mathcal L}_1({\bf p}_0)$ is $\mathcal{L}_1$-space with respect to the measure ${\bf p}_0$. To solve the optimization problem above, we formulate its Lagrangian, 
\begin{equation}
    \max_{g \in \mathcal{L}_1({\bf p}_0)} \mathbb{E}_0[g f(X^+)]-\alpha\left(\mathbb{E}_0[g\log g]-\frac{\lambda}{n}\right)
    \label{eq: lagrangian_kl_dro}
\end{equation}
where $\alpha$ is the Lagrange's multiplier. The solution of the above objective function is given by the following proposition \cite{Petersen,Lam2016RobustSA}:
\begin{proposition}
Under Assumption 1, when $\alpha>0$ is sufficiently large, there exists an unique optimizer of \eqref{eq: lagrangian_kl_dro} given by
\begin{equation}
    g^*({\bf x}^+) = \frac{\exp(\frac{f({\bf x}^+)}{\alpha})}{\mathbb{E}_0\left[\exp{\frac{f(X^+)}{{\alpha}}}\right]}
    \label{eq:optimal_l}
\end{equation}
\end{proposition}
%{\bf Proposition 1} 
%This result is known (e.g., \citep{Petersen}, \citep{Lam2016RobustSA}) and for the complete proof please refer to the appendix section Proof of Proposition 3.1 from \citep{Lam2016RobustSA}.

%By the sufficiency result in Chapter 8, Theorem 1 in \citep{Luenberger1968OptimizationBV}, suppose that we can find $\alpha^*\geq 0$ and $g^*\in \mathcal{L}$ such that $g^*$ maximizes \ref{eq: lagrangian_kl_dro} for $\alpha = \alpha^*$ and $\mathbb{E}_0[g^*\log g^*] = \frac{\lambda}{n}$ then $g^*$ is the optimal solution for  \ref{eq: optimization_likelihood_space}. 
Assume that such $\alpha^*$ and $g^*$ exist and that $\alpha^*$ is sufficiently large then 
$$\frac{\lambda}{n} = \mathbb{E}_0[g^*\log g^*] = \frac{\mathbb{E}_0[{g^*}f(X^+)]}{\alpha} -\log\mathbb{E}_0\left[\exp\left({\frac{f(X^+)}{\alpha^*}}\right)\right] $$
$$= \frac{\beta^*\mathbb{E}_0[f(X^+)\exp(\beta^*f(X^+))]}{\mathbb{E}_0[\exp(\beta^*f(X^+))]}-\log \mathbb{E}_0[\exp{\beta^*f(X^+)}]$$
$$ = \beta^*\psi^{'}(\beta^*)-\psi(\beta^*)$$
where we define $\beta^*=\frac{1}{\alpha^*}$ and ${\psi}(\beta) = \log \mathbb{E}_0[\exp(\beta f(X^+))]$ is the logarithmic moment generating function of $f(X^+)$.

We can write the optimal solution of the objective function \eqref{eq: optimization_likelihood_space} as follows
\begin{equation}
    \mathbb{E}_0[f(X^+)g^*] = \frac{\mathbb{E}_0[f(X^+)\exp(\frac{f(X^+)}{\alpha^*})]}{\mathbb{E}_0[\exp(\frac{f(X^+)}{\alpha^*})]} = \psi^{'}(\beta^*)
    \label{eq: solution1_dro}
\end{equation}
Now let us perform Taylor expansion of the following
$$\beta \psi^{'}(\beta) - \psi(\beta) = \sum_{m=0}^\infty \frac{1}{m!}\kappa_{m+1}\beta^{m+1}-\sum_{m=0}^\infty\frac{1}{m!}\kappa_m\beta^{m}$$
$$=\sum_{m=1}^\infty \left[\frac{1}{(m-1)!}-\frac{1}{m!}\right]\kappa_m\beta^m $$
$$= \sum_{m=2}^\infty \frac{1}{m(m-2)!}\kappa_m\beta^m = \frac{1}{2}\kappa_2\beta^2+\frac{1}{3}\kappa_3\beta^3+\frac{1}{8}\kappa_4\beta^4+\mathcal{O}(\beta^5)$$
In the above expression, $\kappa_m = \psi^{(m)}(0)$ is the m-th derivative of $\psi$ with evaluated at $\beta = 0$ and $\mathcal{O}(\beta^5)$ is continuous in $\beta$. By Assumption 2, we have $\kappa_2>0$. Therefore, for small enough $\frac{\lambda}{n}$, above equation reveals that there is a small $\beta^{*}>0$ that is root to the equation $\frac{\lambda}{n} = \beta\psi^{'}(\beta)-\psi(\beta)$ and the root is unique. This is because by Assumption 2, $\psi(.)$ is strictly convex, and therefore, $\frac{d (\beta\psi^{'}-\psi(\beta)}{d\beta}) = \beta\psi^{''}(\beta)>0$ for $\beta>0$, so that $\beta\psi^{'}(\beta)-\psi(\beta)$ is strictly increasing. 

Since $\alpha^*=\frac{1}{\beta^*}$, this shows that for any sufficiently small $\frac{\lambda}{n}$, we can find a large $\alpha^*>0$ such that the corresponding $g^*$ in \ref{eq:optimal_l} satisfies $\frac{\lambda}{n} = \mathbb{E}_0[g^*\log g^*]$. This means we can write the following
\begin{equation}
    \frac{\lambda}{n} = \frac{1}{2}\kappa_2\beta^{*^{2}}+\frac{1}{3}\kappa_3\beta^{*^{3}}+\frac{1}{8}\kappa_4\beta^{*^{4}}+\mathcal{O}(\beta^{*^{5}})
\end{equation}

We can obtain $\beta^*$ as follow

$$    \beta^* = \sqrt{\frac{2\lambda}{n\kappa_2}}\left(1+\frac{2}{3}\frac{\kappa_3}{\kappa_2}\beta^*+\frac{1}{4}\frac{\kappa_4}{\kappa_2}\beta^{*^{2}}+\mathcal{O}(\beta^{*^{3}})\right)^{-\frac{1}{2}}$$
$$= \sqrt{\frac{2\lambda}{n\kappa_2}}\left(1-\frac{1}{3}\frac{\kappa_3}{\kappa_2}\beta^*+\mathcal{O}(\beta^{*^{2}})\right) 
= \sqrt{\frac{2}{\kappa_2}}\left(\frac{\lambda}{n}\right)^{1/2}-\frac{2}{3}\frac{\kappa_3}{\kappa_2^2}\frac{\lambda}{n}+\mathcal{O}\left(\left(\frac{\lambda}{n}\right)^{\frac{3}{2}}\right)$$
In the above expression, first we use the binomial expansion $(1+x)^{\frac{-1}{2}} = 1-\frac{1}{2}x+\frac{3}{8}x^2.... $ followed by substitution of $\beta^*$ in the second term.
Now, the corresponding optimal solution becomes following

$$\mathbb{E}_0[f(X^+)g^*] = \psi^{'}(\beta^*) = \kappa_1+\kappa_2\beta^*+\kappa_3\frac{\beta^{*^2}}{2}+\mathcal{O}(\beta^{*^{3}})$$
$$=\kappa_1+\sqrt{2\kappa_2}\left(\frac{\lambda}{n}\right)^{\frac{1}{2}}+\frac{1}{3}\frac{\kappa_3}{\kappa_2}\frac{\lambda}{n}+\mathcal{O}\left(\left(\frac{\lambda}{n}\right)^{\frac{3}{2}}\right)$$
In the above equation $\kappa_1 = \bar{f}, \kappa_2 = \text{Var}_n[f(X^+)], \kappa_3 = \mathbb{E}_0[(f(X^+)-\mathbb{E}_0[f(X^+)])^3]$. This completes the proof of Theorem 2.
%\end{proof}

\section{More Examples of Challenging Bags}\label{app:examplebags}\label{app:example_of_20news}
 Figure~\ref{fig:morebagexamples} shows the $p$-$f$ plots for three example challenging bags from three different topics in the 20NewsGroup dataset. As shown, the highest $f$-score from those bags is very low. This implies that the passive learning model predicts all the instances as negative with a high confidence. Using F-Entropy, we may not be able to query any instance from those bags because of low uncertainty. In contrast, by leveraging the standard MIL assumption, the proposed P-F sampling will effectively explore those bags. Once the positive instances from these bags are queried, they  help to accurately identify similar positive instances in the same and different bags to boost the instance prediction performance, as evidenced by our experimental results.

\section{Link to Source Code}\label{app:sourcecode}
For the source code of our experiments, please \href{https://github.com/ritmininglab/ADMIL-P-F}{click here}.

\end{document}